\documentclass[acmsmall,screen]{acmart}
\usepackage{threeparttable}
\usepackage[utf8x]{inputenc}
\usepackage{subfigure}

\usepackage{amsmath}
\usepackage{graphicx}

\usepackage{multicol,balance}

\usepackage{multirow}

\def\BibTeX{{\rm B\kern-.05em{\sc i\kern-.025em b}\kern-.08emT\kern-.1667em\lower.7ex\hbox{E}\kern-.125emX}}

\setcopyright{acmcopyright}
\acmJournal{CSUR}

\begin{document}

%
\title{When Machine Learning Meets Privacy: A Survey and Outlook}

%
\author{Bo Liu}
\email{bo.liu@uts.edu.au}
\orcid{1234-5678-9012}
\authornotemark[1]
\affiliation{%
  \institution{University of Technology Sydney, Australia}
  \streetaddress{}
  \city{Ultimo}
  \state{NSW}
  \postcode{2007}
}

\author{Ming Ding}
\affiliation{%
  \institution{Data61, CSIRO, Australia}}
\email{ming.ding@data61.csiro.au}

\author{Sina Shaham}
\affiliation{\institution{The University of Sydney, Australia}}
\email{sina.shaham@sydney.edu.au}

\author{Wenny Rahayu}
\affiliation{ \institution{La Trobe University, Australia}}
\email{W.Rahayu@latrobe.edu.au}

\author{Farhad Farokhi}
\affiliation{ \institution{The University of Melbourne, Australia}}
\email{Farhad.Farokhi@unimelb.edu.au}

\author{Zihuai Lin}
\affiliation{\institution{The University of Sydney, Australia}}
\email{zihuai.lin@sydney.edu.au}

%
\renewcommand{\shortauthors}{Bo Liu, et al.}

%
\begin{abstract}
The newly emerged machine learning (e.g. deep learning) methods have become a strong driving force to revolutionize a wide range of industries, such as smart healthcare, financial technology, and surveillance systems. Meanwhile, privacy has emerged as a big concern in this machine learning-based artificial intelligence era. It is important to note that the problem of privacy preservation in the context of machine learning is quite different from that in traditional data privacy protection, as machine learning can act as both friend and foe. Currently, the work on the preservation of privacy and machine learning (ML) is still in an infancy stage, as most existing solutions only focus on privacy problems during the machine learning process. Therefore, a comprehensive study on the privacy preservation problems and machine learning is required. This paper surveys the state of the art in privacy issues and solutions for machine learning. The survey covers three categories of interactions between privacy and machine learning: (i) private machine learning, (ii) machine learning aided privacy protection, and (iii) machine learning-based privacy attack and corresponding protection schemes. The current research progress in each category is reviewed and the key challenges are identified. Finally, based on our in-depth analysis of the area of privacy and machine learning, we point out future research directions in this field.
\end{abstract}

%
%


%

%

%
\maketitle

\section{Introduction}
\label{section:introduction}

Since Facebook data privacy scandal in 2018~\cite{Times2020},
privacy has once again become a dominant feature in people’s minds.
This motivates revisiting privacy challenges, particularly with the emergence of intelligent technologies thanks to the big data revolution.
For example, 
newly emerged machine learning (ML) techniques, 
especially the unprecedented powerful deep learning, 
will have paradigm-shifting
impacts on privacy preservation. 
A critical question that needs to be well investigated is: 
What are the privacy challenges and solutions associated with ML?

Some initial work has appeared in the literature with an emphasis on mitigating privacy risks during the machine learning process by paying special attention to the privacy challenges and risks associated with the ML models. In this regard, 
possible attack models~\cite{Fredrikson2015, Tramer2016, Ateniese2015, Shokri2017, melis2019exploiting, Song2017} have been discussed and protection schemes~\cite{Abadi2016, Phong2018, Shokri2015, Bonawitz2017, Papernot2017} have been proposed.
These works demonstrated both ML models and training datasets can be the target of privacy attacks, 
leading to sensitive information leakage. 
Meanwhile, 
researchers have also tried to use ML for privacy protection. 
As an example, 
the authors of~\cite{Yu2017} have developed a method for automatic recognition of privacy-sensitive object classes and adjust users' privacy preference settings. 
In addition, 
there are also several works that develop new privacy protection schemes in the scenarios where ML is used for attacks~\cite{Liu2017, Liu2019a}. 
Overall, 
the current research has only scratched the surface, 
and there are major issues that require further investigation:
\begin{itemize}
  \item ML could play different roles in a privacy protection problem, 
  e.g., protection target, attack tool, and/or protection tool. 
  It may even play multiple roles in the same problem.
  \item ML systems and models have different types, 
  each facing different privacy risks and requires different protection schemes.
  \item There does not exist a unified privacy metric or notion.
  Although differential privacy (DP)~\cite{Dwork2006} is widely accepted in traditional privacy studies, it still has limitations in the context of ML, 
  especially when considering unstructured data, such as text, image, and video.
\end{itemize}

In this context, 
a systematic study of privacy and ML is essential for future research efforts. 
Although there are several surveys on this topic \cite{ji2014, Abadi2017, Zhang2018, Liu2018a}, 
The focus has been on a certain type of ML model or specific methods.

This study attempts to provide the first comprehensive survey on privacy in ML by investigating different scenarios/applications of privacy and ML.
The main contributions of the paper are as follows:
\begin{itemize}
  \item We divide the works in this area by the different roles of ML, 
  i.e., ML as protection target (private ML), 
  protection tool (ML enhanced privacy protection), 
  attack tool (ML-based attack), 
  and analyze the problems and solutions in each category.
  \item For private ML, 
  we categorize the attacks and protection schemes and then compare their difference.
  \item For ML aided privacy protection and ML-based privacy attack, 
  we not only discuss the existing works, 
  but also provide insights on new techniques to achieve privacy preservation.
  \item The study concludes with a discussion on the directions of future research in ML and privacy.
\end{itemize}

Through this comprehensive overview, 
we wish to prepare a solid ground for future research in this field.

The rest of the paper is organized as follows. Section \ref{sec:PrivacyThreatsML} reviews basic concepts of machine learning system and models, and discusses the relationship between privacy and ML.
In Section~\ref{section:PrivateML}, 
we compare and classify existing privacy attacks and protection schemes in ML systems. 
Section~\ref{section:MLaidedaprotection} focuses on ML aided privacy protections, 
followed by the discussion of ML-based attack and corresponding privacy preservation schemes in Section~\ref{Sec:PrivacyAgainstML}. We present our outlook and propose some future directions for this promising research topic in Section~\ref{Sec:outlook}. Finally, 
we conclude our work with a summary in Section~\ref{section:conclusion}.

Moreover, 
the abbreviations used in this paper are listed in Table~\ref{tab:symbols}.

\begin{table}[htbp]
\footnotesize
\renewcommand\arraystretch{1}
\begin{center}
\caption{\label{tab:symbols}Summary of acronyms used in the paper.}
\begin{tabular}{|l|l|}
\toprule
\hline
CNN & convolutional neural network \\
DNN & deep neural network \\
DP$^{\textasteriskcentered}$ & differential privacy\\
ERM & empirical risk minimization \\
FGSM & fast gradient sign method\\
FHE & fully homomorphic encryption \\
GAN & generative adversarial network\\
GNN & generative neural network\\
IoT & Internet of things\\
ML & machine learning \\
SGD & stochastic gradient descent \\
SMC  & secure multi-party computation\\
SVM & support vector machine\\
VAE & variational autoencoder\\
\hline
\end{tabular}
\begin{tablenotes}
\item {\textasteriskcentered} DP in this survey is used as the abbreviation for Differential Privacy, not deep learning.
\end{tablenotes}
  \end{center}
\end{table}

\section{Privacy Threats and Machine Learning}
\label{sec:PrivacyThreatsML}

In this section, 
we discuss the privacy threats in the context of machine learning, 
and further point out various roles of machine learning in the studies of user privacy. 

\subsection{The Machine Learning System and Models}
ML refers to algorithms and statistical models used by computer systems to efficiently perform specific tasks without the use of explicit instructions. It relies on an automated learning process. The ML algorithm constructs a mathematical model of sample data called a "training set" to make predictions or decisions~\cite{Bishop2006}.

Depending on if the output is labelled in the training set, 
ML models can be divided into three different groups: 
supervised, unsupervised, and semi-supervised. 
As supervised learning is used by most practical machine learning algorithms, 
it will be explained here as an example.

A supervised ML model is a parameterized function $f_\theta$ that maps input data $\vec{x} \in \mathbb{X}^d$ (generally a vector of features) to output data $y \in \mathbb{Y}$ (label). 
For a classification problem, 
$\mathbb{X}^d$ is a $d$-dimensional vector space and $\mathbb{Y}$ is the set of classes. 
This function is trained to accurately predict the label of new data that have not seen before.

Moreover, we can divide the ML process into two stages:

\begin{enumerate}
  \item Model training: 
  The training process of a machine learning model is to find the optimal parameters that can accurately capture the relationship between $\mathbb{X}$ and $\mathbb{Y}$. 
  To achieve this, 
  a training dataset $D=\{\vec{x}_i,y_i\}^N_{i=1}$ with $N$ samples is needed. 
  Then a loss function $L$ is adopted to quantify the difference between two outputs, 
  i.e. the ground-truth one $y_i$ and the predicted one $f_\theta(\vec{x}_i)$. 
  The goal of training a model is to minimize this loss function, i.e.,
      \begin{align}
      \theta^\star = \arg \min_\theta(\sum_{i}L(y_i,f_\theta(\vec{x}_i))+\Omega(\theta)),
      \end{align}
      where $\Omega$ is a regularization term to penalize model complexity and avoid overfitting.
  \item Model inference/prediction: 
  After the model training is completed and the optimal parameters $\theta^\star$ are obtained, 
  given an input $\vec{x}$, 
  the corresponding output can be calculated as $y = f_\theta^\star(\vec{x}_i)$. 
  This prediction process is called inference. 
  We can calculate the prediction accuracy of the model over a testing dataset $D_t$ to measure the model's performance. 
\end{enumerate}

Furthermore, 
according to the architecture of the ML systems, 
there are two different models, 
as shown in Fig.~\ref{fig:ML_models}:

\begin{figure}[t!]
\centering
   \subfigure[centralized learning (can be outsourced learning).]{\label{fig:centralized_learning(outsource)}
\includegraphics[height=3.8cm]{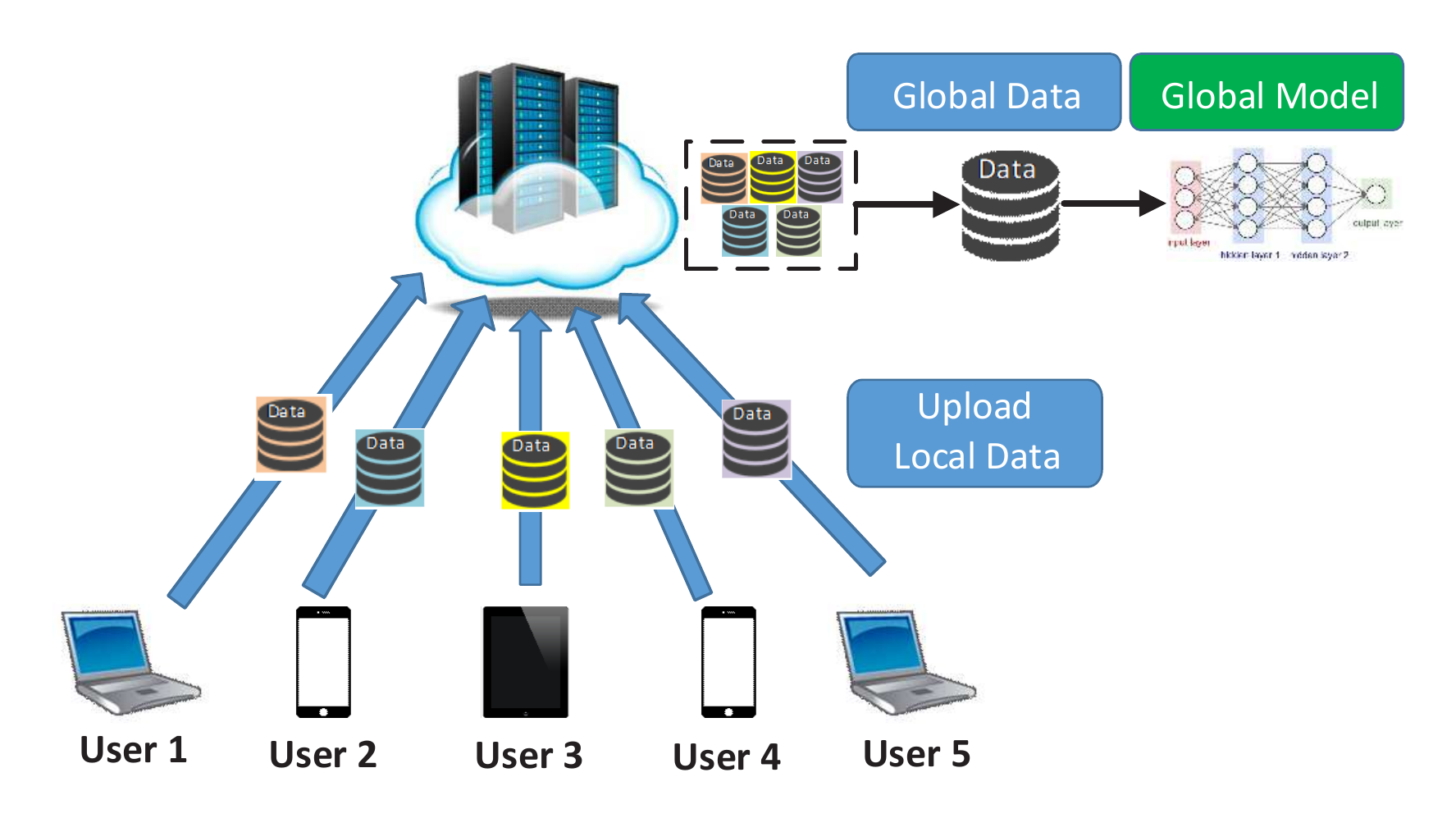}}
\subfigure[distributed learning]{ \label{fig:collaborative_learning}
\includegraphics[height=3.8cm]{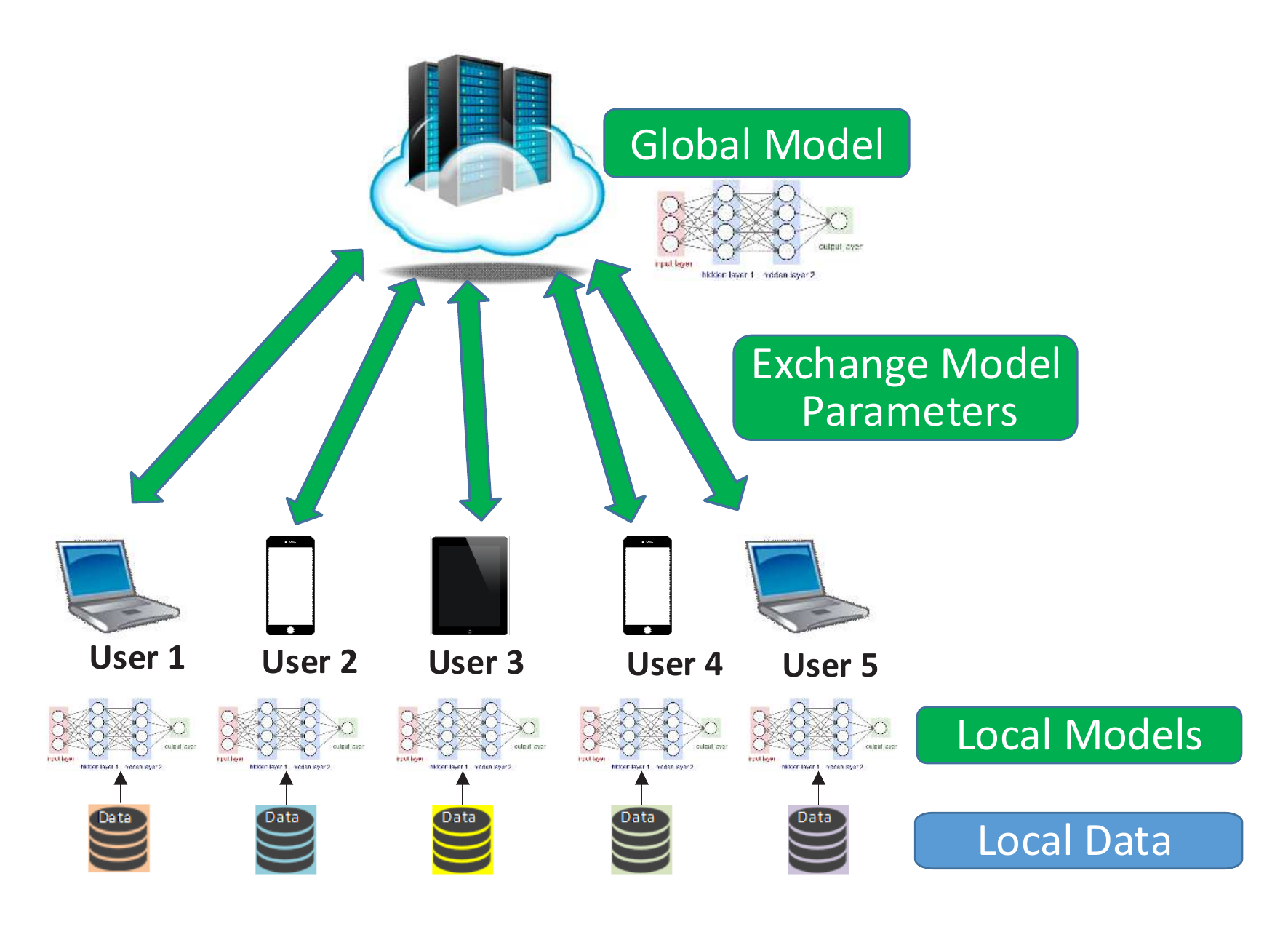}}
\caption{Centralized and distributed ML systems: (a) centralized learning; (b) distributed learning.}\label{fig:ML_models}
  \end{figure}
  
\begin{itemize}
  \item Centralized learning: 
  The training data is centralized in a machine or in a data center, 
  and the centralized entity trains and hosts the models. 
  For example, 
  a researcher could use a cloud platform, 
  to host datasets and train an AI model based on them. 
  It goes without saying that the availability of all data in such a centralized method leads to high efficiency and accuracy~\cite{Hitaj2017}.  
  However, 
  because the centralized operator has direct access to sensitive data, user privacy might be violated.

   As the learning tasks become more and more complicated, 
   many companies start to outsource the training process, 
   i.e., \emph{outsourced learning}, or \emph{ML-as-a-service}. 
   In this case, 
   each user owns his/her training data while the service providers own the models and algorithms. 
   The data holder outsources model creation to a cloud service such as Microsoft Azure ML and Amazon AWS ML, 
   which automate the process of ML. 
   ``Users upload datasets, 
   perform training, 
   and make the resulting models available for use''~\cite{Song2017}. 
   During this process, 
   the users do not have any understanding of the details of model creation. 
   The ``ML provider is the entity that provides ML training codes to data holders''~\cite{Song2017}.
  \item Distributed learning: Centralized learning is sometimes not a good option for several reasons: (i) data is inherently distributed in some scenarios; 
(ii) data is too large to be stored in a single machine; 
(iii) users are not willing to share raw data; 
and (iv) users want to train the neural network with different instances to achieve better predication accuracy. 
In this case, 
ML can be conducted in a distributed manner, 
i.e., distributed learning. 
In general, 
distributed learning is used in a scenario of distributed training data sources and a centralized server. 
There are several variations of distributed learning:
\begin{itemize}
    \item Collaborative learning: Distributed learning involving such collaborations is known as \textit{collaborated learning}. 
    But the settings could be quite different in the literature. 
    For example, 
    the authors of~\cite{Song2018} proposed a collaborative learning framework that trains several classifiers ``simultaneously on the same training data'' to achieve better performance. 
    On the other hand, 
    in the collaborative learning model defined in~\cite{Hitaj2017}, 
    each participant uses its device to train a local AI model. 
    It then shares a fraction of the parameters/coefficients of the model with the other users. 
    Service operators can create a composite model by collecting these parameters and achieve almost the same accuracy as a model built using a centralized approach. 
    The collaborative approach is ``more privacy-friendly'' because the dataset is not directly exposed. 
    Also, 
    if only a small part of the model parameters is shared and the parameters are truncated and/or obfuscated by DP mechanisms, 
    the model exhibits convergence through experiments~\cite{Shokri2015}.
    \item Federated learning: A popular framework for collaborative learning is \textit{Federated learning}~\cite{Konecny2016} introduced by Google. There are currently two different federated learning settings: cross-device and cross-silo~\cite{Kairouz2019}. The cross-device setting normally involves a very large number of mobile or IoT devices, while in the cross-silo settings it ``might involve only a small number of relatively reliable clients’’~\cite{Kairouz2019}, e.g., multiple organizations. In a broader definition of federated learning that covers both settings, each device downloads the current model from a centralized server, 
    improves it by learning from data on a local device, 
    and then sums up the changes in a focused update. 
    Here, ``focused updates are updates'' containing ``the minimum information necessary for the specific learning task’’~\cite{Kairouz2019}. 
    And then the shared model is updated by averaging all users' updates. 
    Since all the training data will not leave local devices, 
    and no updates from individual users are stored in the cloud, 
    the privacy risk has been greatly reduced.
  \item Split learning: Another collaborative learning framework is \textit{Split learning}, in which each user trains the network up to a certain layer known as the cut layer and sends the weights to server. 
    Mathematically speaking, 
    these weights represent and compress the input data to some intermediate feature vectors. 
    The server then trains the network for rest of the layers, 
    and generates the gradients for the final layer, 
    followed by error back-propagation until the cut layer.
    The gradient is then passed over to the users. 
    The rest of the back-propagation is completed by the users~\cite{Vepakomma2018}. 
    In split learning, 
    ``client-side communication costs are significantly reduced as the data to be transmitted is restricted to first few layers of the split neural network prior to the split''.
\end{itemize}
Although some collaborative learning models consider shared training data~\cite{Song2018}, 
which presents a significant privacy risk. 
In this survey, 
however, 
we consider the case that the local raw training data are not shared with the server or amongst users. 
In this learning process, 
the users can collaboratively learn a shared ML model,
thus decoupling ML tasks from the storage of the data in a single device. 
\end{itemize}
Overall, centralized learning is characterized by ``globally stored data'' and ``globally trained model'', 
as shown in Fig.~\ref{fig:centralized_learning(outsource)}, 
while the distributed learning is characterized by ``locally stored data'' and ``locally trained model'', 
as shown in Fig. \ref{fig:collaborative_learning}. 
Although there will be a global model in distributed learning, 
it is not trained globally, 
at least part of the model is trained by individual clients.

\subsection{Relationship of privacy and machine learning}

In contrast to traditional privacy-related research frameworks, 
ML techniques open new challenges and opportunities to privacy protection. 
There has been some initial research embarking on this journey. The existing works can be divided into three categories according to the roles of ML in privacy.

\begin{figure}[t!]
\centering
\includegraphics[width=10cm]{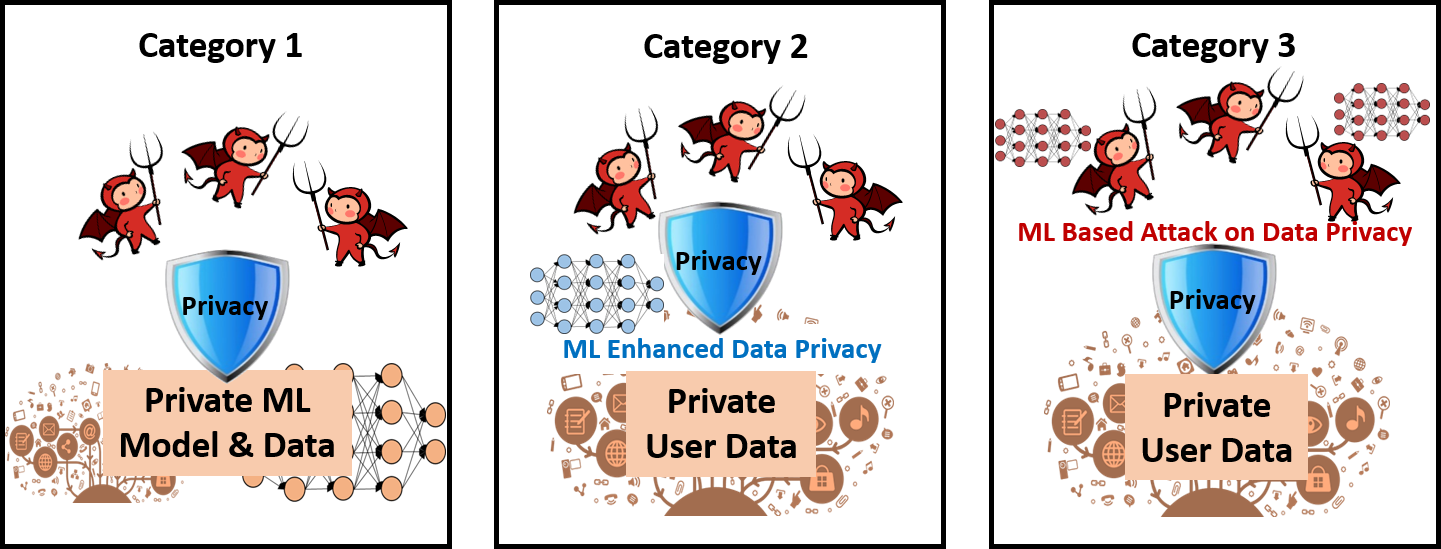}
\hspace{1em}
\centering
\caption{Three different categories of research problems in privacy and ML: (a) Privacy of ML model and data; (b) ML enhanced privacy protection; (c) ML-based privacy attack.}
\label{fig:privacy_categories}
\end{figure}

First, making ML system private, i.e., ML system is the target of privacy protection. As shown in Fig.~\ref{fig:privacy_categories}(a), this category 1 includes making both the ML system (model parameter) and data (training/test dataset and output data) private, 
since the privacy threat may happen in any stage of a data cycle, e.g. the training, publishing, or prediction of data. 
Most of the research in this group relies on the use of differential privacy in ML and deep learning models~\cite{Goodfellow2018}.
For example, Shokri et al. \cite{Shokri2015} developed a differentially private SGD algorithm and a distributed deep learning model training system. 
In such way, multiple entities can cooperatively learn a neural network.

Second, using ML to enhance privacy protection. As shown in Fig.~\ref{fig:privacy_categories}(b), the privacy protection target is the data in this category 2 and ML is a tool to help privacy protection. For example, Liu et al. \cite{Liu2016} utilized ML to enhance private decision-making experience through ML. Orekondy et al. \cite{Orekondy2017} proposed an approach to categorize personal information in images and predict information leakage directly from images. Yuan et al.~\cite{Yuan2017} presented an ML approach to decide whether to share a picture with a specific requester for a particular context.

Third, ML-based privacy attack, i.e., ML is used as an attack tool of the adversary, as shown in Fig. \ref{fig:privacy_categories}(c). 
For example, recent researches have shown that deep learning methods can be used to detect object types, people's  identities, and landmarks, from images posted on Internet. 
When the adversaries use this kind of powerful tools, 
conventional privacy protection methods would be over-powered,
especially being challenged by the mighty deep learning tools. There have been very few works in this category. Liu et al. \cite{Liu2019a} proposed schemes of applying adversarial perturbations images, so that ML systems cannot get private information from them. 

Table~\ref{tab:Cate} summaries three categories of privacy protection problems involving ML systems. 
It is worth mentioning that one technique might belong to more than one category.
For instance, ML might be used as attack and protection tools at the same time, 
which makes the problem more complicated. We will discuss this in more detail in the reminder of the paper.

\begin{table}[htbp]
\footnotesize
\renewcommand\arraystretch{1}
\begin{center}
\caption{\label{tab:Cate}Three categories of privacy protection problems in the context of ML.}
\begin{tabular}{c|c}
\toprule
\hline
{\textbf{Category}} &{\textbf{Role of ML in Privacy Protection}} \\ \hline
Private ML & Protection target \\ \hline
ML enhanced Privacy Protection & Protection tool\\ \hline
ML-based Privacy Attack &Attack tool \\ \hline
\bottomrule
 \end{tabular}
  \end{center}
\end{table}

\begin{figure}[t!]
\centering
\includegraphics[scale=0.6]{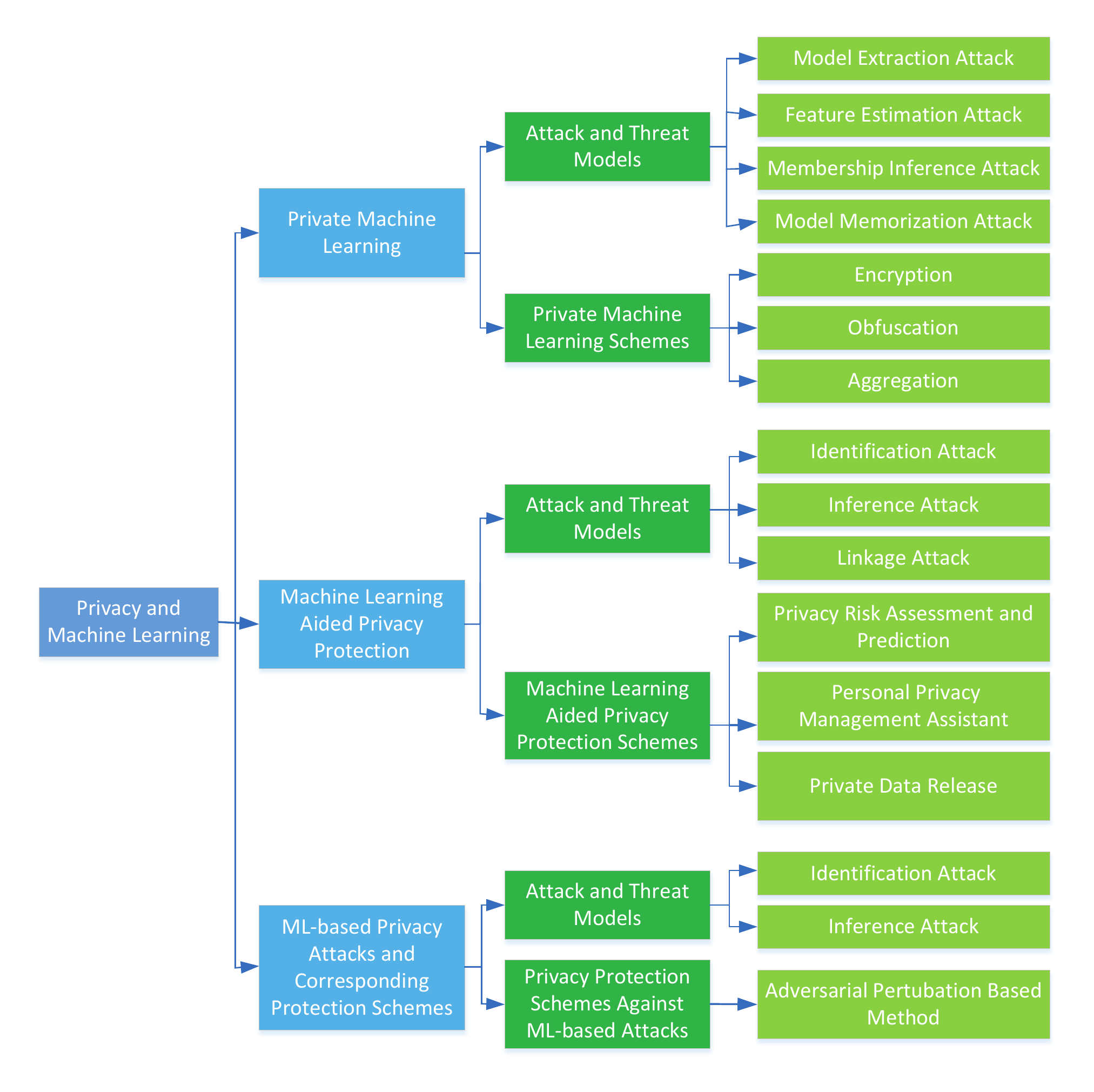}
\hspace{1em}
\centering
\caption{The proposed taxonomy of privacy and ML.}
\label{fig:Taxonomy}
\end{figure}

Fig.~\ref{fig:Taxonomy} summarizes the general taxonomy of the research papers presented in this work. We divide them according to the above mentioned three categories. In each category, we discuss the attack and threat models first and then analyze the works on privacy protection schemes.

\section{Private Machine Learning}
\label{section:PrivateML}

In this section, 
we will discuss the challenges and existing solutions in privacy preservation in ML, 
or simply stated, private ML.

We will first discuss attack and threat models, followed by detailed analysis of privacy preservation schemes, along with some comparisons at the end.

\subsection{Attack and Threat Models}
\label{subsec:AttackModels}

In this subsection, 
we analyze the attack models from three perspectives: 
the attack targets, 
the knowledge of the adversary, 
and the attack methods.

First, as we can see in Section \ref{sec:PrivacyThreatsML}, model and data are two important components in ML that correspond to two different categories of privacy attack targets, 
as shown in Fig.~\ref{fig:privacy_targets}:

\begin{figure}[t!]
\centering
\subfigure[Privacy of the ML model.]{\label{fig:model_privacy}
\includegraphics[height=4.8cm]{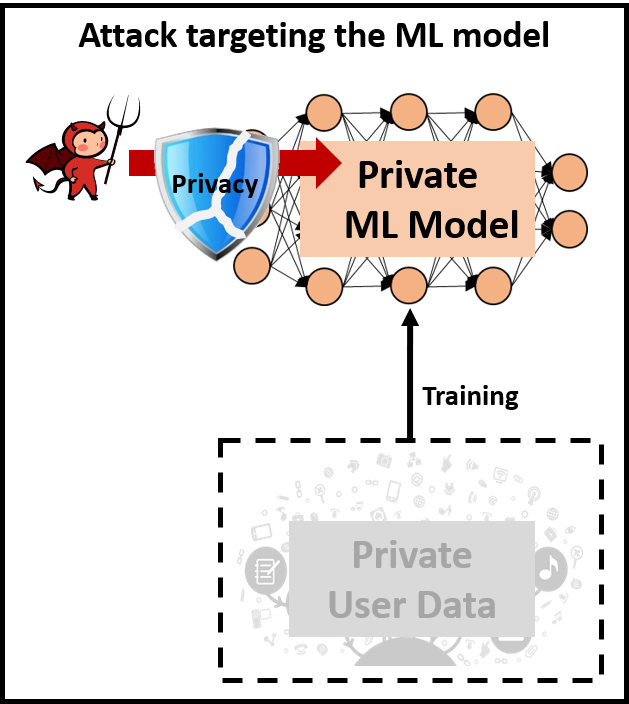}}
\hspace{4em}
\subfigure[Privacy of the underlying data.]{\label{fig:data_privacy}
\includegraphics[height=4.8cm]{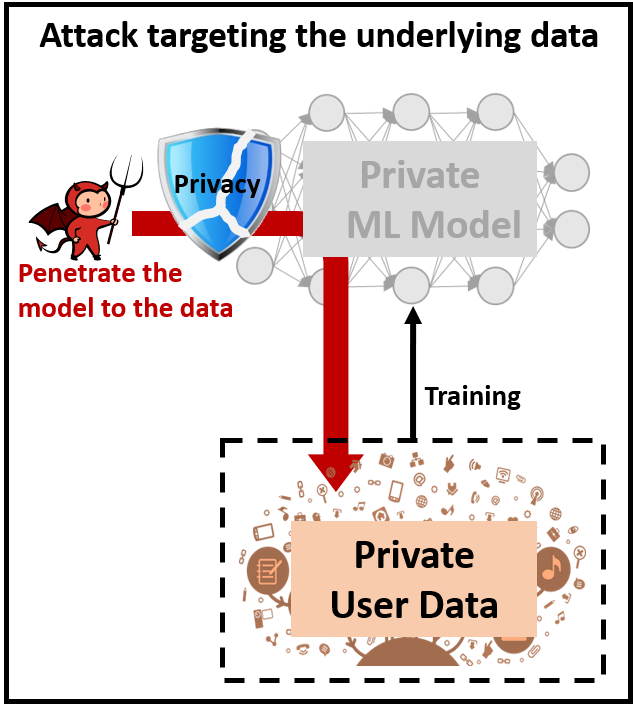}}
\caption{Two different types of privacy attack targets in ML: (a) Model privacy; (b) Training data privacy.}
\label{fig:privacy_targets}
\end{figure}

\begin{enumerate}
  \item Training data privacy: 
  In many cases, 
  a user wants to keep the training data private while using a ML service. 
  For example, 
  for a medical study or a hospital having a model built out of the private medical profiles of some patients. A patient may want to use the model to make a prediction about whether she is likely to contract a certain disease, 
  or the hospital may want to use the model to predict the probability of readmittance. 
  In these cases, 
  the training data is sensitive medical profiles and should not be revealed. 
  Similar cases exist in other areas such as financial records. More specifically, training data privacy includes exact data value, certain features, statistical properties, or membership (whether a certain data is in the training set).
  \item Model privacy: 
  There are also privacy concerns about the ML model including the model parameters, 
  and training algorithms. 
  For example, 
  a financial institution may hold a sensitive model which can accurately predict stock prices or insurance rates. The model is an important commercial and intellectual property. 
  Another example is the commercial ML API services currently provided by Google, Amazon, Microsoft, and other companies. 
  They charge the customers per API access. Revealing their models or algorithms will cause loss of revenue. In summary, the attack target can either be the model structure or parameters.
\end{enumerate}

Second, 
the adversaries have different levels of knowledge according to their access to the information.
\begin{itemize}
  \item White-box access: 
  The adversary has access to the trained model,
  especially the model parameters.
  \item Black-box access: 
  The adversary is an end-user and is only allowed to query the prediction model on his/her inputs through an appropriate interface.
\end{itemize}

Finally, 
the adversary can adopt different attack methods. 
Existing attack methods include model inversion (reverse engineering), 
shadow training models, 
and encoding information into models.

\begin{figure}[th!]
\centering
   \subfigure[Model Extraction Attack]{\label{fig:Model Extraction Attack}
\includegraphics[scale=0.38]{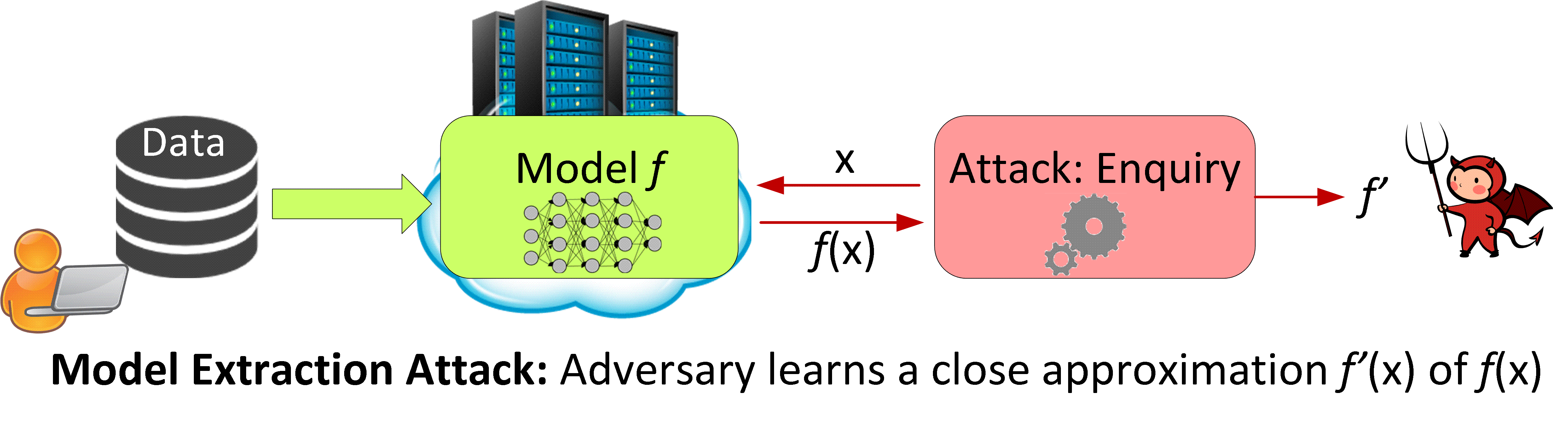}}
\hspace{-1cm}

\subfigure[Feature Estimation Attack]{ \label{fig:feature_estimation}
\includegraphics[scale=0.38]{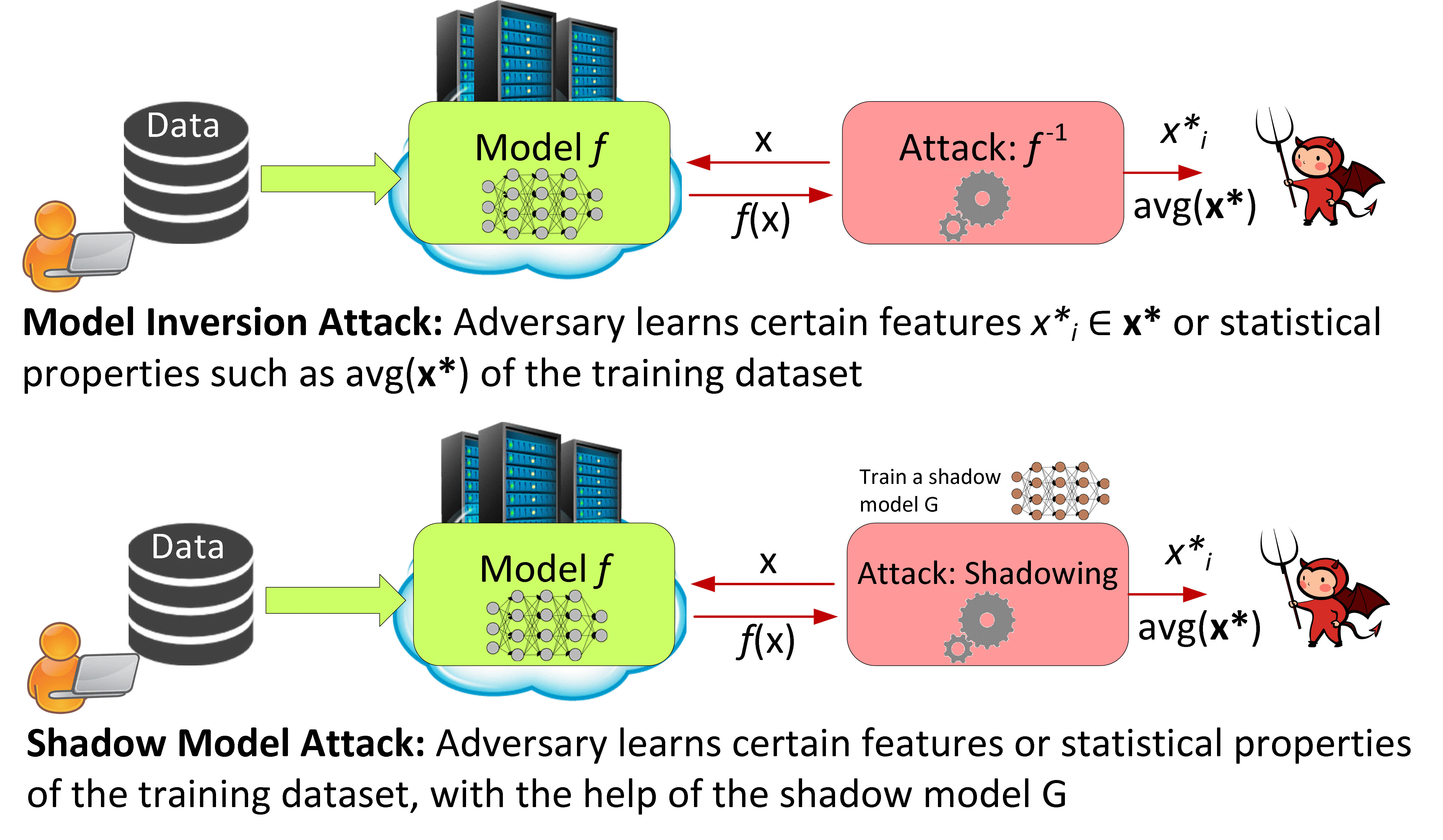}}
\hspace{-1cm}

\subfigure[Membership Inference Attack]{\label{fig:membership_inference_attack}
\includegraphics[scale=0.38]{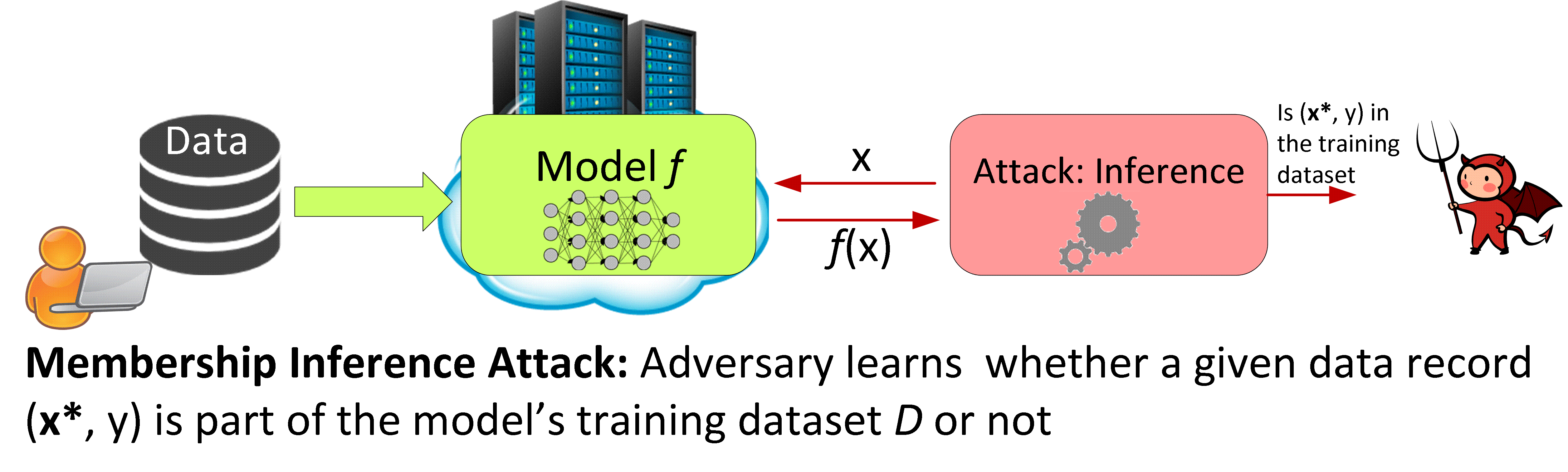}}
\hspace{-1cm}

\subfigure[Model Memorization Attack]{ \label{fig:model_memorization_attack}
\includegraphics[scale=0.38]{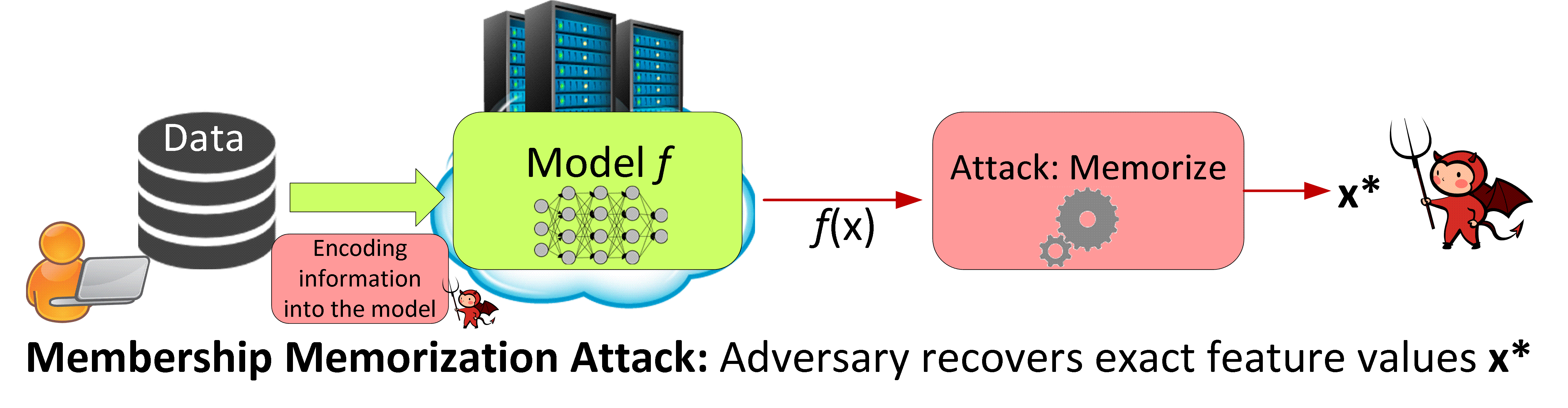}}
\hspace{-1cm}

\caption{Different attack models targeting ML.}\label{fig:attack_models}
\end{figure}

Next, we will group existing popular attack models by attack targets and analyze them from the above mentioned three aspects. 
An illustrative diagram of the attack models is presented in Fig.~\ref{fig:attack_models}.

\subsubsection{Model Extraction Attack}
The model extraction attack targets at the duplication of (i.e., ``steal'') the AI model~\cite{Tramer2016}. 
The outcome of the attack will be a function $f^\prime$ that is approximately the same as the initial function $f$.
An illustration of such an attack can be found in Fig.~\ref{fig:Model Extraction Attack}.

In this attack, the adversary only has black-box access with no prior knowledge of the ML model parameters or training data. Tram\`{e}r et al. ~\cite{Tramer2016} used a shadow training scheme that can ``extract target ML models with near-perfect fidelity for popular ML models'' including logistic regression, decision trees, and neural networks, 
by equation-solving, path finding, or extending the Lowd-Meek approach~\cite{Lowd2005}.

There are several other works following this path. 
Oh et al. \cite{Oh2019} built meta-models to extract more model details such as the neural network architecture. 
Wang et al. \cite{Wang2018} designed an attack to steal the hyperparameters of the machine learning model. 
A hyperparameter is ``used to balance the loss function and regularization term in the objective function''. 
The adversary can obtain this value from the training set and model. Hua et al. \cite{Hua2018} ``investigated reverse-engineering attacks on CNN
models exploiting information leaks through memory and timing side-channels''.

\subsubsection{Feature Estimation Attack}

A feature estimation attack aims to estimate certain features $x^\star_i\in\vec{x}^\star$ or statistical properties such as $avg(\vec{x}^\star)$ of the training dataset~\cite{Fredrikson2014,Fredrikson2015}. 
In practice, 
it can be implemented by model inversion attack, shadow model attack or power side-channel attack.
An illustration of such an attack can be found in Fig.~\ref{fig:feature_estimation}.

First, \emph{Model Inversion Attack} mostly works in a white-box model, 
although it also can use black-box attack~\cite{Fredrikson2015} with lower effectiveness. Fredrikson et al.~\cite{Fredrikson2014} showed a white-box attack that can ``learn sensitive genomic information about individuals''. 
The basic idea of \cite{Fredrikson2014} is to complete the target feature vector ``with each of the possible values, 
and then computes a weighted probability estimate that this is the correct value'', 
given the knowledge of a linear regression model $f$. 
Then in~\cite{Fredrikson2015} they extended the attack to facial recognition models to achieve two different targets: 
the \emph{reconstruction attack} that produces ``an image of the person associated with a given label'' and the deblurring attack that generates the deblurred image of a certain individual given ``an image containing a blurred-out face''. The idea behind these attacks is ``to use gradient descent (GD) to minimize a cost function involving $f$''.

Overall, 
the model inversion attack works with a simple philosophy: 
we can reverse-engineer (find $f^{-1}$) by following the gradient in a trained network to adjust the weights and obtain the features for all classes in the network. 
Even for classes that we do not have prior information, we can still reproduce the prototype example. 
This type of attack suggests that any accurate deep learning machine, 
regardless of training methods, 
may leak information on the distinguishable classes. 
Extensive research has shown that generative adversarial network (GAN) generated sample data are similar to the training data.
And thus, 
the results given by the model inversion attack may even ``reveal more private information about the training data compared to the average samples''~\cite{Ateniese2015}.

Second, \emph{Shadow Model Attack} means the attacker trains other ML models to achieve the target.
It can happen in either black-box or white-box way. 
For example, Ateniese et al.~\cite{Ateniese2015} designed a ``meta-classifier that can be trained to hack into other ML classifiers to infer patterns or private information from the training set'', e.g. they were able to extract accent information from trained speech recognition systems.

Hitaj et al.~\cite{Hitaj2017} designed an attack in the context of collaborative learning. 
They consider the adversary is an insider of the collaborative learning process who wants to infer sensitive information from the peers. The adversary can see and use internal parameters of the model, so it is a white-box attack. The adversary uses GANs~\cite{Goodfellow2014} to extract and reconstruct information of the victim. 
``This process is similar to facial composite imaging used by the police to identify suspects, 
where the composite artist generates sketches based on eyewitness identification of the suspect's face. Although the composite artist (GAN) has never seen a real face, the final image is based on eyewitness feedback''~\cite{Hitaj2017}.

Finally, Wei et al.~\cite{Wei2018} proposed to use power side-channel attack on an FPGA-based convolutional neural network accelerator, which can successfully recover the input image using the power traces at the inference stage.


\subsubsection{Membership Inference Attack}
Membership inference attack refers to acquiring the knowledge about whether a certain data record $(\vec{x}^\star,
y^\star)$ belongs to the model's training dataset $D$ or not~\cite{Shokri2017, melis2019exploiting}.
An illustration of such an attack can be found in Fig.~\ref{fig:membership_inference_attack}.

Shokri et al.~\cite{Shokri2017} introduced a ``black-box membership inference'' that used a shadow training technique to imitate the behavior of the target model.
The trained inference model is used ``to recognize differences in the target model's predictions'' on training and non-training inputs. They also found that overfitting, the structure and type of the model are the main factors that cause a model to be vulnerable to membership inference attack. Long et al.~\cite{Long2018} and Yeom et al.~\cite{Yeom2018} investigated ``the relationship between overfitting and privacy leakage''. Salem et al.~\cite{Salem2019} proposed a membership inference attack method using an unsupervised binary classification, ``which does not need to train any shadow model and does not assume knowledge of model or data distribution''.

Membership inference attacks are also studied in Generative Adversarial Networks (GANs). For example, Liu et al.~\cite{Liu2018} trained an attacker network to launch membership attacks against Variational Autoencoders (VAEs) and GANs. Hayes et al.~\cite{Hayes2019} focused on ``generative models in ML-as-a-service applications and train GANs to recognize training inputs''.

Melis et al.~\cite{melis2019exploiting} studied membership inference in collaborative learning. 
The attack is achieved by ``analyzing periodic updates to the shared model during training''. The reason that this attack is effective is that the gradients in neural networks are based on features, 
``thus observations of the participants' gradient updates can be used to infer the feature values, 
which are in turn based on these participants' private training data''. Wang et al.~\cite{Wang2019} considered membership inference attack ``against the user-level privacy on the federated learning framework by the attack from a malicious server. The proposed attack framework exploits GAN with a multi-task discriminator, which simultaneously discriminates category, reality and client identity of input samples, and doing so recovers user-specific private data''.

\subsubsection{Model Memorization Attack}
Song et al.~\cite{Song2017} first proposed the model memorization attack that targets recovering the exact feature values on individual samples. 
They consider a ``malicious ML provider'' specialized in model-training for the customers. 
In such a business model, 
the provider does not observe the training, 
but has access to the resulting model. 
He can steal the sensitive samples and encode the values into the model parameters or outputs. 
Another malicious party can retrieve sensitive information from the model during model serving.
An illustration of such an attack can be found in Fig.~\ref{fig:model_memorization_attack}.

Model memorization attack can happen both in white-box and black-box cases. 
In the white-box case, Song et al.~\cite{Song2017} proposed several techniques for the adversary to encode sensitive data into the models. 
(1) LSB encoding: the adversary can encode the ``training dataset in the least significant (lower) bits of the model parameters''. 
(2) Correlated value encoding: the adversary can ``gradually encode information while training model parameters''. 
For instance, 
``the adversary can add a malicious term to the loss function which maximizes the correlation between the parameters and the data he wants to encode''. 
(3) Sign encoding: similar to correlated value encoding, the adversary can use ``the sign of model parameters to interpret as bit strings'', 
e.g., positive parameters represent 1 and negative parameters represent 0.

In the black-box case, 
the adversary is assumed to have no access to the model parameters. They designed a scheme in which the adversary can ``augment the training dataset with synthetic inputs whose labels encode the critical information''. 
Then the information is leaked via the outputs of these added inputs.

Model memorization attack studies how malicious training algorithms deliberately create models that leak information about their training data sets. ``This threat model is more generous to the adversary, so it can extract more information about the training data than any other attack''~\cite{Song2017}.

\subsection{Private Machine Learning Schemes}
\label{subsec:PrivateMLSchemes}

In this subsection, we present several private ML schemes, including encryption, obfuscation, and aggregation. 

\subsubsection{Encryption}

Encryption or cryptography-based methods can be divided into two groups:

\begin{itemize}
  \item Encrypting training data. 
  The mainstream technique is homomorphic encryption. 
  As adding homomorphic encryption to the process will make the process at least an order of magnitude slower, 
  initially it is applied on training data for relatively simple classifiers~\cite{Brickell2007, Graepel2012, Bost2015}. 
  For example, 
  Graepel et al.~\cite{Graepel2012} found that training over encrypted data is possible when the training algorithm can be expressed as a low degree polynomial. 
  Bost et al.~\cite{Bost2015} applied this technique in three classifiers: 
  hyperplane decision, Naive Bayes and decision trees. 
  Then researchers try to extend the work to deep neural networks (DNN). Dowlin et al.~\cite{Dowlin2016} proposed CryptoNets which demonstrates how to efficiently convert learned neural networks to make it applicable to encrypted input data. 
  While Hesamifard et al.~\cite{Hesamifard2018} proposed a framework to train the neural network over encrypted data. Li et al.~\cite{Li2017} investigate the case of collaborative learning where datasets are encrypted with different keys, 
  and propose a solution based on multi-key fully homomorphic encryption (FHE).
  \item Encrypting ML model. 
  The encryption technique is also used to protect the model privacy. Phong et al.~\cite{Phong2018} proposed to use ``additively homomorphic encryption on the gradients''. 
  The scheme can prevent information leakage to the ``honest-but-curious cloud server'' in the condition of collaborative deep learning.
\end{itemize}

Overall, 
training neural networks especially DNNs over encrypted data is still challenging. 
Computational complexity is a major challenge. The network is slow even when trained on plaintext. Adding homomorphic encryption to a process will make it at least an order of magnitude slower. 
Since the level of the computed polynomial is proportional to the number of backpropagation steps done, the deceleration is more likely to get worse. Another challenging aspect of encryption is the lack of data scientists' ability to examine data and train models, correct mislabelled items, add functionality, and further tune the network~\cite{Dowlin2016}.

Secure multi-party computation (SMC) is the extension of encryption under the multiparty setting. 
In SMC, multiple non-colluding parties use a combination of encryption and oblivious transfer to privately finish the computation without seeing the individual components. For ML, it means to compute model updates without having access to both the data and the model.

SMC has been used for a variety of traditional ML models, 
including decision trees~\cite{Agrawal2000}, 
linear regression~\cite{Du2004, Sanil2004, Schoppmann2016, Nikolaenko2013, Jia2018a}, logistic regression~\cite{Slavkovic2007, Wu2013},
Naive Bayes classifiers~\cite{Vaidya2008}, 
and $k$-means clustering~\cite{Bunn2007, Jagannathan2005}.

In general, SMC techniques impose non-trivial computational overheads and their application to privacy-preserving neural networks especially deep learning remains a challenging task. 
SecureML~\cite{Mohassel2017} is a recent example of SMC. It uses ``two-party computations to privately train logistic regression models and neural networks''.

In summary, SMC based method can cover both data/model privacy concerns, 
at the cost of communication overhead.

\subsubsection{Obfuscation/Perturbation (Differential Private Learning)}

Obfuscation mechanisms in the context of privacy protection in ML aim at reducing the precision of the data or model.
It is can be achieved by adding noises to the model parameters or the original dataset. 
It is very popular because the DP scheme is usually implemented by obfuscation in practical applications.

The obfuscation can be applied to the model or data. 
When obfuscation mechanism is for the model, 
it has another name in the community, 
i.e., differentially private machine learning. There are some early works on traditional machine learning with differential privacy. For example, Rubinstein et al.~\cite{Rubinstein2012} proposed differentially-private support vector machine (SVM) learning mechanisms by adding noise to the output classifier and they yield close approximations to the non-private SVM. Chaudhuri et al.~\cite{Chaudhuri2011} provided the model objective perturbation to produce deferentially private empirical risk minimization (ERM) classifier. Song et al.~\cite{Song2013} derived differentially private SGD for general convex objectives and validated the effectiveness of the approach using logistic regression for classification.
One of the well-known early methods of implementing differential privacy in deep learning is~\cite{Shokri2015}. 
They trained the ML model ``in a distributed manner by updating the selected local gradients and adding noise to them within the privacy budget of each parameter''.
Based on this work, Abadi et al.~\cite{Abadi2016} introduced ``a simpler differential
private SGD (DPSGD) algorithm that ensures DP by cutting the gradients to a maximum $l_2$ norm for each layer''. And then add the noise bounded by the ``$l_2$ norm-clipping-bound''.
It was shown that ``high-quality models can be trained through privacy under a moderate privacy budget'' with the DPSGD algorithm. In DPSGD, the DP noise is added to the gradients and the whole training process involves multiple iterations. Therefore, it is important to compute the overall privacy loss of the training, i.e, privacy accounting. Although the composition theorem~\cite{Dwork2014} can be used to generate the overall privacy loss, it can be quite loose. Abadi et al.~\cite{Abadi2016} introduced a moments accountant method that can track privacy loss across multiple training iterations and generate a tighter bound. 
Another closely related notion is R\'{e}nyi differential privacy, 
which ``offers quantitatively accurate way of tracking cumulative privacy loss'' throughout a multi-round DP mechanisms~\cite{Mironov2017}.

Prior to \cite{McMahan2018}, all considered methods used ``record-level differential privacy as a framework to protect private information''. In many real-world work environments, users have multiple data sources. They may be relevant and should be protected as a whole. 
Therefore, in some cases, the DPSGD method results in a loss of privacy at a higher level (e.g., user level). McMahan et al.~\cite{McMahan2018} introduced a ``user
level differential private algorithm called the DP-FedAvg algorithm to protect all the data of a user''. Instead of limiting the ``contribution of a single record'', the DP-FedAvg algorithm limits the contribution of the user data set to the learning model. The DPSGD algorithm was ``combined with the FederatedAveraging algorithm'' from ~\cite{BrendanMcMahan2017} which uses a server that performs model averaging.

Obfuscation on training data has not been investigated extensively in the context of ML, 
because it has been deemed as similar to traditional big data privacy. 
One notable research from Zhang et al.~\cite{Zhang2018a} proposed an obfuscate function and applied it to the training data before feeding them to the model training task. 
This function adds random noise to existing samples, 
or augments the dataset with new samples. 
By doing so, sensitive information about the properties of individual samples, 
or statistical properties of a group of samples, 
is hidden. 
Meanwhile, the model trained from the obfuscated dataset can still achieve high accuracy.

Apart from the above-mentioned works, 
there are other research works in the closely relevant area,
such as tensor/matrix factorizations and functional optimization schemes. 
In more detail, 
the authors of~\cite{Imtia2018tensor,Imtiaz2018} discussed differentially private algorithms for tensor decomposition, 
in both centralized and distributed settings~\cite{Imtiaz2018}. The authors of~\cite{Friedman2016} applied a DP framework in the matrix factorization process with four different possible perturbation: 
input perturbation, 
private stochastic gradient perturbation, 
alternating least squares (ALS) with output perturbation, 
and output perturbation. 
The authors of~\cite{Zhang2012} proposed a functional mechanism framework to achieve an $\epsilon$-DP in analyses, 
which involves solving an optimization problem with a perturbed objective function. 

\subsubsection{Aggregation}

Aggregation is a technique that generally comes along with distributed/collaborated learning, 
in which multiple parties join a machine learning task while wishing to keep their respective dataset private.

Aggregation can be applied both in and after the training process.
It often works together with the encryption scheme (especially SMC) when used during the training process. 
For example, Pathak et al.~\cite{Pathak2010} proposed an aggregation scheme for independently trained classifiers. 
They average the parameters using DP and SMC. 
But they do not consider the accuracy of their approach formally.
The first part of later research \cite{Shokri2015} also focuses on aggregation. 
They reduce the communication costs and improve the model accuracy by selectively ``sharing a subset of parameters in each round of communication''.

Another popular framework using aggregation for collaborative learning is federated learning~\cite{Konecny2016, McMahan2016} introduced by Google, which has been described before. 


Compared with \cite{Shokri2015}, federated learning considers different constraints on the training dataset, 
i.e., Non-IID, unbalanced, and massively distributed, 
which is claimed to be more practical in some scenarios such as using mobile devices for the local training.

Federated learning algorithm introduces techniques for quickly and safely aggregating gradients. This scheme focuses on optimizing the communication efficiency of the aggregation process and making the protocol robust against adversaries. However, it lacks guarantees on the amount of user information leakage during training.

Bonawitz et al.~\cite{Bonawitz2017} enhance the privacy of federated learning by leveraging SMC to compute sums of model parameter updates, 
i.e., federated Learning with secure aggregation.

On the other hand, 
using aggregation schemes for privacy protection in ML after the training process, 
i.e., using ensembles of models is also reasonable. 
If an ensemble contains enough of models, and each model is trained with disjoint subsets of the training data in a distributed manner, then ``any predictions made by most of the models should not be based on any particular part of the training data''~\cite{Abadi2017}. The private aggregation of teacher ensembles (PATE) is based on this idea~\cite{Papernot2019}. In more detail, the ensemble is seen as a set of ``teachers'' for a new ``student'' model. The student is linked to the teachers only by their prediction capabilities. And the student is trained by ``querying the teachers about unlabelled examples''. The prediction result is disjoined from the training data through this process. Therefore the data privacy can be protected. The privacy budget for PATE is much lower than traditional DP ML approaches. But it may not work in many practical scenarios as it relies on an unlabelled public dataset.

Until now, 
the above works consider aggregation from the perspective of the model. 
Dwork et al.~\cite{Dwork2018} proposed a scheme that aggregates the prediction output rather than the model. 
In more details, they partition the dataset $D$ into several subsamples $D_1$, . . . ,
$D_r$ and run a non-private learning algorithm on each of those subsamples to obtain predictors $f_1$, . . . , $f_r$, 
then use a differentially private aggregation technique on values $f_1(x)$, . . . , $f_r(x)$ and output the result. 
This subsample-and-aggregate technique is easy to implement as it does not require a new learning algorithm. 
It focuses on training data privacy via private prediction.

\subsection{Summary on Private ML}
 
In this subsection, 
we sum up the key points on private ML.

\subsubsection{Discussions of attack models}

We summarize the attack models and related papers in Table~\ref{table: summary of attack model} and Table~\ref{table: papers of attack model}.

\begin{table*}[htbp]
\scriptsize
\renewcommand\arraystretch{1}
\begin{center}
\caption{\label{table: summary of attack model}Summary of Attack Models.}
\begin{tabular}{c|c|c|c|c|c}
\toprule
\hline
\multicolumn{2}{c|}{\textbf{Adversary features}} &{\textbf{ Model Extraction}} &{\textbf{Feature Estimation}}  &{\textbf{Membership Inference}} &{\textbf{Model Memorization}} \\ \hline
\multirow{2}{*}{\textbf{Knowledge}}&\textbf{Black-box}   &\checkmark & \checkmark &\checkmark&\\
  &\textbf{White-box} &  &\checkmark  &&\checkmark \\\hline
\multirow{4}{*}{\textbf{Target}}  &\textbf{Model} & \checkmark  &  &\\
  &\textbf{Data features} &  & \checkmark &&\\
  &\textbf{Exact data values} &  &   &&\checkmark\\
  &\textbf{Membership} &   &   &\checkmark& \\\hline
\multirow{4}{*}{\textbf{Scheme}}  &\textbf{Model inversion} &   & \checkmark  & &  \\
  &\textbf{Shadow training} & \checkmark &   \checkmark &\checkmark&\\
  &\textbf{Encoding} &    & &&\checkmark\\
\hline
\bottomrule
 \end{tabular}

  \end{center}
\end{table*}

\begin{table*}[htbp]
\tiny
\renewcommand\arraystretch{1}
\begin{center}
\caption{\label{table: papers of attack model}Comparisons of Attack Methods.}
\begin{tabular}{c|c|c|c|c|c|c|c}
\toprule
\hline
\multirow{2}{*}{\textbf{Attack and Threat}} &\multirow{2}{*}{\textbf{ME}} &\multirow{2}{*}{\textbf{FE}}  &\multirow{2}{*}{\textbf{MI}} &\multirow{2}{*}{\textbf{MM}} &{\textbf{Adversary's}} &{\textbf{Attack}}&{\textbf{System}}\\
& &  & & &{\textbf{Knowledge}}  &{\textbf{Method}}&{\textbf{Settings}}\\\hline
{\textbf{\cite{Tramer2016}}}   &\checkmark &  && &Black-box    &Shadow training&ML-as-a-service \\
{\textbf{\cite{Oh2019}}}   &\checkmark &  &&& Black-box     &Metamodel &Centralised\\
{\textbf{\cite{Wang2018}}}   &\checkmark && & &Black-box    &Hyperparameter-stealing&Centralised\\
{\textbf{\cite{Hua2018}}}   &\checkmark && & &Black-box    &Reverse-engineering&Centralised\\\hline
{\textbf{\cite{Fredrikson2014}}}   & & \checkmark && &White-box    &Model inversion &Centralised\\
{\textbf{\cite{Fredrikson2015}}}   & &\checkmark  && &Black-box    &Model inversion &Centralised\\
{\textbf{\cite{Ateniese2015}}}   && \checkmark & & &White-box   &Shadow training&Centralised \\
{\textbf{\cite{Hitaj2017}}}   & & \checkmark && &White-box    &GAN & Distributed \\
{\textbf{\cite{Wei2018}}}   & & \checkmark && &Black-box    &Power side-channel attack&Centralised\\
\hline
{\textbf{\cite{Shokri2017}}}   & &  &\checkmark& &Black-box    &Shadow training &Centralised\\
{\textbf{\cite{Liu2018}}}   & &  &\checkmark& &White-box    &Shadow training&Centralised\\
{\textbf{\cite{Salem2019}}}   & &  &\checkmark& &Black-box    &Unsupervised binary classification&ML-as-a-service\\
{\textbf{\cite{Hayes2019}}}   & &  &\checkmark& &White/Black-box    &GAN&Centralised \\
{\textbf{\cite{melis2019exploiting}}}   &&  &\checkmark & &White-box   &Gradient-based & Distributed \\\hline
{\textbf{\cite{Song2017}}}   & &  && \checkmark&White/Black-box    &Encoding &Centralised\\
\hline
\bottomrule
\end{tabular}
 \begin{tablenotes}
\item[\textasteriskcentered] ME: Model Extraction; FE: Feature Estimation; MI: Membership Inference; MM: Model Memorization.
\end{tablenotes}
  \end{center}
\end{table*}

The attack models listed in Section~\ref{subsec:AttackModels} are not interdependent. 
For example, 
many attacks might be launched on top of the model extraction attack,
because it converts the condition from black-box to white-box. 
Once the black-box attack is finished, 
the adversary can continue to launch the white-box attack, 
e.g., a model inversion attack followed by a model extraction attack.

\subsubsection{Attack models and protection schemes}

Table~\ref{tab:Private_ML} summarizes the private ML schemes and their effectiveness against different attacks in different situations. Generally speaking, encryption can maintain the adversary's knowledge to a black-box case, 
thus it is effective to white-box attacks like model inversion attack. 
Obfuscation~\cite{Zhang2018a} influences most attacks as it blurs the information to reduce the privacy risk at the cost of utility. 
Aggregation is mostly used in distributed systems and often comes along with the other schemes.

Another important question is the relationship of attack models, protection schemes and DP. 
Among all the mentioned attack schemes, the membership inference attack works along with DP, 
because the DP definition makes individuals indistinguishable. 
The other attack models cannot be well countered and evaluated by DP. 
For example, model inversion uses the output of a model to infer certain features of the hidden input. From a DP perspective, it does not necessarily lead to privacy breaches. For example, in a face recognition scenario, a single person is associated with an output class of the model. 
As all training images for this class include various photos of the same person, 
an adversary can orchestra a model inversion attack by creating an artificial image capturing the average information from the person's photos. 
In most of the cases, this average can be identified as that person. In summary, the average of the features produced by the model inversion can represent the entire output class at most. 
It does not construct a particular member of the training data set. 
Moreover,
given an input and a model, 
it determines whether to use that particular input to train the model.

Therefore, 
model inversion attack is even effective with DP applied collaborative learning~\cite{Shokri2015} and Federated learning~\cite{Konecny2016, McMahan2016}. Because DP is being applied to the parameters of the model, and the granularity is set at the record/instance level. However, once the model becomes accurate, it must eventually contain noise added to the learning parameters. 
Model inversion attack works as long as the model can accurately classify the class and will generate representations of that class. 
It should be noted that the DP scheme proposed in~\cite{Shokri2015} can only prevent the recovery of specific elements, that is, membership inference attack.

Overall, 
the DP criterion cannot provide comprehensive privacy evaluation in private machine learning, 
due to the complexity of the data (unstructured and multimedia data) and privacy protection target (not only membership, 
but also features of the dataset). 
Therefore, defining new privacy metrics and criteria is still an open question.

\begin{table*}[htbp]
\tiny
\renewcommand\arraystretch{1}
\begin{center}
\caption{\label{tab:Private_ML}Comparisons of Private ML Schemes.}
\begin{tabular}{c|c|c|c|c|c|c|c}
\toprule
\hline
\textbf{Private ML Schemes}&{\textbf{ME}} &{\textbf{FE}}  &{\textbf{MI}} &{\textbf{MM}} &\textbf{Categories} &\textbf{Methods}&\textbf{System Settings}   \\\hline
{\textbf{\cite{Brickell2007, Graepel2012, Bost2015, Dowlin2016, Li2017, Hesamifard2018}}}   & & \checkmark &&\checkmark &Encryption   &Homomorphic encryption (training data)&Centralised\\
{\textbf{\cite{Phong2018}}}   && \checkmark  &&\checkmark&   Encryption   &Homomorphic encryption (model)&Distributed \\
{\textbf{\cite{Mohassel2017, Du2004, Jia2018a,Agrawal2000, Wu2013,Vaidya2008,Bunn2007}}}   & &\checkmark& &\checkmark &Encryption &  SMC&Distributed \\\hline
{\textbf{\cite{Rubinstein2012}}}   & \checkmark&  &\checkmark& & Obfuscation    &DP SVM&Centralised \\
{\textbf{\cite{Chaudhuri2011}}}   & \checkmark&  &\checkmark& & Obfuscation    &DP ERM &Centralised \\
{\textbf{\cite{Song2013}}}   & \checkmark&  &\checkmark& & Obfuscation    &DP-SGD for convex objectives &Centralised \\
{\textbf{\cite{Shokri2015}}}   & \checkmark&  &\checkmark& & Obfuscation    &DPSGD&Distributed \\
{\textbf{\cite{Abadi2016}}}   & \checkmark&  &\checkmark& & Obfuscation    &DPSGD&Centralised \\
{\textbf{\cite{Mironov2017}}}   & \checkmark&  &\checkmark& & Obfuscation    &multi-round DP&Centralised \\
{\textbf{\cite{McMahan2018}}}   &\checkmark &  &\checkmark& &Obfuscation    &DP-FedAvg&Distributed \\
{\textbf{\cite{Zhang2018a}}}   && & \checkmark & &Obfuscation   &Training data obfuscation&Centralised \\\hline

{\textbf{\cite{Pathak2010}}}   & \checkmark&  &\checkmark& \checkmark& Aggregation/Obfuscation    &DP+Aggregation&Distributed \\
{\textbf{\cite{Konecny2016, McMahan2016}}}   & &  && \checkmark&Aggregation    &Federated learning &Distributed\\
{\textbf{\cite{Bonawitz2017}}}   &&  &\checkmark & \checkmark&Aggregation/Encryption  &Federated learning + SMC &Distributed \\
{\textbf{\cite{Papernot2017}}}   & &  &\checkmark& \checkmark&Aggregation    &PATE &Centralised\\
{\textbf{\cite{Dwork2018}}}   & &  &\checkmark& &Aggregation/Obfuscation    &Output aggregation + DP&Centralised \\
\hline
\bottomrule
 \end{tabular}
 \begin{tablenotes}
\item[\textasteriskcentered] ME: Model Extraction; FE: Feature Estimation; MI: Membership Inference; MM: Model Memorization.
\end{tablenotes}
  \end{center}
\end{table*}

\subsubsection{Privacy in Distributed Learning Systems}
Training ML in a distributed manner can naturally provide a certain level of privacy protection, 
as the local training data points are usually not shared among users. 
Moreover, 
different privacy protection schemes in centralised learning, 
such as encryption,
perturbation, 
can be easily extended to the distributed learning settings~\cite{wu2019value}. In this sense, private ML in distributed systems have a lot in common with that of centralised ML. But there are several special features.
\begin{itemize}
    \item Distributed ML requires some forms of data sharing among the training nodes because distributed ML is fundamentally different from stand-alone ML. 
    Such shared data, 
    albeit not raw data, 
    could take the forms of model parameters, 
    feature vectors, 
    classification results, etc., 
    and such data would still reveal users' privacy from an information theory point of view.
    Hence, 
    we need to carefully design the data sharing mechanism in distributed ML. 
    \item SMC and aggregation are quite often adopted in the distributed ML systems.
    However, 
    the above mechanisms are not adequate to protect users' privacy, 
    especially when there exist inside attackers~\cite{Hitaj2017,Nasr2019}.
\end{itemize}

\subsubsection{Backdoor attacks and privacy}
Some recent work raised the awareness of backdoor attacks against machine learning and deep learning systems, where misclassification behaviours are hidden in models and can be triggered by specific inputs.  Gu et al.~\cite{gu2017badnets} introduced BadNets that builds a backdoor in DNN models by injecting a square-like trigger with a fixed location to some training data with a target label. Ahmed et al.~\cite{Salem2020} extended this work by using dynamic trigger patterns and locations. Liu et al.~\cite{Liu2018c} proposed a backdoor attack called the Trojan attack, which reverse-engineers the target model to synthesize training data so that it does not require access to the original training set. Yao et al.~\cite{Yao2019} proposed a latent backdoor attack method in which they embed the backdoors in teacher models to survive the transfer learning process. In general, current backdoor attacks are mostly considered to be security risks, e.g., it may cause various severe consequences in critical ML applications like autonomous driving. But we can also expect potential privacy risks in the future, for example, backdoor attacks against authentication systems that might enable an adversary to access sensitive information. 

\section{Machine Learning Aided Privacy Protection}
\label{section:MLaidedaprotection}

In this section, we will focus on the case that ML is used to help privacy protection. We will first discuss traditional data privacy risks and threats. These threats have existed for a while, but the newly emerging ML gives us new tools to combat them.

\subsection{Attack and Threat Models}

Along with the proliferation of the mobile network, 
people spend more and more time on the Internet, 
using web-based applications, 
mobile applications and social networks. 
These all pose privacy risks. 
For example, 
online photo sharing has become more popular than any time before. Users are increasingly sharing their images on various social media, such as Facebook, Google+ and Flickr. 
Shared images can reveal sensitive information about people and their surroundings~\cite{Squicciarini2014,Zerr2012}.
Consider a person sharing a photo of a family gathering. 
Not only this photo can expose the people who may or may not wanted to be in the picture, 
but it can also reveal sensitive information about the family such as religious beliefs, traditions, and food habits. 
Therefore, 
sharing photos online can severely violate privacy and disclose sensitive information~\cite{Fox2015}.

\begin{figure}[t!]
\centering
   \subfigure[Re-identification Attack]{\label{fig:Re-identification Attack}
\includegraphics[scale=0.35]{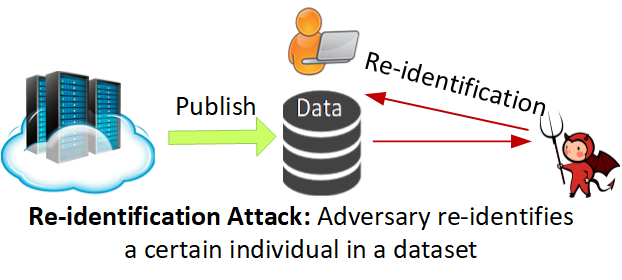}}
\hspace{0.2cm}
\subfigure[Inference Attack]{ \label{fig:inference attack}
\includegraphics[scale=0.35]{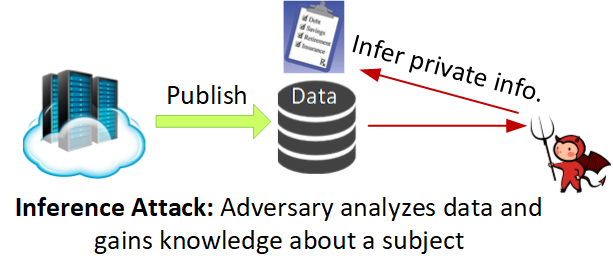}}
\hspace{0.2cm}
\subfigure[Linkage Attack]{\label{fig:linkage attack}
\includegraphics[scale=0.35]{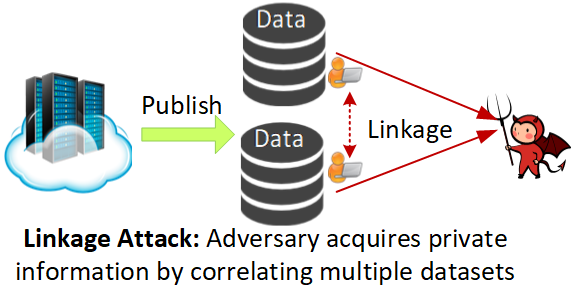}}
\caption{Different privacy attack and threat models.}
\label{fig:attack_models_ML_aided1}
\end{figure}

Major traditional privacy attacks include identification attacks,
inference attacks, 
and linkage attacks, 
as shown in Fig.~\ref{fig:attack_models_ML_aided1}.

\begin{enumerate}
  \item Identification attack: 
  Identification attack identifies a user's name or identity-based on some public dataset~\cite{Li2018}. 
  It is also called re-identification~\cite{Henriksen-Bulmer2016, Hussien2013} when anonymisation is reversed.
  Such kind of attack is illustrated in Fig.~\ref{fig:Re-identification Attack}.
  \item Inference attack: 
  This type of attack aims at ``analyzing data in order to illegitimately gain knowledge about a subject''~\cite{Krumm2007}.
  Such an attack is illustrated in Fig.~\ref{fig:inference attack}.
  \item Linkage attack: 
  The adversary aims to achieve a target's information by correlating multiple data sources. 
  For example, 
  Narayanan et al.~\cite{Narayanan2006} showed that an adversary ``can identify a subscriber's record in the Netflix Prize dataset'', linking it to an Internet Movie Database.
  Such an attack is illustrated in Fig.~\ref{fig:linkage attack}.
\end{enumerate}

\subsection{Machine Learning Aided Privacy Protection Schemes}
Many privacy protection schemes have been introduced. Obfuscation/perturbation~\cite{Dwork2008, Shokri2015}, 
anonymization~\cite{Aggarwal2005, Aggarwal2005a},
reducing information sharing~\cite{Shokri2014, Liu2016}, 
and cryptographic mechanisms~\cite{Pinkas2002, Gai2016} are the major technologies.

However, the traditional privacy protection schemes focus on structured data,
such as an entry in the databases~\cite{Wang2010}. 
With the introduction of new applications such as Internet of Things (IoT) and vehicular networks, 
both the volume and the complexity of the data is increasing.
Traditional protection schemes cannot handle all cases and it also becomes more difficult for both common users and even data curators to understand the risk, 
select correct schemes and manage their privacy.

\begin{figure}[t!]
\centering
   \subfigure[Privacy risk assessment and prediction]{\label{fig:Privacy risk assessment and prediction}
\includegraphics[scale=0.45]{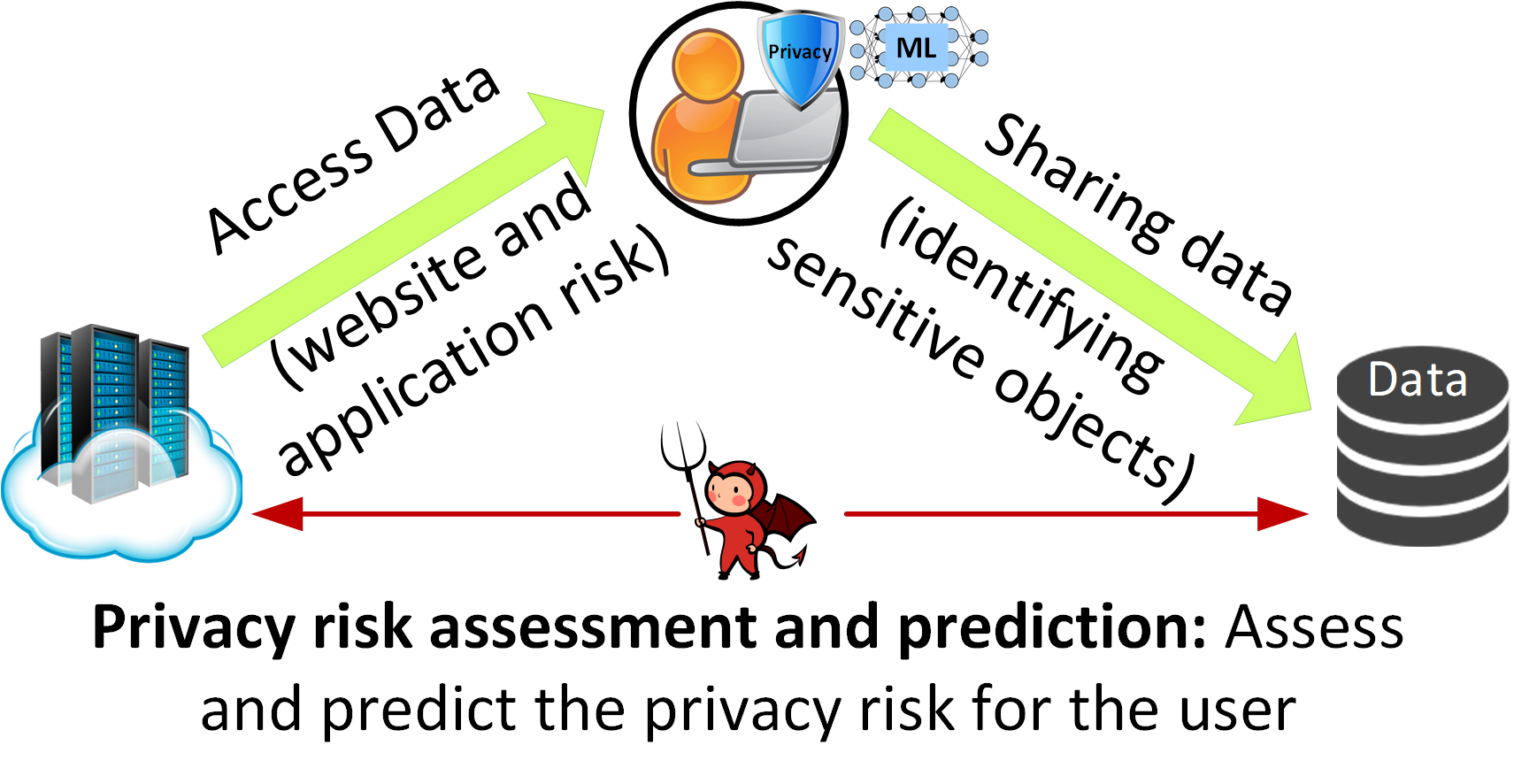}}
\subfigure[Personal privacy management assistant]{ \label{fig:Personal privacy management assistan}
\includegraphics[scale=0.45]{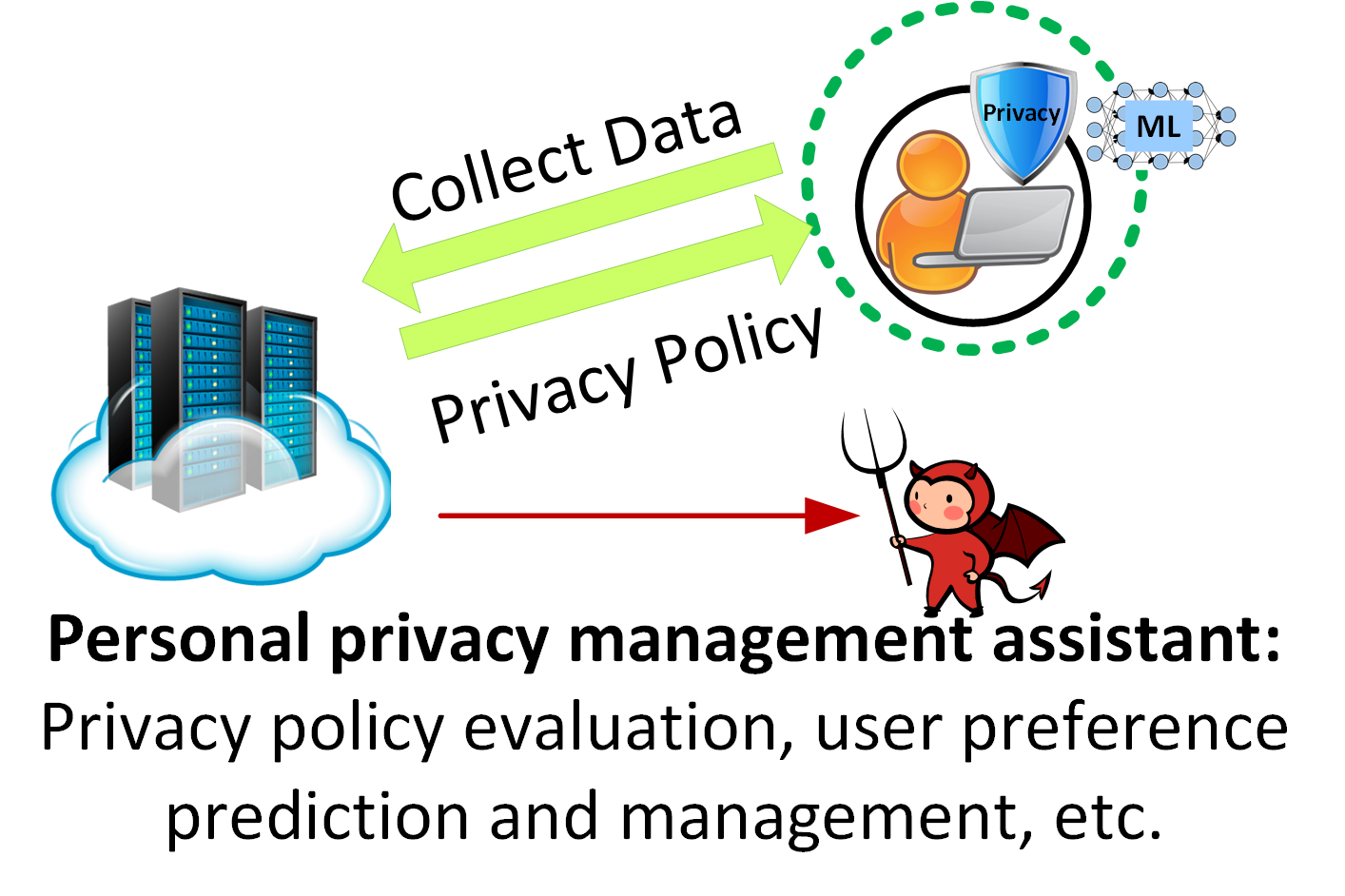}}
\subfigure[Private  data  release]{\label{fig:Private  data  release}
\includegraphics[scale=0.45]{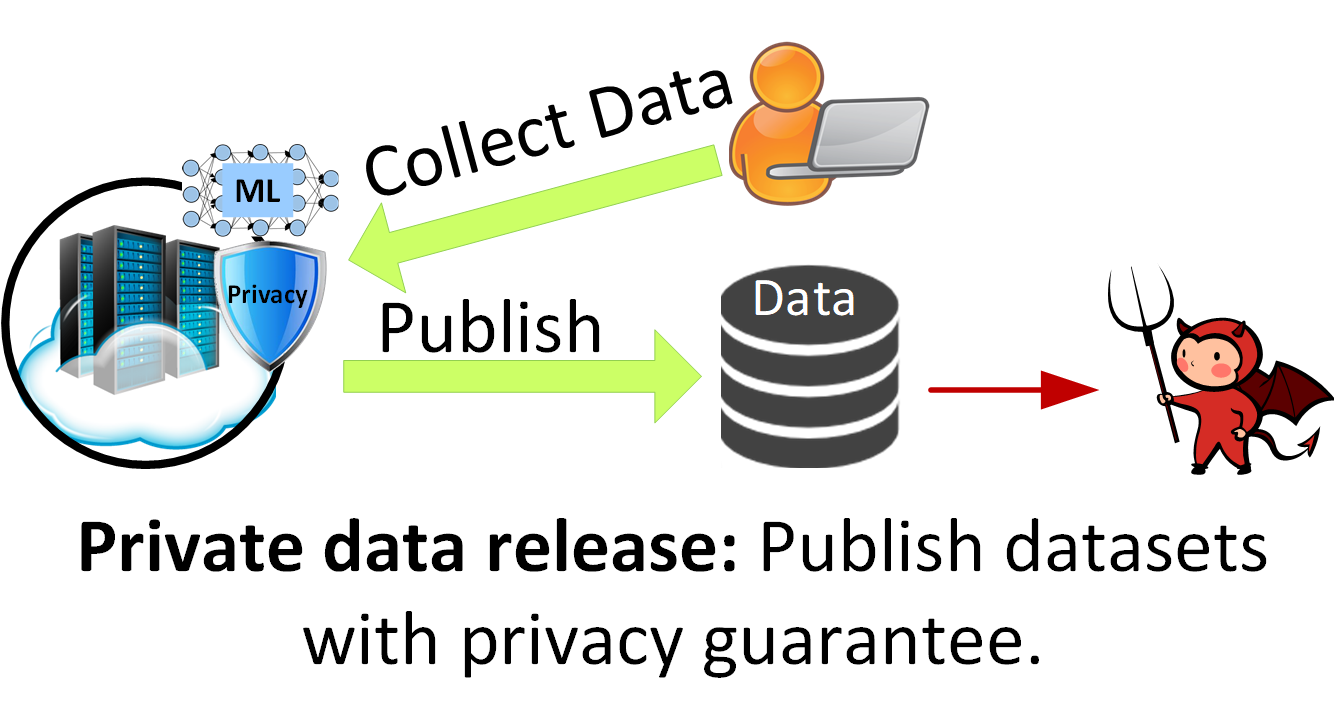}}
\caption{ML-aided privacy protection schemes.}\label{fig:attack_models_ML_aided}
\end{figure}

Under these circumstances, 
ML has been introduced to enhance privacy protection during the past few years. 
The efforts including research in several aspects.
\begin{itemize}
  \item Privacy risk assessment and prediction: 
  Assess and predict the privacy risk for the user during the processes of ``access'' and ``sharing''. 
  As shown in Fig.~\ref{fig:Privacy risk assessment and prediction},
  ML is used to evaluate both the input and output data streams to find the risk and then privacy protection schemes can be deployed accordingly.
  \item Personal privacy management assistant: 
  This includes privacy policy evaluation, 
  user preference prediction and management, 
  as shown in Fig.~\ref{fig:Personal privacy management assistan}.
  \item Private data release: 
  Publish datasets with privacy guarantee. 
  The schemes are generally adopted by data curators rather than an individual user, 
  as shown in Fig.~\ref{fig:Private  data  release}.
\end{itemize}

\subsubsection{Privacy Risk Assessment and Prediction}
\label{subsec: risk assessment}

The privacy risk exists either when the user is just accessing the application (passively collected information by malicious attackers) or sharing on social networks (actively sharing information). 
In both cases, 
ML can help to prevent the loss of sensitive information.
An illustration of such a defence mechanism is shown in Fig.~\ref{fig:Privacy risk assessment and prediction}.

\textbf{\emph{Website and application privacy risk prediction}:} ML can make browsing the websites safer. 
The proposed browser extension in~\cite{Sebastiani2002} collects information about websites that users visit and provides feedback to users based on ML to let them know the privacy quality of the site. Manek et al.~\cite{Manek2016} proposed a method based on a Bayesian classifier to detect and identify websites that can be malicious or threatening to the privacy of users. 
The proposed approach analyzes online reviews written for websites to decide whether they are reliable or not.

The work in~\cite{Fu2019} uses an SVM classifier to rate the privacy risks of applications. 
The results indicate that privacy risks can be identified with over $90\%$ accuracy. 
Understanding the privacy risks of mobile phone applications with the aid of ML have been considered in~\cite{Avdiienko2015,Gorla2014}.

\textbf{\emph{Identifying sensitive information when sharing}:}
Identifying sensitive information in multimedia data has been difficult in the past. 
With the help of the state-of-the-art ML techniques, 
users can prevent loss of their personal information while sharing their photos on social media.

Squicciarini et al.~\cite{Squicciarini2017} considered visual-content features and images' metadata to develop and contrast several learning models. 
The ML models can classify the photos and evaluate the degree of sensitivity so as to make the decision based on past decisions of the users. Yu et al. \cite{Yu2017} proposed a tool called ``iPrivacy (image privacy)'' to reduce the burden of specifying privacy setting by users when they are sharing photos online. 
iPrivcy utilizes ML to automate the process. 
It finds privacy-sensitive objects from images and classifies them according to their privacy sensitivity. 
Based on the classification, 
iPrivacy notifies the users if there are objects, 
which should be suppressed/masked due to privacy concerns before sharing. 
Moreover, 
iPrivacy provides privacy settings recommendation based on user preferences and shared images. Orekondy et al.~\cite{Orekondy2018} proposed the first large-scale private images dataset, with pixel and instance level annotations. And they proposed the first model to automatically redaction various private information. Hasan et al.~\cite{Hasan2020} proposed a method to automatically identify bystanders ``solely based on the visual information present in an image''.

\subsubsection{Personal Privacy Management Assistant}

As the user connectivity increases and web applications become ubiquitous, 
the responsibility of privacy management transfers more and more to individuals. 
Unfortunately, 
given the complexity of the environments and the lack of awareness about privacy attacks by adversaries, 
it is improbable that the users can manage and fine-tune their privacy preferences correctly~\cite{McCrae2002}.
Therefore, 
there is an immediate need to develop automated privacy management systems to help users in protecting their privacy.
An illustration of such a defence mechanism is shown in Fig.~\ref{fig:Personal privacy management assistan}.

The authors of~\cite{Acquisti2015} indicate that users continuously modify their privacy requirements to reach their expected level of privacy, 
and also, appropriately change their privacy preferences. 
Moreover, 
mobile and web applications are attempting to customize their services according to individual preferences to grant personalized experience to customers. 
Such a customized service results in potential risk for the users~\cite{Pearson2010}. 
This evidence points to the fact that it is crucial to develop assistants to help users with the management of their privacy configurations. 
ML can be an invaluable asset in this regard. 
For example, 
it can help users to manage their privacy configurations and reduce the burden of time and human resources required to ensure the preservation of privacy.

We have divided the applications of ML for privacy management in two broad categories: 
(i) privacy policy evaluation, 
and (ii) user preference prediction and management.

\textbf{\emph{Privacy policy evaluation}:} Users are usually prompted to agree with the provider's privacy policies when almost using any software and web applications.
 Privacy policies provide complete information on the collection, storage, and sharing of personal data. 
 Therefore, 
 they are critical to the privacy of users. 
 Unfortunately, 
 most of this information is written using technical jargon and challenging to read terms. 
 Hence, 
 most of the readers prefer to accept the policy unconditionally without thoroughly realizing the consequences~\cite{Costante2011}. 
 To help users with the decision making, Costante et al.~\cite{Costante2012} developed a system to evaluate the completeness of privacy policies based on preferences of the users. 
 The system uses natural language processing to analyze and verify the existence of the privacy measures that users specify, 
 and also, assess the level of completeness. Nugent et al.~\cite{Nugent2018} graded the privacy policies that the users encounter based on factors such as security, cookies, and purpose which helps users to check the results and identify if their desired privacy requirements are satisfied. Tesfay et al.~\cite{Tesfay2018} proposed an ML approach to ``summarize the long privacy policies'' into a short paragraph so that it is readable and understandable for users. Shayegh et al.~\cite{Shayegh2017} considered methods to improve the privacy notices given to users in IoT networks. 
 With the aid of ML, 
 the authors extract notice and choice statements from the privacy policies for IoT devices, 
 so as to help users to better understand the implications of privacy notices. Lebanoff et al.~\cite{Lebanoff2018} investigated automatic detection of vague contents on privacy policies and used GANs to characterize the vagueness of sentences.

\textbf{\emph{User Privacy Preference Prediction and Management}:} Another difficulty in user privacy protection is caused by the fact that each user has a different privacy sensitivity and preference.
Nowadays, 
applications often provide many functionalities with different levels of privacy guarantees. 
While installing the applications, 
users are usually prompted for permissions to access resources that have an impact on their privacy. 
It is important that the users can well coordinate their own privacy preference with the actual privacy risk.

ML techniques are implemented to predict user privacy preferences and help decision making. 
It was initially proved feasible as some early studies found that user privacy preferences are related to some statistical and environmental parameters. 
For example, 
the quantitative research in~\cite{Wijesekera2017} uncovered that a significant number of users would rather prevent at least one permission request involved in the study. 
Also, several works have shown that the context of the applications is highly related to user privacy preferences~\cite{Olejnik2017,Wijesekera2015}. Lee et al.~\cite{Lee2017} surveyed 172 participants and uncovered contextual factors that violate the privacy of users in IoT.

Based on the contextual factors and features, 
ML models can be developed to predict user privacy preferences and take privacy management decision. Mehrpouyan et al.~\cite{Mehrpouyan2017} used openness, conscientiousness, neuroticism, extroversion, and agreeableness as inputs to ML models to predict desired users' preferences. Das et al.~\cite{Das2018} generated ML models of people's privacy preferences and expectations.

Wijesekera et al.~\cite{Wijesekera2015} proposed a run-time permission system to infer privacy requirements of users automatically. 
The proposed system grants the resource allocation permission based on the type of the application requesting the permissions, 
the request time, 
and in what circumstances it is requested. Liu et al.~\cite{Liu2016} investigated ML to enhance privacy decision-making experience. 
The results show that providing users with ``recommendations based on clusters of like-minded users and using predictive models of people's privacy preferences work to the users' satisfaction''. Wijesekera et al.~\cite{Wijesekera2017,Wijesekera2018} built a classifier to work as a middle-man and make privacy decisions on behalf of users. 
The classifier adjusts and preserves privacy by changes that happen in the context predicated on the past behaviors of the users.

Orekondy et al.~\cite{Orekondy2017} proposed a method named ``Visual Privacy Advisor'' that ``extends this concept to image'' contents.
They classify ``personal information in images into 68 attributes and train models that directly predict such information from images''. A user study has been done to understand the privacy
preferences with respect to these attributes. They also proposed models that ``predict user specific privacy score from images''. Yuan et al.~\cite{Yuan2017} presented an ML approach to decide whether to share a picture with a specific requester at a particular context, 
and if yes, at which granularity.

\subsubsection{Private Data Release}

Database release is currently an important process in data analytic applications. 
Different entities generate different types of data, 
e.g., health data from medical centers. 
Then, 
such data will be transmitted to data custodians such as government agencies.
Then, 
the data custodian maintains a platform that organizes, stores and provides data access to data consumers, 
such as other government departments, individuals, analysts, etc.
Privacy preservation processing is highly required when the data custodians release the data. 
An illustration of such a defence mechanism is shown in Fig.~\ref{fig:Private  data  release}.

A frequently used traditional private data release mechanism is obfuscation by adding noise to the original dataset.
Whereas the ML techniques provide a new solution to this problem,
i.e., using a generative neural network (GNN) or generative adversarial network (GAN)~\cite{Goodfellow2014} to generate synthetic dataset~\cite{Acs2019}.

Although the technique of GNN itself has existed for a while, using it for private data release has just been linked to privacy preservation very recently. Denton et al.~\cite{Denton2015} used the GAN framework in the context of image processing to generate natural synthetic images. Gregor et al.~\cite{Gregor2015} introduced a model called ``Deep Recurrent Attentive Writer (DRAW)'' to create synthetic images. 
The principal idea of the approach is to use two recurrent neural networks as encoder and decoder trained end-to-end with SGD. Vinyals et al.~\cite{Vinyals2015} proposed a generative model predicated on recurrent neural network architecture. 
The approach combines the natural processing ML tools with computer vision for the generation of natural scenes. 
Using generative models has also been considered for the generation of audios. Oord et al.~\cite{Oord2016} introduced a DNN model to produce raw audios and applied the approach to ``text-to-speech and validated by human listeners for natural sounding''. 
A modified version of the proposed model is used for singing synthesis in~\cite{Blaauw2017}. Kulkarni et al.~\cite{Kulkarni2018} created spatiotemporal trajectories in large scale by training the models based on realistic data, 
and then, creating synthetic data using the trained models. 
The authors investigate the utility-privacy trade-off of the approach by experiments. Ouyang et al.~\cite{Ouyang2018} proposed a non-sequential non-parametric generative model for spatiotemporal trajectories. 
The authors generate ``synthetic data by training a generative adversarial neural network, 
which can learn geographic patterns''. Liu et al.~\cite{Liu2018b} aim at the addition of geo-privacy protection layer for publication of spatiotemporal datasets based on synthetic trajectory generation. Choi et al.~\cite{Choi2017} proposed an approach for the generation of synthetic patient records based on GANs and autoencoders. 
In this work, 
the performance of the proposed generative model is examined by comparing the generated synthetic patient records with the real data. Cheung et al.~\cite{Cheung2018} used GNNs for the transformation of sensitive images so that they can preserve privacy of individuals. 
The authors focus on the generation of synthetic facial images and how they can be used for classification of actual images. Zhang et al.~\cite{Zhang2018b} proposed a novel approach based on GNNs to increase privacy of users while releasing semantic rich data such as text, image, and video. Triastcyn et al.~\cite{Triastcyn2019} used GAN to generating artificial data that retain statistical properties of the real data while reducing the risk of information disclosure. Sun et al.~\cite{sun2018natural} proposed GAN-based head inpainting obfuscation technique to preserve the identity of users when sharing their photos online. Huai et al.~\cite{Huai2019} considered the differentially private release of crowdsourcing data. They proposed the PrisCrowd approach ``in which the data collector learns about underlying patterns of the data and then samples a set of candidate synthetic data from the learned density. The synthetic data are subjected to a privacy test and the ones that pass will be released''.  

\begin{figure}[th!]
\centering
\includegraphics[scale=0.38]{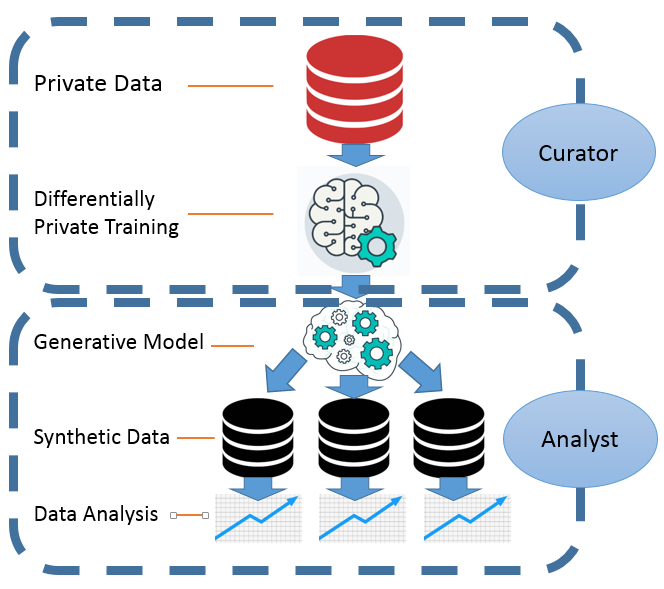}
\centering
\caption{Privacy preserving framework based on generative model approach.}
\label{g1}
\end{figure}

Overall, 
the latest deep learning techniques show the ability to synthesize fake dataset that is statistically similar to the original one. 
This technique can be used for private data release. 
Fig.~\ref{g1} presents the generative model framework used for privacy preservation of rich semantic data. 
The process can best be explained by an example. 
Consider a clinical data sharing scenario, 
in which the data curator instead of directly releasing the data,
trains a deep generative model using the original data in a differentially private manner, 
and then publishes synthetic dataset generated by the model. 
In a more general case, 
the data curator may publish the deep generative model from which ``an unlimited amount of synthetic data for arbitrary analysis tasks'' can be produced~\cite{Xu2018}. 
The use of generative models can significantly increase the privacy of users as the training process of the models can be conducted based on synthetic data instead of the real data belonging to individuals. 
Meanwhile, the utility of the dataset can be guaranteed as the statistical similarity of models trained based on synthetic data and realist data has been shown repeatedly in the literature. 
For example, Park et al.~\cite{Park2018} proved the statistical similarity of the generated synthetic tabular data and original data. Xu et al.~\cite{Xu2018} developed training deep neural networks for the generation of synthetic data that closely resemble the actual medical records of patients.

Although research in GNNs for privacy preservation is in its initial stages, 
the outlook of the approach is promising. 
Generation of synthetic data is particularly crucial as traditional methods such as anonymity and obfuscation are ineffective for privacy preservation of semantic-rich data. 
Moreover, 
this approach is not associated with the drawbacks of other traditional anonymization approaches such as having background knowledge or linking the data to other sources.

\subsection{Summary on ML-aided Privacy Protection}
The three different groups of ML aided privacy protection schemes introduced in Subsection 4.2 work in various stages of privacy protection. Privacy risk assessment and prediction is a pre-process before privacy protection, that identifies what do we need to protect. Personal privacy management assistants help to improve access control over sensitive information. Private data release can be applied directly to the data. These protection schemes do not have a one-on-one relationship with the attack models listed in Subsection 4.1. They can be effective against multiple attack models and will work best if combined correctly in specific scenarios.

The two main types of ML models used for privacy protection are classification and object detection. 
Classification is used for privacy risk prediction and assessment.
Object detection is used for identifying sensitive information. Additionally, 
schemes discussed in~\ref{subsec: risk assessment} do not directly provide privacy protection. 
They are currently playing a supporting role, 
and other subsequent privacy protection schemes are still needed.

GNN opens a new direction for privacy protection research,
especially for unstructured data such as image and video. 
But it is still challenging, 
as there are no unified metrics for privacy measurements in those complicated cases.

\section{Machine Learning-based Privacy Attacks and Corresponding Protection Schemes}
\label{Sec:PrivacyAgainstML}

Besides serving as a privacy protection tool,
ML can also be used as an attack tool. 
It urges us to revisit the definition and scope of privacy. 
In particular, 
the emerging deep learning technique can ``automatically collect and process millions of photos or videos to extract private/sensitive information from social networks''~\cite{Liu2019a}. 
Traditional privacy-preserving methods are over-powered when combating deep learning tools. 
It is time to seriously discuss new threats and corresponding solutions.

\subsection{Attack and Threat Models}

The riskiest personal information leakage source is the social network. 
While there are a variety of social network platforms enriching people's interactivity and relationship, 
the shared posts including check-ins, activities, thoughts (tweets, status updates, etc.), pictures, videos often come along with sensitive information. 
The information poses high privacy risks and they are likely to hand over their privacy unintentionally. 
A growing number of companies and start-ups specialize in analyzing shared pictures on social media to exploit them for commercial purposes or selling them to other companies. 
Therefore, 
the most advanced DNNs have been used to launch privacy attacks.

For example, 
the adversary can use geo-location information to initiate a localized attack that focuses on finding the position and time information of the person.
Gu et al. and Mahmud et al.~\cite{Gu2016, Mahmud2014} showed a dangerous attack that is designed to ``find important locations such as homes and workplaces''. 
There have been some researches discussing the home location identification problem, 
either based on the ``content of the posts''~\cite{Cheng2010}, 
or the ``geo-tags in the check-ins''~\cite{Cho2011}. 
And ``the research shows that the identification accuracy might be over 90$\%$ in many cases''~\cite{Liu2017}.

 Besides the simple location information, 
 multimedia data poses more risk under the attack of ML tools.
 Companies apply advanced DNNs to cluster photos or infer preference of users to facilitate marketers to send targeted ads~\cite{Meng2014, Reznichenko2014}. 
 DNNs are considered one of the most practical tools in ML as they take advantage of efficient training algorithms and large datasets which enables them to outperform other existing ML techniques. 
 The power of such ML tools has become a problem itself that may compromise the privacy of photos once they are shared on social media and a challenging problem that needs to be addressed.

The privacy of sensitive data, 
photos and videos become more crucial in IoT networks, 
as users might not even be aware of their information such as pictures and videos being recorded. 
For instance, 
areas controlled under surveillance cameras can severely compromise user privacy as people lose control of how their photos and videos are being captured and managed. 
It is likely that the surveillance system applies techniques such as face recognition and detection to identify the users without their permission. Pew Internet survey in 2014 reported that over 91 percent of participants ``strongly agree'' or ``agree'' that ``they have lost their control over how their personal information is being collected and used by companies''~\cite{Center2014}.

\begin{figure}[t!]
\centering
\subfigure[Identification attack]{\label{fig:Identification attack}
\includegraphics[scale=0.5]{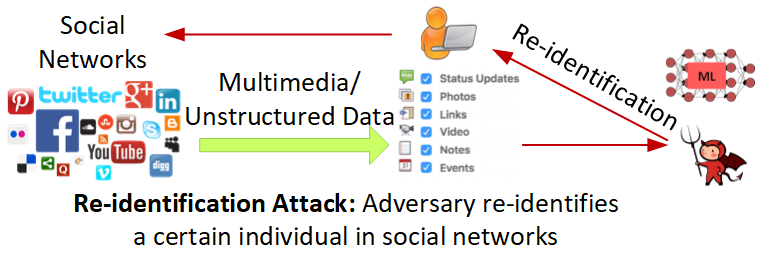}}
\hspace{0.2cm}
\subfigure[Inference attack]{ \label{fig:Inference attack}
\includegraphics[scale=0.5]{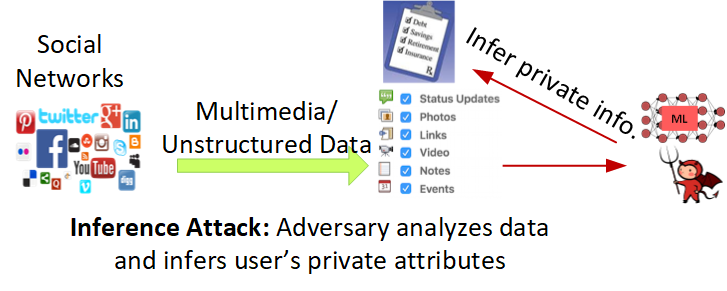}}
\caption{Different privacy attack and threat models when ML is used as the attack tool.}\label{fig:attack_models_ML}
\end{figure}

Major ML attack models include re-identification attacks and inference attacks, as shown in Fig.~\ref{fig:attack_models_ML}. 
These attack models are different from those described in Section~\ref{section:MLaidedaprotection} in the sense that ML is used as an attack tool here.

\begin{itemize}
    \item The re-identification attack can be launched by face recognition techniques. 
The recent advance in DNN makes it more harmful from two aspects. First, 
the process becomes automatic with high accuracy~\cite{Wilber2016, Sun2018, JoonOh2015}. 
Second, 
traditional protection schemes such as obfuscation no longer work effectively~\cite{Oh2016, McPherson2016}.
An illustration of the re-identification attack can be found in Fig.~\ref{fig:Identification attack}.
\item Inference attack has also become more powerful when equipped with ML. 
ML classifiers can be used to infer a target user's private information (e.g., location, occupation, hobby, political view) from its public data (e.g., twitters, movie rating scores)~\cite{Cheng2010, Cho2011}. 
Moreover, a series of research work have demonstrated how the advanced artificial neural networks can be used as an adversarial tool to detect sensitive information in images, including people's age~\cite{Jahanbekam2010}, relationship~\cite{Sun2017} and vehicle license plates~\cite{Zhou2012} from ordinary or even obfuscated images. 
An illustration of the inference attack can be found in Fig.~\ref{fig:Inference attack}.
Therefore, 
it is quite urgent to accelerate the research on privacy protection schemes against ML aided attacks.
\end{itemize}

\subsection{Protection Schemes Against ML-based Attacks}

There has been some preliminary research in this area. 
For privacy protection against traditional ML attack, Liu et al.~\cite{Liu2017} designed community-based information sharing scheme that changes the overall spatial and temporal features so that the clustering-based privacy attack~\cite{Liu2014} no longer works.

The problem becomes more challenging when deep learning is involved. 
The solutions may come from a better understanding of deep learning itself. 
Some researchers recently found that there are limitations to deep learning. Specifically, ``it is proved to be vulnerable to some well-designed inputs termed \emph{adversarial examples}''~\cite{Sharif2016,Fawzi2015}. 
Szegedy et al.~\cite{Szegedy2014} first discovered that the superposition of ``imperceptible noise onto the original image'' would mislead DNNs to the wrong classification. 
Then, Goodfellow et al.~\cite{Goodfellow2014} proposed the ``fast gradient sign method (FGSM) that can be used to generate this type of adversarial examples''. 
Other algorithms to generate such noise can be found in~\cite{Moosavi-Dezfooli2016, Kurakin2019, Rozsa2016}.

According to~\cite{Papernot2016a}, 
the primary reason for why neural networks are vulnerable to adversarial examples is the linear nature of the neural networks. 
The authors formalize the space of adversaries against DNNs, 
which are mostly originated from ML techniques itself. 
In simple words, 
ML is used as a tool to breach the ML classifiers. 
Kurakin et al.~\cite{Kurakin2019} focused on adversarial training and how they can be scaled to large datasets. 
Sharif et al.~\cite{Sharif2016} proposed an algorithm for manufacturing adversarial examples based on ML to disable DNN detection systems from finding objects in shared photos. 
Additionally, 
a significant point about adversarial examples is its transferability property~\cite{Goodfellow2014}. 
It means that if they are able to fool one model, 
they are often likely to mislead another model with a different set of parameters and architecture~\cite{Szegedy2014}. 
This is even true if the other model is trained on a different training set or model~\cite{Papernot2016}. 
This leads to the idea of universal perturbation~\cite{Moosavi-Dezfooli2017, Poursaeed2018}. 
It is even possible to ``generate adversarial examples that fool both human and computer alike''. 
Elsayed et al.~\cite{Elsayed2018} exploited ML to construct adversarial examples that transfer from models created based on computer vision to the human visual system. 
The authors generated adversarial examples without utilizing the parameters of the model's architecture, 
and then mimic the visual processing of humans using ML.

Enlighted by the idea of adversarial examples, 
researchers started to focus on the generation of adversarial examples based on ML to improve the privacy of users against attacks mostly based on DNNs. 
Liu et al.~\cite{Liu2017a} proposed an algorithm that is against automatic detection using adversarial examples based on the ``Faster RCNN framework''.
Jia et al.~\cite{Jia2018} proposed a two-phase framework called AttriGuard to defend against attribute inference attacks launched by a classifier. Liu et al.~\cite{Liu2019a} investigated schemes for using adversarial examples in ML systems so that they cannot identify the sensitive information from images. Oh et al.~\cite{oh2017adversarial} set up a game-theoretical framework and studied the effectiveness of adversarial image perturbations for privacy protection. Li et al.~\cite{Li2019} proposed to use adversarial perturbation for face de-identification. Friedrich et al.~\cite{Friedrich2020} proposed a privacy-preserving shareable representation of medical texts for a de-identification classifier. 

\subsection{Summary on Privacy Protection against ML}
Previously, 
the common understanding of privacy protection is to prevent human adversaries from knowing some sensitive information about people. 
For example, 
obfuscating faces in images is a well-researched topic. 
However, 
the situation has dramatically changed recently. 
First, the growth of data volume has reached a point where it is physically impossible for anyone to browse everything with their eyes. Second, as a result, people increasingly rely on machines with advanced algorithms to extract relevant information of interest. Third, the booming of ML open source community makes ML tools easy to be obtained by anyone. This brings up a new problem, that is, it is now possible to automatically process data to infer sensitive user information, such as personal identity, social relationships, location, and context.
Indeed, 
ML has recently been used by malicious parties as an efficient tool to launch new types of privacy attacks, 
especially for social media data. 
Therefore, 
we would expect that privacy protection against machines is as important as privacy protection against humans.

ML-based privacy attacks are more challenging to defend against, 
due to three main reasons. 
First, 
the average user is not aware of the capability of state-of-the-art ML methods in extracting personal information. 
Second, privacy in some contexts such as multimedia data is not obvious. Third, privacy threats also arise from organizations and government sectors that collect and analyze data on a large scale. Therefore, 
we need to prevent ML algorithms from automatically mining private information, 
either intentionally or unintentionally.

In summary, privacy protection against the fast-evolving ML techniques is the most challenging task among all three categories we discussed in the paper.   
The methodology is to exploit the weakness and limitations of ML methods. 
Although there have been some initial solutions to this problem using adversarial machine learning, there are still many research problems that require further investigations.


\section{Outlook and Future Directions}
\label{Sec:outlook}
Significant previous work focuses on making ML algorithms differentially private to preserve the privacy of training sets. However, we should be aware that machine learning, as a whole, also provide potent tools for privacy research (not just for the training datasets), both from attack and defense perspectives.

\subsection{Perturbation in Deep Learning}
The goal of perturbation in deep learning is to train a model while ensuring DP concerning information about individual training examples. Theoretically, the noise can be added to either the input data, the model parameters (through gradient updates), or the model output. In practical, the majority of work proposed to inject noise into gradients. The main disadvantage of this group of methods is that amount of injected noise is dependent on the number of training epochs, and it potentially can accumulate too much noise due to the significant number of parameters. 

Directly adding noise to input data is an option, but it is similar to a typical big data privacy problem and does not closely related to deep learning. Output perturbation and objective perturbation seem to be reasonable directions in the future. 

Output perturbation adds noise to the output of the ML system, e.g., the logits at the prediction stage. This method is fast and easy to implement. However, it can suffer from degradation from an attack of repeated querying by an adversarial. Therefore, it is important to restrict the number of queries~\cite{Rahimian2020}. One potential solution is to use output perturbation in certain intermediate outputs, such as the teacher voting output in PATE frame work~\cite{Papernot2019}. 

Objective perturbation is one of the most effective methods for differential privacy ML. This technique adds a random linear term to the objective function. Objective perturbation has been extensively studied in convex optimization. Recently, Iyengar et al.~\cite{Iyengar2019} has provided a practical algorithm for differentially private convex optimization, which is a big step towards practical deployment of this technique. Moreover, Neel et al. ~\cite{Neel2019}  has extended this approach to non-convex optimization problems. Despite the success in traditional ML, applying objective perturbation to deep neural network is still challenges due to several obstacles: 1) the sensitivity calculation is difficult because the objective functions of deep learning models are mostly non-convex and do not have closed-form expressions; 2) the privacy guarantee is implicitly based on the rank-one assumption on the Hessian of the loss, which is difficult to verify; 3) the privacy guarantee holds only at the exact minima (at least the approximate minima as proposed in~\cite{Iyengar2019}) of the optimization problem, which is hard to be guaranteed in practical deep learning systems. One possible solution is to use a convex approximation of the loss function~\cite{Phan2017}. However, the approximation error might outweigh the reduced perturbation due to smaller sensitivity. It is  expected to see more effective methods following this path.

Moreover, instead of perturbing the final output, it is also possible to add noise to the middle layers of the neural networks. Lecuyer et al.~\cite{lecuyer2019certified} proposed the PixelDP framework that includes a DP noise layer in the DNN. Although the purpose of PixelDP is ``to increase robustness to adversarial examples'', the idea can be further investigated to serve for privacy preservation. For example, PixelDP scheme enforces that the output prediction function is DP provided the input changes on a small number of pixels (when the input is an image). Potential extensions to PixelDP include: 1) enforcing DP for given different input samples so that it can provide privacy preservation for the training set against membership inference attacks; 2) adding DP noise to the hidden layer of an autoencoder. With the post-processing property of DP, the output of the autoencoder remains to be DP as well. This idea is briefly mentioned in~\cite{lecuyer2019certified}. But we can further explore it in different applications. For instance, we can protect a social network image by generating a perturbed version using this autoencoder with a DP guarantee.

\subsection{Defending ML-based Privacy Attack: Adversarial Examples}
As we have discussed in Section \ref{Sec:PrivacyAgainstML}, when ML is used as a privacy attack method, adversarial examples become a powerful way of privacy protection. Despite the preliminaries work on this topic, there are several issues that need to be solved:
\begin{itemize}
    \item Adversarial example generation methods fall into two categories of attack scenarios: white-box and black-box. The research of using an adversarial example for privacy protection usually assumes that the deep learning model is known, using the white-box setting. In practice, the black-box scenario seems to be a more realistic assumption, e.g., the latest black-box adversarial generation methods such as ZOO~\cite{chen2017zoo}, $\mathcal{N}$ attack~\cite{li2019nattack} and AdvFlow~\cite{dolatabadi2020advflow}, could be potentially used for privacy protection.
    \item It is still hard to evaluate the effectiveness of this mechanism with respect to privacy and utility. The existing works use the change of ML outputs (labels) to evaluate the privacy protection methods. We need to prompt more concise and better evaluation metrics.
    \item There have been some recent research works that connect the DP framework and adversarial example~\cite{lecuyer2019certified}. The PixelDP algorithm~\cite{lecuyer2019certified} proposed to add a DP-noise to the input or any middle layer to the network’s architecture to provide guaranteed robustness against adversarial examples., In more details, if we consider ``a DNN’s input (e.g., images) as databases in DP parlance, and individual features (e.g., pixels) as rows in DP’’, randomizing the output prediction function to enforce DP can guarantee the robustness of predictions against adversarial examples. PixelDP cannot effectively preserve privacy in the training set as the input changes are restricted to ``a small number of pixels’’~\cite{lecuyer2019certified}. Phan et al.~\cite{Phan2019} proposed a heterogeneous Gaussian Mechanism (HGM) that can preserve DP in training data and provide provable robustness against adversarial examples at the same time. They further proposed the stochastic batch mechanism in~\cite{Phan2020} that can retain higher model utility and is more scalable to large DNNs and datasets, compared with HGM. Overall, the interplay among DP, adversarial example and certified robustness would be a very interesting future topic.
\end{itemize}

\subsection{ML-aided Privacy Protection: GAN and VAE}
Excessive amounts of unstructured data including images, videos, audios and texts are being generated constantly and are being used by the government and a wide range of industries. According to the projections of the international data corporation, unstructured data will constitute approximately 80 percent of worldwide data by 2025. Unstructured data, especially image and videos, often containing rich personal information, play a key role in the future privacy preservation ecosystem. And the problem of private data release for unstructured data will be a hot topic in the future. 

We expect GAN to play an important role in this area, as it has demonstrated the capability to preserve high utility for ML algorithms while protecting sensitive information in the dataset.
Moreover, GAN, as part of VAE, might also be used for privacy protection for a signal data entry (i.e., an image). In this case, we can encode an original data entry and then decode it with some additional privacy protection.

\section{Conclusion}
\label{section:conclusion}
This study surveys the literature on privacy in the context of machine learning. 
By classifying the existing research into three groups: 
(i) private machine learning, 
(ii)machine learning aided privacy protection, 
and (iii) privacy protection against machine learning attack, 
we comprehensively review the state-of-art techniques on this topic and draw several conclusions as follows.
\begin{itemize}
    \item The private machine learning problem has drawn the most attention recently. 
    In this category of research works, 
    many try to use the differential privacy criterion during the analysis. 
    However, 
    DP notation cannot provide comprehensive privacy evaluation due to the complexity of the data and privacy protection target.
    Therefore, 
    how to define new privacy metrics and notations is still an open question.
    \item The research on machine learning aided privacy protection is gaining momentum these days. 
    For example, 
    using GNN to generate synthetic datasets opens the new  direction  for  privacy  protection  research, 
    especially for unstructured data such as image and video.
    \item Research on protection schemes against ML-based privacy attack is in its infancy. 
    But it is expected to fly in the future due to the proliferation of AI techniques in every corner of the future networks. 
    Currently, mainstream technology in this category is the adversarial example/perturbation technique.
\end{itemize}

We believe our timely study will shed valuable light on the research
problems associated with privacy and machine learning. 
With the increasing attention paid to this topic, 
we would expect to see increasing research activities in this area.

%
\bibliographystyle{ACM-Reference-Format}
\bibliography{PrivacyML-survey-citations}


\begin{thebibliography}{181}


\ifx \showCODEN    \undefined \def \showCODEN     #1{\unskip}     \fi
\ifx \showDOI      \undefined \def \showDOI       #1{#1}\fi
\ifx \showISBNx    \undefined \def \showISBNx     #1{\unskip}     \fi
\ifx \showISBNxiii \undefined \def \showISBNxiii  #1{\unskip}     \fi
\ifx \showISSN     \undefined \def \showISSN      #1{\unskip}     \fi
\ifx \showLCCN     \undefined \def \showLCCN      #1{\unskip}     \fi
\ifx \shownote     \undefined \def \shownote      #1{#1}          \fi
\ifx \showarticletitle \undefined \def \showarticletitle #1{#1}   \fi
\ifx \showURL      \undefined \def \showURL       {\relax}        \fi
\providecommand\bibfield[2]{#2}
\providecommand\bibinfo[2]{#2}
\providecommand\natexlab[1]{#1}
\providecommand\showeprint[2][]{arXiv:#2}

\bibitem[\protect\citeauthoryear{Abadi, Erlingsson, Goodfellow, McMahan,
  Mironov, Papernot, Talwar, and Zhang}{Abadi et~al\mbox{.}}{2017}]%
        {Abadi2017}
\bibfield{author}{\bibinfo{person}{Mart{\'{i}}n Abadi},
  \bibinfo{person}{{\'{U}}lfar Erlingsson}, \bibinfo{person}{Ian Goodfellow},
  \bibinfo{person}{H~Brendan McMahan}, \bibinfo{person}{Ilya Mironov},
  \bibinfo{person}{Nicolas Papernot}, \bibinfo{person}{Kunal Talwar}, {and}
  \bibinfo{person}{Li Zhang}.} \bibinfo{year}{2017}\natexlab{}.
\newblock \showarticletitle{{On the Protection of Private Information in
  Machine Learning Systems: Two Recent Approches}}. In
  \bibinfo{booktitle}{\emph{Proceedings of IEEE Computer Security Foundations
  Symposium (CSF'17)}}. \bibinfo{pages}{1--6}.
\newblock
\showISBNx{9781538632161}
\showISSN{19401434}
\urldef\tempurl%
\url{https://doi.org/10.1109/CSF.2017.10}
\showDOI{\tempurl}


\bibitem[\protect\citeauthoryear{Abadi, McMahan, Chu, Mironov, Zhang,
  Goodfellow, and Talwar}{Abadi et~al\mbox{.}}{2016}]%
        {Abadi2016}
\bibfield{author}{\bibinfo{person}{Mart{\'{i}}n Abadi},
  \bibinfo{person}{H.~Brendan McMahan}, \bibinfo{person}{Andy Chu},
  \bibinfo{person}{Ilya Mironov}, \bibinfo{person}{Li Zhang},
  \bibinfo{person}{Ian Goodfellow}, {and} \bibinfo{person}{Kunal Talwar}.}
  \bibinfo{year}{2016}\natexlab{}.
\newblock \showarticletitle{{Deep learning with differential privacy}}. In
  \bibinfo{booktitle}{\emph{Proceedings of the ACM Conference on Computer and
  Communications Security (CCS'16)}}. \bibinfo{pages}{308--318}.
\newblock
\showISBNx{9781450341394}
\showISSN{15437221}
\urldef\tempurl%
\url{https://doi.org/10.1145/2976749.2978318}
\showDOI{\tempurl}


\bibitem[\protect\citeauthoryear{Acquisti, Brandimarte, and
  Loewenstein}{Acquisti et~al\mbox{.}}{2015}]%
        {Acquisti2015}
\bibfield{author}{\bibinfo{person}{Alessandro Acquisti}, \bibinfo{person}{Laura
  Brandimarte}, {and} \bibinfo{person}{George Loewenstein}.}
  \bibinfo{year}{2015}\natexlab{}.
\newblock \showarticletitle{Privacy and human behavior in the age of
  information}.
\newblock \bibinfo{journal}{\emph{Science}} \bibinfo{volume}{347},
  \bibinfo{number}{6221} (\bibinfo{year}{2015}), \bibinfo{pages}{509--514}.
\newblock
\urldef\tempurl%
\url{https://doi.org/10.1126/science.aaa1465}
\showDOI{\tempurl}


\bibitem[\protect\citeauthoryear{Acs, Melis, Castelluccia, and {De
  Cristofaro}}{Acs et~al\mbox{.}}{2019}]%
        {Acs2019}
\bibfield{author}{\bibinfo{person}{Gergely Acs}, \bibinfo{person}{Luca Melis},
  \bibinfo{person}{Claude Castelluccia}, {and} \bibinfo{person}{Emiliano {De
  Cristofaro}}.} \bibinfo{year}{2019}\natexlab{}.
\newblock \showarticletitle{{Differentially Private Mixture of Generative
  Neural Networks}}.
\newblock \bibinfo{journal}{\emph{IEEE Trans. Knowl. Data Eng.}}
  \bibinfo{volume}{31}, \bibinfo{number}{6} (\bibinfo{year}{2019}),
  \bibinfo{pages}{1109--1121}.
\newblock
\showISSN{15582191}
\urldef\tempurl%
\url{https://doi.org/10.1109/TKDE.2018.2855136}
\showDOI{\tempurl}


\bibitem[\protect\citeauthoryear{Aggarwal}{Aggarwal}{2005}]%
        {Aggarwal2005}
\bibfield{author}{\bibinfo{person}{Charu~C Aggarwal}.}
  \bibinfo{year}{2005}\natexlab{}.
\newblock \showarticletitle{{On k-anonymity and the curse of dimensionality}}.
  In \bibinfo{booktitle}{\emph{Proceedings of the 31st international conference
  on Very large data bases (VLDB'05)}}. \bibinfo{pages}{901--909}.
\newblock


\bibitem[\protect\citeauthoryear{Aggarwal, Feder, Kenthapadi, Motwani,
  Panigrahy, Thomas, and Zhu}{Aggarwal et~al\mbox{.}}{2005}]%
        {Aggarwal2005a}
\bibfield{author}{\bibinfo{person}{Gagan Aggarwal}, \bibinfo{person}{Tom{\'a}s
  Feder}, \bibinfo{person}{Krishnaram Kenthapadi}, \bibinfo{person}{Rajeev
  Motwani}, \bibinfo{person}{Rina Panigrahy}, \bibinfo{person}{Dilys Thomas},
  {and} \bibinfo{person}{An Zhu}.} \bibinfo{year}{2005}\natexlab{}.
\newblock \showarticletitle{Anonymizing tables}. In
  \bibinfo{booktitle}{\emph{International Conference on Database Theory}}.
  Springer, \bibinfo{pages}{246--258}.
\newblock


\bibitem[\protect\citeauthoryear{Agrawal and Srikant}{Agrawal and
  Srikant}{2000}]%
        {Agrawal2000}
\bibfield{author}{\bibinfo{person}{Rakesh Agrawal} {and}
  \bibinfo{person}{Ramakrishnan Srikant}.} \bibinfo{year}{2000}\natexlab{}.
\newblock \showarticletitle{{Privacy-preserving data mining}}. In
  \bibinfo{booktitle}{\emph{Proceedings of the ACM SIGMOD international
  conference on Management of data (SIGMOD'00)}}. \bibinfo{publisher}{ACM
  Press}, \bibinfo{address}{New York, New York, USA},
  \bibinfo{pages}{439--450}.
\newblock
\showISBNx{1581132174}
\urldef\tempurl%
\url{https://doi.org/10.1145/342009.335438}
\showDOI{\tempurl}


\bibitem[\protect\citeauthoryear{Ateniese, Mancini, Spognardi, Villani, Vitali,
  and Felici}{Ateniese et~al\mbox{.}}{2015}]%
        {Ateniese2015}
\bibfield{author}{\bibinfo{person}{Giuseppe Ateniese}, \bibinfo{person}{Luigi~V
  Mancini}, \bibinfo{person}{Angelo Spognardi}, \bibinfo{person}{Antonio
  Villani}, \bibinfo{person}{Domenico Vitali}, {and} \bibinfo{person}{Giovanni
  Felici}.} \bibinfo{year}{2015}\natexlab{}.
\newblock \showarticletitle{{Hacking smart machines with smarter ones: How to
  extract meaningful data from machine learning classifiers}}.
\newblock \bibinfo{journal}{\emph{International Journal of Security and
  Networks}} \bibinfo{volume}{10}, \bibinfo{number}{3} (\bibinfo{year}{2015}),
  \bibinfo{pages}{137--150}.
\newblock
\showISSN{17478413}
\urldef\tempurl%
\url{https://doi.org/10.1504/IJSN.2015.071829}
\showDOI{\tempurl}


\bibitem[\protect\citeauthoryear{Avdiienko, Kuznetsov, Gorla, Zeller, Arzt,
  Rasthofer, and Bodden}{Avdiienko et~al\mbox{.}}{2015}]%
        {Avdiienko2015}
\bibfield{author}{\bibinfo{person}{Vitalii Avdiienko},
  \bibinfo{person}{Konstantin Kuznetsov}, \bibinfo{person}{Alessandra Gorla},
  \bibinfo{person}{Andreas Zeller}, \bibinfo{person}{Steven Arzt},
  \bibinfo{person}{Siegfried Rasthofer}, {and} \bibinfo{person}{Eric Bodden}.}
  \bibinfo{year}{2015}\natexlab{}.
\newblock \showarticletitle{{Mining apps for abnormal usage of sensitive
  data}}. In \bibinfo{booktitle}{\emph{Proceedings International Conference on
  Software Engineering (ICSE'15)}}, Vol.~\bibinfo{volume}{1}.
  \bibinfo{pages}{426--436}.
\newblock
\showISBNx{9781479919345}
\showISSN{02705257}
\urldef\tempurl%
\url{https://doi.org/10.1109/ICSE.2015.61}
\showDOI{\tempurl}


\bibitem[\protect\citeauthoryear{Bishop}{Bishop}{2006}]%
        {Bishop2006}
\bibfield{author}{\bibinfo{person}{Christopher~M Bishop}.}
  \bibinfo{year}{2006}\natexlab{}.
\newblock \bibinfo{booktitle}{\emph{{Pattern Recognition and Machine Learning
  (Information Science and Statistics)}}}.
\newblock \bibinfo{publisher}{Springer-Verlag New York, Inc.},
  \bibinfo{address}{Secaucus, NJ, USA}.
\newblock
\showISBNx{0387310738}


\bibitem[\protect\citeauthoryear{Blaauw and Bonada}{Blaauw and Bonada}{2017}]%
        {Blaauw2017}
\bibfield{author}{\bibinfo{person}{Merlijn Blaauw} {and} \bibinfo{person}{Jordi
  Bonada}.} \bibinfo{year}{2017}\natexlab{}.
\newblock \showarticletitle{{A neural parametric singing synthesizer}}. In
  \bibinfo{booktitle}{\emph{Proceedings of the Annual Conference of the
  International Speech Communication Association (INTERSPEECH'17)}},
  Vol.~\bibinfo{volume}{2017-Augus}. \bibinfo{pages}{4001--4005}.
\newblock
\showISSN{19909772}


\bibitem[\protect\citeauthoryear{Bonawitz, Ivanov, Kreuter, Marcedone, McMahan,
  Patel, Ramage, Segal, and Seth}{Bonawitz et~al\mbox{.}}{2017}]%
        {Bonawitz2017}
\bibfield{author}{\bibinfo{person}{Keith Bonawitz}, \bibinfo{person}{Vladimir
  Ivanov}, \bibinfo{person}{Ben Kreuter}, \bibinfo{person}{Antonio Marcedone},
  \bibinfo{person}{H~Brendan McMahan}, \bibinfo{person}{Sarvar Patel},
  \bibinfo{person}{Daniel Ramage}, \bibinfo{person}{Aaron Segal}, {and}
  \bibinfo{person}{Karn Seth}.} \bibinfo{year}{2017}\natexlab{}.
\newblock \showarticletitle{{Practical secure aggregation for
  privacy-preserving machine learning}}. In
  \bibinfo{booktitle}{\emph{Proceedings of the ACM Conference on Computer and
  Communications Security (CCS'17)}}. \bibinfo{pages}{1175--1191}.
\newblock
\showISBNx{9781450349468}
\showISSN{15437221}
\urldef\tempurl%
\url{https://doi.org/10.1145/3133956.3133982}
\showDOI{\tempurl}


\bibitem[\protect\citeauthoryear{Bost, Popa, Tu, and Goldwasser}{Bost
  et~al\mbox{.}}{2015}]%
        {Bost2015}
\bibfield{author}{\bibinfo{person}{Raphael Bost}, \bibinfo{person}{Raluca~Ada
  Popa}, \bibinfo{person}{Stephen Tu}, {and} \bibinfo{person}{Shafi
  Goldwasser}.} \bibinfo{year}{2015}\natexlab{}.
\newblock \showarticletitle{{Machine Learning Classification over Encrypted
  Data}}. In \bibinfo{booktitle}{\emph{Proceeding of The Network and
  Distributed System Security Symposium (NDSS'15)}}.
\newblock


\bibitem[\protect\citeauthoryear{{Brendan McMahan}, Moore, Ramage, Hampson, and
  {Ag{\"{u}}era y Arcas}}{{Brendan McMahan} et~al\mbox{.}}{2017}]%
        {BrendanMcMahan2017}
\bibfield{author}{\bibinfo{person}{H. {Brendan McMahan}},
  \bibinfo{person}{Eider Moore}, \bibinfo{person}{Daniel Ramage},
  \bibinfo{person}{Seth Hampson}, {and} \bibinfo{person}{Blaise {Ag{\"{u}}era y
  Arcas}}.} \bibinfo{year}{2017}\natexlab{}.
\newblock \showarticletitle{{Communication-efficient learning of deep networks
  from decentralized data}}. In \bibinfo{booktitle}{\emph{Proceedings of the
  20th International Conference on Artificial Intelligence and Statistics
  (AISTATS'17)}}.
\newblock


\bibitem[\protect\citeauthoryear{Brickell, Porter, Shmatikov, and
  Witchel}{Brickell et~al\mbox{.}}{2007}]%
        {Brickell2007}
\bibfield{author}{\bibinfo{person}{Justin Brickell}, \bibinfo{person}{Donald~E
  Porter}, \bibinfo{person}{Vitaly Shmatikov}, {and} \bibinfo{person}{Emmett
  Witchel}.} \bibinfo{year}{2007}\natexlab{}.
\newblock \showarticletitle{{Privacy-preserving remote diagnostics}}. In
  \bibinfo{booktitle}{\emph{Proceedings of the 14th ACM conference on Computer
  and communications security (CCS'07)}}. \bibinfo{pages}{498--507}.
\newblock


\bibitem[\protect\citeauthoryear{Bunn and Ostrovsky}{Bunn and
  Ostrovsky}{2007}]%
        {Bunn2007}
\bibfield{author}{\bibinfo{person}{Paul Bunn} {and} \bibinfo{person}{Rafail
  Ostrovsky}.} \bibinfo{year}{2007}\natexlab{}.
\newblock \showarticletitle{{Secure two-party k-means clustering}}. In
  \bibinfo{booktitle}{\emph{Proceedings of the 14th ACM conference on Computer
  and communications security (CCS'07)}}. \bibinfo{pages}{486--497}.
\newblock


\bibitem[\protect\citeauthoryear{Center}{Center}{2014}]%
        {Center2014}
\bibfield{author}{\bibinfo{person}{Pew~Research Center}.}
  \bibinfo{year}{2014}\natexlab{}.
\newblock \showarticletitle{{Public Perceptions of Privacy and Security}}.
\newblock \bibinfo{journal}{\emph{Pew Research Center}} (\bibinfo{year}{2014}).
\newblock
\urldef\tempurl%
\url{http://www.pewinternet.org/2014/11/12/public-privacy-perceptions/}
\showURL{%
\tempurl}


\bibitem[\protect\citeauthoryear{Chaudhuri, Monteleoni, and Sarwate}{Chaudhuri
  et~al\mbox{.}}{2011}]%
        {Chaudhuri2011}
\bibfield{author}{\bibinfo{person}{Kamalika Chaudhuri}, \bibinfo{person}{Claire
  Monteleoni}, {and} \bibinfo{person}{Anand~D Sarwate}.}
  \bibinfo{year}{2011}\natexlab{}.
\newblock \showarticletitle{{Differentially private empirical risk
  minimization}}.
\newblock \bibinfo{journal}{\emph{Journal of Machine Learning Research}}
  \bibinfo{volume}{12}, \bibinfo{number}{Mar} (\bibinfo{year}{2011}),
  \bibinfo{pages}{1069--1109}.
\newblock


\bibitem[\protect\citeauthoryear{Chen, Zhang, Sharma, Yi, and Hsieh}{Chen
  et~al\mbox{.}}{2017}]%
        {chen2017zoo}
\bibfield{author}{\bibinfo{person}{Pin-Yu Chen}, \bibinfo{person}{Huan Zhang},
  \bibinfo{person}{Yash Sharma}, \bibinfo{person}{Jinfeng Yi}, {and}
  \bibinfo{person}{Cho-Jui Hsieh}.} \bibinfo{year}{2017}\natexlab{}.
\newblock \showarticletitle{Zoo: Zeroth order optimization based black-box
  attacks to deep neural networks without training substitute models}. In
  \bibinfo{booktitle}{\emph{Proceedings of the 10th ACM Workshop on Artificial
  Intelligence and Security}}. \bibinfo{pages}{15--26}.
\newblock


\bibitem[\protect\citeauthoryear{Cheng, Caverlee, and Lee}{Cheng
  et~al\mbox{.}}{2010}]%
        {Cheng2010}
\bibfield{author}{\bibinfo{person}{Zhiyuan Cheng}, \bibinfo{person}{James
  Caverlee}, {and} \bibinfo{person}{Kyumin Lee}.}
  \bibinfo{year}{2010}\natexlab{}.
\newblock \showarticletitle{{You are where you tweet: a content-based approach
  to geo-locating twitter users}}. In \bibinfo{booktitle}{\emph{Proceedings of
  ACM international conference on Information and knowledge management
  (CIKM'10)}}. \bibinfo{pages}{759--768}.
\newblock


\bibitem[\protect\citeauthoryear{Cheung, Wildfeuer, Nikkhah, Zhu, and
  Tan}{Cheung et~al\mbox{.}}{2018}]%
        {Cheung2018}
\bibfield{author}{\bibinfo{person}{Sen-Ching~Samson Cheung},
  \bibinfo{person}{Herb Wildfeuer}, \bibinfo{person}{Mehdi Nikkhah},
  \bibinfo{person}{Xiaoqing Zhu}, {and} \bibinfo{person}{Waitian Tan}.}
  \bibinfo{year}{2018}\natexlab{}.
\newblock \showarticletitle{{Learning Sensitive Images Using Generative
  Models}}. In \bibinfo{booktitle}{\emph{Proceedings of the 25th IEEE
  International Conference on Image Processing (ICIP'18)}}.
  \bibinfo{pages}{4128--4132}.
\newblock


\bibitem[\protect\citeauthoryear{Cho, Myers, and Leskovec}{Cho
  et~al\mbox{.}}{2011}]%
        {Cho2011}
\bibfield{author}{\bibinfo{person}{Eunjoon Cho}, \bibinfo{person}{Seth~A
  Myers}, {and} \bibinfo{person}{Jure Leskovec}.}
  \bibinfo{year}{2011}\natexlab{}.
\newblock \showarticletitle{{Friendship and mobility: user movement in
  location-based social networks}}. In \bibinfo{booktitle}{\emph{Proceedings of
  the ACM International Conference on Knowledge Discovery and Data Mining
  (KDD'11)}}. \bibinfo{pages}{1082--1090}.
\newblock


\bibitem[\protect\citeauthoryear{Choi, Biswal, Malin, Duke, Stewart, and
  Sun}{Choi et~al\mbox{.}}{2017}]%
        {Choi2017}
\bibfield{author}{\bibinfo{person}{Edward Choi}, \bibinfo{person}{Siddharth
  Biswal}, \bibinfo{person}{Bradley Malin}, \bibinfo{person}{Jon Duke},
  \bibinfo{person}{Walter~F. Stewart}, {and} \bibinfo{person}{Jimeng Sun}.}
  \bibinfo{year}{2017}\natexlab{}.
\newblock \showarticletitle{{Generating Multi-label Discrete Patient Records
  using Generative Adversarial Networks}}.
\newblock \bibinfo{journal}{\emph{Proceedings of the Machine Learning for
  Healthcare Conference}} (\bibinfo{year}{2017}).
\newblock
\urldef\tempurl%
\url{http://proceedings.mlr.press/v68/choi17a.html
  http://arxiv.org/abs/1703.06490}
\showURL{%
\tempurl}


\bibitem[\protect\citeauthoryear{Costante, {Den Hartog}, and Petkovic}{Costante
  et~al\mbox{.}}{2011}]%
        {Costante2011}
\bibfield{author}{\bibinfo{person}{Elisa Costante}, \bibinfo{person}{Jerry {Den
  Hartog}}, {and} \bibinfo{person}{Milan Petkovic}.}
  \bibinfo{year}{2011}\natexlab{}.
\newblock \showarticletitle{{On-line trust perception: What really matters}}.
  In \bibinfo{booktitle}{\emph{Proceedings of the 1st Workshop on
  Socio-Technical Aspects in Security and Trust (STAST'11)}}.
  \bibinfo{pages}{52--59}.
\newblock


\bibitem[\protect\citeauthoryear{Costante, Sun, Petkovic, and {Den
  Hartog}}{Costante et~al\mbox{.}}{2012}]%
        {Costante2012}
\bibfield{author}{\bibinfo{person}{Elisa Costante}, \bibinfo{person}{Yuanhao
  Sun}, \bibinfo{person}{Milan Petkovic}, {and} \bibinfo{person}{Jerry {Den
  Hartog}}.} \bibinfo{year}{2012}\natexlab{}.
\newblock \showarticletitle{{A machine learning solution to assess privacy
  policy completeness (short paper)}}. In \bibinfo{booktitle}{\emph{Proceedings
  of the ACM Conference on Computer and Communications Security (CCS'12)}}.
  \bibinfo{pages}{91--96}.
\newblock
\showISBNx{9781450316637}
\showISSN{15437221}
\urldef\tempurl%
\url{https://doi.org/10.1145/2381966.2381979}
\showDOI{\tempurl}


\bibitem[\protect\citeauthoryear{Das, Degeling, Smullen, and Sadeh}{Das
  et~al\mbox{.}}{2018}]%
        {Das2018}
\bibfield{author}{\bibinfo{person}{Anupam Das}, \bibinfo{person}{Martin
  Degeling}, \bibinfo{person}{Daniel Smullen}, {and} \bibinfo{person}{Norman
  Sadeh}.} \bibinfo{year}{2018}\natexlab{}.
\newblock \showarticletitle{{Personalized privacy assistants for the internet
  of things: Providing users with notice and choice}}.
\newblock \bibinfo{journal}{\emph{IEEE Pervasive Computing}}
  \bibinfo{volume}{17}, \bibinfo{number}{3} (\bibinfo{year}{2018}),
  \bibinfo{pages}{35--46}.
\newblock
\showISSN{15582590}


\bibitem[\protect\citeauthoryear{Denton, Chintala, Szlam, and Fergus}{Denton
  et~al\mbox{.}}{2015}]%
        {Denton2015}
\bibfield{author}{\bibinfo{person}{Emily Denton}, \bibinfo{person}{Soumith
  Chintala}, \bibinfo{person}{Arthur Szlam}, {and} \bibinfo{person}{Rob
  Fergus}.} \bibinfo{year}{2015}\natexlab{}.
\newblock \showarticletitle{{Deep generative image models using a laplacian
  pyramid of adversarial networks}}. In \bibinfo{booktitle}{\emph{Advances in
  Neural Information Processing Systems (NIPS'15)}},
  Vol.~\bibinfo{volume}{2015-Janua}. \bibinfo{pages}{1486--1494}.
\newblock
\showISSN{10495258}


\bibitem[\protect\citeauthoryear{Dolatabadi, Erfani, and Leckie}{Dolatabadi
  et~al\mbox{.}}{2020}]%
        {dolatabadi2020advflow}
\bibfield{author}{\bibinfo{person}{Hadi~M Dolatabadi}, \bibinfo{person}{Sarah
  Erfani}, {and} \bibinfo{person}{Christopher Leckie}.}
  \bibinfo{year}{2020}\natexlab{}.
\newblock \showarticletitle{AdvFlow: Inconspicuous Black-box Adversarial
  Attacks using Normalizing Flows}.
\newblock \bibinfo{journal}{\emph{arXiv preprint arXiv:2007.07435}}
  (\bibinfo{year}{2020}).
\newblock


\bibitem[\protect\citeauthoryear{Dowlin, Gilad-Bachrach, Laine, Lauter,
  Naehrig, and Wernsing}{Dowlin et~al\mbox{.}}{2016}]%
        {Dowlin2016}
\bibfield{author}{\bibinfo{person}{Nathan Dowlin}, \bibinfo{person}{Ran
  Gilad-Bachrach}, \bibinfo{person}{Kim Laine}, \bibinfo{person}{Kristin
  Lauter}, \bibinfo{person}{Michael Naehrig}, {and} \bibinfo{person}{John
  Wernsing}.} \bibinfo{year}{2016}\natexlab{}.
\newblock \showarticletitle{{Cryptonets: Applying neural networks to encrypted
  data with high throughput and accuracy}}. In
  \bibinfo{booktitle}{\emph{Proceedings of the 33rd International Conference on
  Machine Learning (ICML'16)}}, Vol.~\bibinfo{volume}{1}.
  \bibinfo{pages}{342--351}.
\newblock
\showISBNx{9781510829008}


\bibitem[\protect\citeauthoryear{Du, Han, and Chen}{Du et~al\mbox{.}}{2004}]%
        {Du2004}
\bibfield{author}{\bibinfo{person}{Wenliang Du}, \bibinfo{person}{Yunghsiang~S
  Han}, {and} \bibinfo{person}{Shigang Chen}.} \bibinfo{year}{2004}\natexlab{}.
\newblock \showarticletitle{{Privacy-preserving multivariate statistical
  analysis: Linear regression and classification}}. In
  \bibinfo{booktitle}{\emph{SIAM Proceedings Series}}.
  \bibinfo{pages}{222--233}.
\newblock
\urldef\tempurl%
\url{https://doi.org/10.1137/1.9781611972740.21}
\showDOI{\tempurl}


\bibitem[\protect\citeauthoryear{Dwork}{Dwork}{2008}]%
        {Dwork2008}
\bibfield{author}{\bibinfo{person}{Cynthia Dwork}.}
  \bibinfo{year}{2008}\natexlab{}.
\newblock \showarticletitle{Differential privacy: A survey of results}. In
  \bibinfo{booktitle}{\emph{International conference on theory and applications
  of models of computation}}. Springer, \bibinfo{pages}{1--19}.
\newblock


\bibitem[\protect\citeauthoryear{Dwork, McSherry, Nissim, and Smith}{Dwork
  et~al\mbox{.}}{2006}]%
        {Dwork2006}
\bibfield{author}{\bibinfo{person}{Cynthia Dwork}, \bibinfo{person}{Frank
  McSherry}, \bibinfo{person}{Kobbi Nissim}, {and} \bibinfo{person}{Adam
  Smith}.} \bibinfo{year}{2006}\natexlab{}.
\newblock \showarticletitle{Calibrating noise to sensitivity in private data
  analysis}. In \bibinfo{booktitle}{\emph{Proceedings of the theory of
  cryptography conference (TCC'06)}}. Springer, \bibinfo{pages}{265--284}.
\newblock


\bibitem[\protect\citeauthoryear{Dwork, Roth, et~al\mbox{.}}{Dwork
  et~al\mbox{.}}{2014}]%
        {Dwork2014}
\bibfield{author}{\bibinfo{person}{Cynthia Dwork}, \bibinfo{person}{Aaron
  Roth}, {et~al\mbox{.}}} \bibinfo{year}{2014}\natexlab{}.
\newblock \showarticletitle{The algorithmic foundations of differential
  privacy.}
\newblock \bibinfo{journal}{\emph{Foundations and Trends in Theoretical
  Computer Science}} \bibinfo{volume}{9}, \bibinfo{number}{3-4}
  (\bibinfo{year}{2014}), \bibinfo{pages}{211--407}.
\newblock


\bibitem[\protect\citeauthoryear{Dwork and Theory}{Dwork and Theory}{2018}]%
        {Dwork2018}
\bibfield{author}{\bibinfo{person}{C Dwork} {and} \bibinfo{person}{V~Feldman
  Theory}.} \bibinfo{year}{2018}\natexlab{}.
\newblock \showarticletitle{{Privacy-preserving Prediction}}.
\newblock \bibinfo{journal}{\emph{Proceedings of the Machine Learning for
  Healthcare Conference}} (\bibinfo{year}{2018}).
\newblock
\urldef\tempurl%
\url{http://proceedings.mlr.press/v75/dwork18a.html}
\showURL{%
\tempurl}


\bibitem[\protect\citeauthoryear{Elsayed, Papernot, Shankar, Kurakin, Cheung,
  Goodfellow, and Sohl-Dickstein}{Elsayed et~al\mbox{.}}{2018}]%
        {Elsayed2018}
\bibfield{author}{\bibinfo{person}{Gamaleldin~F Elsayed},
  \bibinfo{person}{Nicolas Papernot}, \bibinfo{person}{Shreya Shankar},
  \bibinfo{person}{Alexey Kurakin}, \bibinfo{person}{Brian Cheung},
  \bibinfo{person}{Ian Goodfellow}, {and} \bibinfo{person}{Jascha
  Sohl-Dickstein}.} \bibinfo{year}{2018}\natexlab{}.
\newblock \showarticletitle{{Adversarial examples that fool both computer
  vision and time-limited humans}}. In \bibinfo{booktitle}{\emph{Advances in
  Neural Information Processing Systems (NIPS'18)}},
  Vol.~\bibinfo{volume}{2018-Decem}. \bibinfo{pages}{3910--3920}.
\newblock
\showISSN{10495258}


\bibitem[\protect\citeauthoryear{Fawzi, Fawzi, and Frossard}{Fawzi
  et~al\mbox{.}}{2015}]%
        {Fawzi2015}
\bibfield{author}{\bibinfo{person}{Alhussein Fawzi}, \bibinfo{person}{Omar
  Fawzi}, {and} \bibinfo{person}{Pascal Frossard}.}
  \bibinfo{year}{2015}\natexlab{}.
\newblock \showarticletitle{{Fundamental limits on adversarial robustness}}.
\newblock \bibinfo{journal}{\emph{Proceedings of International Conference on
  Machine Learning (ICML'15), Workshop Deep Learning}} (\bibinfo{year}{2015}),
  \bibinfo{pages}{1--7}.
\newblock


\bibitem[\protect\citeauthoryear{Fox and Moreland}{Fox and Moreland}{2015}]%
        {Fox2015}
\bibfield{author}{\bibinfo{person}{Jesse Fox} {and} \bibinfo{person}{Jennifer~J
  Moreland}.} \bibinfo{year}{2015}\natexlab{}.
\newblock \showarticletitle{{The dark side of social networking sites: An
  exploration of the relational and psychological stressors associated with
  Facebook use and affordances}}.
\newblock \bibinfo{journal}{\emph{Computers in Human Behavior}}
  \bibinfo{volume}{45} (\bibinfo{year}{2015}), \bibinfo{pages}{168--176}.
\newblock
\showISSN{07475632}
\urldef\tempurl%
\url{https://doi.org/10.1016/j.chb.2014.11.083}
\showDOI{\tempurl}


\bibitem[\protect\citeauthoryear{Fredrikson, Jha, and Ristenpart}{Fredrikson
  et~al\mbox{.}}{2015}]%
        {Fredrikson2015}
\bibfield{author}{\bibinfo{person}{Matt Fredrikson}, \bibinfo{person}{Somesh
  Jha}, {and} \bibinfo{person}{Thomas Ristenpart}.}
  \bibinfo{year}{2015}\natexlab{}.
\newblock \showarticletitle{{Model inversion attacks that exploit confidence
  information and basic countermeasures}}. In
  \bibinfo{booktitle}{\emph{Proceedings of the ACM Conference on Computer and
  Communications Security (CCS'15)}}. \bibinfo{publisher}{ACM},
  \bibinfo{pages}{1322--1333}.
\newblock
\showISBNx{9781450338325}
\showISSN{15437221}
\urldef\tempurl%
\url{https://doi.org/10.1145/2810103.2813677}
\showDOI{\tempurl}


\bibitem[\protect\citeauthoryear{Fredrikson, Lantz, Jha, Lin, Page, and
  Ristenpart}{Fredrikson et~al\mbox{.}}{2014}]%
        {Fredrikson2014}
\bibfield{author}{\bibinfo{person}{Matthew Fredrikson}, \bibinfo{person}{Eric
  Lantz}, \bibinfo{person}{Somesh Jha}, \bibinfo{person}{Simon Lin},
  \bibinfo{person}{David Page}, {and} \bibinfo{person}{Thomas Ristenpart}.}
  \bibinfo{year}{2014}\natexlab{}.
\newblock \showarticletitle{{Privacy in pharmacogenetics: An end-to-end case
  study of personalized warfarin dosing}}. In
  \bibinfo{booktitle}{\emph{Proceedings of the 23rd USENIX Security Symposium
  (USENIX'14)}}. \bibinfo{pages}{17--32}.
\newblock
\showISBNx{9781931971157}


\bibitem[\protect\citeauthoryear{Friedman, Berkovsky, and Kaafar}{Friedman
  et~al\mbox{.}}{2016}]%
        {Friedman2016}
\bibfield{author}{\bibinfo{person}{Arik Friedman}, \bibinfo{person}{Shlomo
  Berkovsky}, {and} \bibinfo{person}{Mohamed~Ali Kaafar}.}
  \bibinfo{year}{2016}\natexlab{}.
\newblock \showarticletitle{{A differential privacy framework for matrix
  factorization recommender systems}}.
\newblock \bibinfo{journal}{\emph{User Modeling and User-Adapted Interaction}}
  \bibinfo{volume}{26}, \bibinfo{number}{5} (\bibinfo{date}{Dec}
  \bibinfo{year}{2016}), \bibinfo{pages}{425--458}.
\newblock
\showISSN{15731391}


\bibitem[\protect\citeauthoryear{Friedrich, K{\"{o}}hn, Wiedemann, and
  Biemann}{Friedrich et~al\mbox{.}}{2020}]%
        {Friedrich2020}
\bibfield{author}{\bibinfo{person}{Max Friedrich}, \bibinfo{person}{Arne
  K{\"{o}}hn}, \bibinfo{person}{Gregor Wiedemann}, {and} \bibinfo{person}{Chris
  Biemann}.} \bibinfo{year}{2020}\natexlab{}.
\newblock \showarticletitle{{Adversarial learning of privacy-preserving text
  representations for de-identification of medical records}}. In
  \bibinfo{booktitle}{\emph{Proceedings of the 57th Annual Meeting of the
  Association for Computational Linguistics (ACL'20)}}.
  \bibinfo{pages}{5829--5839}.
\newblock
\showISBNx{9781950737482}
\urldef\tempurl%
\url{https://doi.org/10.18653/v1/p19-1584}
\showDOI{\tempurl}


\bibitem[\protect\citeauthoryear{Fu, Zheng, Zhu, and Mohapatra}{Fu
  et~al\mbox{.}}{2019}]%
        {Fu2019}
\bibfield{author}{\bibinfo{person}{Hao Fu}, \bibinfo{person}{Zizhan Zheng},
  \bibinfo{person}{Sencun Zhu}, {and} \bibinfo{person}{Prasant Mohapatra}.}
  \bibinfo{year}{2019}\natexlab{}.
\newblock \showarticletitle{{Keeping Context in Mind: Automating Mobile App
  Access Control with User Interface Inspection}}. In
  \bibinfo{booktitle}{\emph{Proceedings of IEEE International Conference on
  Computer Communications (INFOCOM'19)}}. \bibinfo{pages}{2089--2097}.
\newblock
\showISBNx{9781728105154}
\showISSN{0743166X}
\urldef\tempurl%
\url{https://doi.org/10.1109/INFOCOM.2019.8737510}
\showDOI{\tempurl}


\bibitem[\protect\citeauthoryear{Gai, Qiu, Zhao, and Xiong}{Gai
  et~al\mbox{.}}{2016}]%
        {Gai2016}
\bibfield{author}{\bibinfo{person}{Keke Gai}, \bibinfo{person}{Meikang Qiu},
  \bibinfo{person}{Hui Zhao}, {and} \bibinfo{person}{Jian Xiong}.}
  \bibinfo{year}{2016}\natexlab{}.
\newblock \showarticletitle{{Privacy-Aware Adaptive Data Encryption Strategy of
  Big Data in Cloud Computing}}. In \bibinfo{booktitle}{\emph{Proceedings of
  3rd IEEE International Conference on Cyber Security and Cloud Computing
  (CSCloud'16)}}. \bibinfo{pages}{273--278}.
\newblock
\showISBNx{9781509009459}
\urldef\tempurl%
\url{https://doi.org/10.1109/CSCloud.2016.52}
\showDOI{\tempurl}


\bibitem[\protect\citeauthoryear{Goodfellow}{Goodfellow}{2018}]%
        {Goodfellow2018}
\bibfield{author}{\bibinfo{person}{Ian Goodfellow}.}
  \bibinfo{year}{2018}\natexlab{}.
\newblock \showarticletitle{{Defense Against the Dark Arts: An overview of
  adversarial example security research and future research directions}}.
\newblock \bibinfo{journal}{\emph{arXiv:1806.04169}} (\bibinfo{year}{2018}).
\newblock
\showeprint{1806.04169}
\urldef\tempurl%
\url{http://arxiv.org/abs/1806.04169}
\showURL{%
\tempurl}


\bibitem[\protect\citeauthoryear{Goodfellow, Pouget-Abadie, Mirza, Xu,
  Warde-Farley, Ozair, Courville, and Bengio}{Goodfellow et~al\mbox{.}}{2014}]%
        {Goodfellow2014}
\bibfield{author}{\bibinfo{person}{Ian~J. Goodfellow}, \bibinfo{person}{Jean
  Pouget-Abadie}, \bibinfo{person}{Mehdi Mirza}, \bibinfo{person}{Bing Xu},
  \bibinfo{person}{David Warde-Farley}, \bibinfo{person}{Sherjil Ozair},
  \bibinfo{person}{Aaron Courville}, {and} \bibinfo{person}{Yoshua Bengio}.}
  \bibinfo{year}{2014}\natexlab{}.
\newblock \showarticletitle{{Generative adversarial nets}}. In
  \bibinfo{booktitle}{\emph{Advances in Neural Information Processing Systems
  (NIPS'14)}}, Vol.~\bibinfo{volume}{3}. \bibinfo{pages}{2672--2680}.
\newblock
\showISSN{10495258}


\bibitem[\protect\citeauthoryear{Gorla, Tavecchia, Gross, and Zeller}{Gorla
  et~al\mbox{.}}{2014}]%
        {Gorla2014}
\bibfield{author}{\bibinfo{person}{Alessandra Gorla}, \bibinfo{person}{Ilaria
  Tavecchia}, \bibinfo{person}{Florian Gross}, {and} \bibinfo{person}{Andreas
  Zeller}.} \bibinfo{year}{2014}\natexlab{}.
\newblock \showarticletitle{{Checking app behavior against app descriptions}}.
  In \bibinfo{booktitle}{\emph{Proceedings of the 36th International Conference
  on Software Engineering (ICSE'14)}}. \bibinfo{pages}{1025--1035}.
\newblock


\bibitem[\protect\citeauthoryear{Graepel, Lauter, and Naehrig}{Graepel
  et~al\mbox{.}}{2012}]%
        {Graepel2012}
\bibfield{author}{\bibinfo{person}{Thore Graepel}, \bibinfo{person}{Kristin
  Lauter}, {and} \bibinfo{person}{Michael Naehrig}.}
  \bibinfo{year}{2012}\natexlab{}.
\newblock \showarticletitle{{ML confidential: Machine learning on encrypted
  data}}. In \bibinfo{booktitle}{\emph{Proceedings of the International
  Conference on Information Security and Cryptology (ICISC'12)}}.
  \bibinfo{pages}{1--21}.
\newblock


\bibitem[\protect\citeauthoryear{Gregor, Danihelka, Graves, Rezende, and
  Wierstra}{Gregor et~al\mbox{.}}{2015}]%
        {Gregor2015}
\bibfield{author}{\bibinfo{person}{Karol Gregor}, \bibinfo{person}{Ivo
  Danihelka}, \bibinfo{person}{Alex Graves}, \bibinfo{person}{Danilo~Jimenez
  Rezende}, {and} \bibinfo{person}{Daan Wierstra}.}
  \bibinfo{year}{2015}\natexlab{}.
\newblock \showarticletitle{{DRAW: A recurrent neural network for image
  generation}}. In \bibinfo{booktitle}{\emph{Proceedings of the 32nd
  International Conference on Machine Learning (ICML'15)}},
  Vol.~\bibinfo{volume}{2}. \bibinfo{pages}{1462--1471}.
\newblock
\showISBNx{9781510810587}


\bibitem[\protect\citeauthoryear{Gu, Dolan-Gavitt, and Garg}{Gu
  et~al\mbox{.}}{2017}]%
        {gu2017badnets}
\bibfield{author}{\bibinfo{person}{Tianyu Gu}, \bibinfo{person}{Brendan
  Dolan-Gavitt}, {and} \bibinfo{person}{Siddharth Garg}.}
  \bibinfo{year}{2017}\natexlab{}.
\newblock \showarticletitle{Badnets: Identifying vulnerabilities in the machine
  learning model supply chain}.
\newblock \bibinfo{journal}{\emph{arXiv preprint arXiv:1708.06733}}
  (\bibinfo{year}{2017}).
\newblock


\bibitem[\protect\citeauthoryear{Gu, Yao, Liu, and Song}{Gu
  et~al\mbox{.}}{2016}]%
        {Gu2016}
\bibfield{author}{\bibinfo{person}{Yulong Gu}, \bibinfo{person}{Yuan Yao},
  \bibinfo{person}{Weidong Liu}, {and} \bibinfo{person}{Jiaxing Song}.}
  \bibinfo{year}{2016}\natexlab{}.
\newblock \showarticletitle{{We know where you are: Home location
  identification in location-based social networks}}. In
  \bibinfo{booktitle}{\emph{Proceedings of the 25th International Conference on
  Computer Communications and Networks (ICCCN'16)}}. \bibinfo{pages}{1--9}.
\newblock
\showISBNx{9781509022793}
\urldef\tempurl%
\url{https://doi.org/10.1109/ICCCN.2016.7568598}
\showDOI{\tempurl}


\bibitem[\protect\citeauthoryear{Hasan, Crandall, and Kapadia}{Hasan
  et~al\mbox{.}}{2020}]%
        {Hasan2020}
\bibfield{author}{\bibinfo{person}{Rakibul Hasan}, \bibinfo{person}{David
  Crandall}, {and} \bibinfo{person}{Mario Fritz~Apu Kapadia}.}
  \bibinfo{year}{2020}\natexlab{}.
\newblock \showarticletitle{{Automatically Detecting Bystanders in Photos to
  Reduce Privacy Risks}}. In \bibinfo{booktitle}{\emph{Proceedings of the IEEE
  Symposium on Security and Privacy (SP'20)}}. \bibinfo{pages}{318--335}.
\newblock
\urldef\tempurl%
\url{https://doi.org/10.1109/SP40000.2020.00097}
\showDOI{\tempurl}


\bibitem[\protect\citeauthoryear{Hayes, Melis, Danezis, and {De
  Cristofaro}}{Hayes et~al\mbox{.}}{2019}]%
        {Hayes2019}
\bibfield{author}{\bibinfo{person}{Jamie Hayes}, \bibinfo{person}{Luca Melis},
  \bibinfo{person}{George Danezis}, {and} \bibinfo{person}{Emiliano {De
  Cristofaro}}.} \bibinfo{year}{2019}\natexlab{}.
\newblock \showarticletitle{{LOGAN: Membership Inference Attacks Against
  Generative Models}}. In \bibinfo{booktitle}{\emph{Proceedings on Privacy
  Enhancing Technologies (PETS'19)}}. \bibinfo{pages}{133--152}.
\newblock
\urldef\tempurl%
\url{https://doi.org/10.2478/popets-2019-0008}
\showDOI{\tempurl}


\bibitem[\protect\citeauthoryear{Henriksen-Bulmer and Jeary}{Henriksen-Bulmer
  and Jeary}{2016}]%
        {Henriksen-Bulmer2016}
\bibfield{author}{\bibinfo{person}{Jane Henriksen-Bulmer} {and}
  \bibinfo{person}{Sheridan Jeary}.} \bibinfo{year}{2016}\natexlab{}.
\newblock \showarticletitle{{Re-identification attacks—A systematic
  literature review}}.
\newblock \bibinfo{journal}{\emph{International Journal of Information
  Management}} \bibinfo{volume}{36}, \bibinfo{number}{6}
  (\bibinfo{year}{2016}), \bibinfo{pages}{1184--1192}.
\newblock


\bibitem[\protect\citeauthoryear{Hesamifard, Takabi, Ghasemi, and
  Wright}{Hesamifard et~al\mbox{.}}{2018}]%
        {Hesamifard2018}
\bibfield{author}{\bibinfo{person}{Ehsan Hesamifard}, \bibinfo{person}{Hassan
  Takabi}, \bibinfo{person}{Mehdi Ghasemi}, {and} \bibinfo{person}{Rebecca~N.
  Wright}.} \bibinfo{year}{2018}\natexlab{}.
\newblock \showarticletitle{{Privacy-preserving Machine Learning as a
  Service}}. In \bibinfo{booktitle}{\emph{Proceedings on Privacy Enhancing
  Technologies (PETS'19)}}. \bibinfo{pages}{123--142}.
\newblock


\bibitem[\protect\citeauthoryear{Hitaj, Ateniese, and Perez-Cruz}{Hitaj
  et~al\mbox{.}}{2017}]%
        {Hitaj2017}
\bibfield{author}{\bibinfo{person}{Briland Hitaj}, \bibinfo{person}{Giuseppe
  Ateniese}, {and} \bibinfo{person}{Fernando Perez-Cruz}.}
  \bibinfo{year}{2017}\natexlab{}.
\newblock \showarticletitle{{Deep Models under the GAN: Information leakage
  from collaborative deep learning}}. In \bibinfo{booktitle}{\emph{Proceedings
  of the ACM Conference on Computer and Communications Security (CCS'17)}}.
  \bibinfo{publisher}{ACM}, \bibinfo{pages}{603--618}.
\newblock
\showISBNx{9781450349468}
\showISSN{15437221}
\urldef\tempurl%
\url{https://doi.org/10.1145/3133956.3134012}
\showDOI{\tempurl}


\bibitem[\protect\citeauthoryear{Hua, Zhang, and Suh}{Hua
  et~al\mbox{.}}{2018}]%
        {Hua2018}
\bibfield{author}{\bibinfo{person}{Weizhe Hua}, \bibinfo{person}{Zhiru Zhang},
  {and} \bibinfo{person}{G~Edward Suh}.} \bibinfo{year}{2018}\natexlab{}.
\newblock \showarticletitle{{Reverse engineering convolutional neural networks
  through side-channel information leaks}}. In
  \bibinfo{booktitle}{\emph{Proceedings of the Design Automation Conference
  (DAC'18)}}, Vol.~\bibinfo{volume}{Part F1377}. \bibinfo{pages}{1--6}.
\newblock
\showISBNx{9781450357005}
\showISSN{0738100X}
\urldef\tempurl%
\url{https://doi.org/10.1145/3195970.3196105}
\showDOI{\tempurl}


\bibitem[\protect\citeauthoryear{Huai, Wang, Miao, Xu, and Zhang}{Huai
  et~al\mbox{.}}{2019}]%
        {Huai2019}
\bibfield{author}{\bibinfo{person}{Mengdi Huai}, \bibinfo{person}{Di Wang},
  \bibinfo{person}{Chenglin Miao}, \bibinfo{person}{Jinhui Xu}, {and}
  \bibinfo{person}{Aidong Zhang}.} \bibinfo{year}{2019}\natexlab{}.
\newblock \showarticletitle{{Privacy-aware Synthesizing for Crowdsourced
  Data}}. In \bibinfo{booktitle}{\emph{Proceedings of the Twenty-Eighth
  International Joint Conference on Artificial Intelligence}},
  Vol.~\bibinfo{volume}{2019-Augus}. \bibinfo{publisher}{International Joint
  Conferences on Artificial Intelligence Organization},
  \bibinfo{address}{California}, \bibinfo{pages}{2542--2548}.
\newblock
\showISBNx{978-0-9992411-4-1}
\showISSN{10450823}


\bibitem[\protect\citeauthoryear{Hussien, Hamza, and Hefny}{Hussien
  et~al\mbox{.}}{2013}]%
        {Hussien2013}
\bibfield{author}{\bibinfo{person}{Abou-el-ela~Abdou Hussien},
  \bibinfo{person}{Nermin Hamza}, {and} \bibinfo{person}{Hesham~A Hefny}.}
  \bibinfo{year}{2013}\natexlab{}.
\newblock \showarticletitle{{Attacks on Anonymization-Based Privacy-Preserving:
  A Survey for Data Mining and Data Publishing}}.
\newblock \bibinfo{journal}{\emph{Journal of Information Security}}
  \bibinfo{volume}{04}, \bibinfo{number}{02} (\bibinfo{year}{2013}),
  \bibinfo{pages}{101--112}.
\newblock
\showISSN{2153-1234}
\urldef\tempurl%
\url{https://doi.org/10.4236/jis.2013.42012}
\showDOI{\tempurl}


\bibitem[\protect\citeauthoryear{Imtia and Sarwate}{Imtia and Sarwate}{2018}]%
        {Imtia2018tensor}
\bibfield{author}{\bibinfo{person}{Hafiz Imtia} {and} \bibinfo{person}{Anand~D.
  Sarwate}.} \bibinfo{year}{2018}\natexlab{}.
\newblock \showarticletitle{{Improved Algorithms for Differentially Private
  Orthogonal Tensor Decomposition}}. In \bibinfo{booktitle}{\emph{Proceedings
  of IEEE International Conference on Acoustics, Speech and Signal Processing
  (ICASSP'18)}}. \bibinfo{publisher}{IEEE}, \bibinfo{pages}{2201--2205}.
\newblock
\showISBNx{9781538646588}
\showISSN{15206149}
\urldef\tempurl%
\url{https://doi.org/10.1109/ICASSP.2018.8461303}
\showDOI{\tempurl}


\bibitem[\protect\citeauthoryear{Imtiaz and Sarwate}{Imtiaz and
  Sarwate}{2018}]%
        {Imtiaz2018}
\bibfield{author}{\bibinfo{person}{Hafiz Imtiaz} {and}
  \bibinfo{person}{Anand~D. Sarwate}.} \bibinfo{year}{2018}\natexlab{}.
\newblock \showarticletitle{{Distributed Differentially Private Algorithms for
  Matrix and Tensor Factorization}}.
\newblock \bibinfo{journal}{\emph{IEEE J. Sel. Topics Signal Process.}}
  \bibinfo{volume}{12}, \bibinfo{number}{6} (\bibinfo{year}{2018}),
  \bibinfo{pages}{1449--1464}.
\newblock
\showISSN{19324553}
\urldef\tempurl%
\url{https://doi.org/10.1109/JSTSP.2018.2877842}
\showDOI{\tempurl}


\bibitem[\protect\citeauthoryear{Iyengar, Near, Song, Thakkar, Thakurta, and
  Wang}{Iyengar et~al\mbox{.}}{2019}]%
        {Iyengar2019}
\bibfield{author}{\bibinfo{person}{Roger Iyengar}, \bibinfo{person}{Joseph~P.
  Near}, \bibinfo{person}{Dawn Song}, \bibinfo{person}{Om Thakkar},
  \bibinfo{person}{Abhradeep Thakurta}, {and} \bibinfo{person}{Lun Wang}.}
  \bibinfo{year}{2019}\natexlab{}.
\newblock \showarticletitle{{Towards practical differentially private convex
  optimization}}. In \bibinfo{booktitle}{\emph{Proceedings of the IEEE
  Symposium on Security and Privacy (SP'19).}},
  Vol.~\bibinfo{volume}{2019-May}. \bibinfo{publisher}{IEEE},
  \bibinfo{pages}{299--316}.
\newblock
\showISBNx{9781538666609}
\showISSN{10816011}
\urldef\tempurl%
\url{https://doi.org/10.1109/SP.2019.00001}
\showDOI{\tempurl}


\bibitem[\protect\citeauthoryear{Jagannathan and Wright}{Jagannathan and
  Wright}{2005}]%
        {Jagannathan2005}
\bibfield{author}{\bibinfo{person}{Geetha Jagannathan} {and}
  \bibinfo{person}{Rebecca~N Wright}.} \bibinfo{year}{2005}\natexlab{}.
\newblock \showarticletitle{{Privacy-preserving distributed k-means clustering
  over arbitrarily partitioned data}}. In \bibinfo{booktitle}{\emph{Proceedings
  of the ACM International Conference on Knowledge Discovery and Data Mining
  (KDD'05)}}. \bibinfo{pages}{593--599}.
\newblock
\urldef\tempurl%
\url{https://doi.org/10.1145/1081870.1081942}
\showDOI{\tempurl}


\bibitem[\protect\citeauthoryear{Jahanbekam, Bauckhage, and Thurau}{Jahanbekam
  et~al\mbox{.}}{2010}]%
        {Jahanbekam2010}
\bibfield{author}{\bibinfo{person}{Amirhossein Jahanbekam},
  \bibinfo{person}{Christian Bauckhage}, {and} \bibinfo{person}{Christian
  Thurau}.} \bibinfo{year}{2010}\natexlab{}.
\newblock \showarticletitle{{Age recognition in the wild}}. In
  \bibinfo{booktitle}{\emph{Proceedings of the International Conference on
  Pattern Recognition (ICPR'10)}}. \bibinfo{pages}{392--395}.
\newblock
\showISBNx{9780769541099}
\showISSN{10514651}
\urldef\tempurl%
\url{https://doi.org/10.1109/ICPR.2010.104}
\showDOI{\tempurl}


\bibitem[\protect\citeauthoryear{Ji, Lipton, and Elkan}{Ji
  et~al\mbox{.}}{2014}]%
        {ji2014}
\bibfield{author}{\bibinfo{person}{Zhanglong Ji}, \bibinfo{person}{Zachary~C
  Lipton}, {and} \bibinfo{person}{Charles Elkan}.}
  \bibinfo{year}{2014}\natexlab{}.
\newblock \showarticletitle{{Differential Privacy and Machine Learning: a
  Survey and Review}}.
\newblock \bibinfo{journal}{\emph{arXiv:1412.7584}} (\bibinfo{year}{2014}).
\newblock
\urldef\tempurl%
\url{http://arxiv.org/abs/1412.7584}
\showURL{%
\tempurl}


\bibitem[\protect\citeauthoryear{Jia and Gong}{Jia and Gong}{2018}]%
        {Jia2018}
\bibfield{author}{\bibinfo{person}{Jinyuan Jia} {and}
  \bibinfo{person}{Neil~Zhenqiang Gong}.} \bibinfo{year}{2018}\natexlab{}.
\newblock \showarticletitle{{AttriGuard: A practical defense against attribute
  inference attacks via adversarial machine learning}}. In
  \bibinfo{booktitle}{\emph{Proceedings of the 27th USENIX Security Symposium
  (USENIX'18)}}. \bibinfo{pages}{513--529}.
\newblock
\showISBNx{9781939133045}


\bibitem[\protect\citeauthoryear{Jia, Guo, Jin, and Fang}{Jia
  et~al\mbox{.}}{2018}]%
        {Jia2018a}
\bibfield{author}{\bibinfo{person}{Qi Jia}, \bibinfo{person}{Linke Guo},
  \bibinfo{person}{Zhanpeng Jin}, {and} \bibinfo{person}{Yuguang Fang}.}
  \bibinfo{year}{2018}\natexlab{}.
\newblock \showarticletitle{{Preserving model privacy for machine learning in
  distributed systems}}.
\newblock \bibinfo{journal}{\emph{IEEE Trans. Parallel Distrib. Syst.}}
  \bibinfo{volume}{29}, \bibinfo{number}{8} (\bibinfo{year}{2018}),
  \bibinfo{pages}{1808--1822}.
\newblock
\showISSN{10459219}
\urldef\tempurl%
\url{https://doi.org/10.1109/TPDS.2018.2809624}
\showDOI{\tempurl}


\bibitem[\protect\citeauthoryear{{Joon Oh}, Benenson, Fritz, and Schiele}{{Joon
  Oh} et~al\mbox{.}}{2015}]%
        {JoonOh2015}
\bibfield{author}{\bibinfo{person}{Seong {Joon Oh}}, \bibinfo{person}{Rodrigo
  Benenson}, \bibinfo{person}{Mario Fritz}, {and} \bibinfo{person}{Bernt
  Schiele}.} \bibinfo{year}{2015}\natexlab{}.
\newblock \showarticletitle{{Person recognition in personal photo
  collections}}. In \bibinfo{booktitle}{\emph{Proceedings of the IEEE
  International Conference on Computer Vision (CVPR'15)}}.
  \bibinfo{pages}{3862--3870}.
\newblock


\bibitem[\protect\citeauthoryear{Kairouz, McMahan, Avent, Bellet, Bennis,
  Bhagoji, Bonawitz, Charles, Cormode, Cummings, D'Oliveira, Rouayheb, Evans,
  Gardner, Garrett, Gasc{\'{o}}n, Ghazi, Gibbons, Gruteser, Harchaoui, He, He,
  Huo, Hutchinson, Hsu, Jaggi, Javidi, Joshi, Khodak, Kone{\v{c}}n{\'{y}},
  Korolova, Koushanfar, Koyejo, Lepoint, Liu, Mittal, Mohri, Nock,
  {\"{O}}zg{\"{u}}r, Pagh, Raykova, Qi, Ramage, Raskar, Song, Song, Stich, Sun,
  Suresh, Tram{\`{e}}r, Vepakomma, Wang, Xiong, Xu, Yang, Yu, Yu, and
  Zhao}{Kairouz et~al\mbox{.}}{2019}]%
        {Kairouz2019}
\bibfield{author}{\bibinfo{person}{Peter Kairouz}, \bibinfo{person}{H.~Brendan
  McMahan}, \bibinfo{person}{Brendan Avent}, \bibinfo{person}{Aur{\'{e}}lien
  Bellet}, \bibinfo{person}{Mehdi Bennis}, \bibinfo{person}{Arjun~Nitin
  Bhagoji}, \bibinfo{person}{Keith Bonawitz}, \bibinfo{person}{Zachary
  Charles}, \bibinfo{person}{Graham Cormode}, \bibinfo{person}{Rachel
  Cummings}, \bibinfo{person}{Rafael G.~L. D'Oliveira},
  \bibinfo{person}{Salim~El Rouayheb}, \bibinfo{person}{David Evans},
  \bibinfo{person}{Josh Gardner}, \bibinfo{person}{Zachary Garrett},
  \bibinfo{person}{Adri{\`{a}} Gasc{\'{o}}n}, \bibinfo{person}{Badih Ghazi},
  \bibinfo{person}{Phillip~B. Gibbons}, \bibinfo{person}{Marco Gruteser},
  \bibinfo{person}{Zaid Harchaoui}, \bibinfo{person}{Chaoyang He},
  \bibinfo{person}{Lie He}, \bibinfo{person}{Zhouyuan Huo},
  \bibinfo{person}{Ben Hutchinson}, \bibinfo{person}{Justin Hsu},
  \bibinfo{person}{Martin Jaggi}, \bibinfo{person}{Tara Javidi},
  \bibinfo{person}{Gauri Joshi}, \bibinfo{person}{Mikhail Khodak},
  \bibinfo{person}{Jakub Kone{\v{c}}n{\'{y}}}, \bibinfo{person}{Aleksandra
  Korolova}, \bibinfo{person}{Farinaz Koushanfar}, \bibinfo{person}{Sanmi
  Koyejo}, \bibinfo{person}{Tancr{\`{e}}de Lepoint}, \bibinfo{person}{Yang
  Liu}, \bibinfo{person}{Prateek Mittal}, \bibinfo{person}{Mehryar Mohri},
  \bibinfo{person}{Richard Nock}, \bibinfo{person}{Ayfer {\"{O}}zg{\"{u}}r},
  \bibinfo{person}{Rasmus Pagh}, \bibinfo{person}{Mariana Raykova},
  \bibinfo{person}{Hang Qi}, \bibinfo{person}{Daniel Ramage},
  \bibinfo{person}{Ramesh Raskar}, \bibinfo{person}{Dawn Song},
  \bibinfo{person}{Weikang Song}, \bibinfo{person}{Sebastian~U. Stich},
  \bibinfo{person}{Ziteng Sun}, \bibinfo{person}{Ananda~Theertha Suresh},
  \bibinfo{person}{Florian Tram{\`{e}}r}, \bibinfo{person}{Praneeth Vepakomma},
  \bibinfo{person}{Jianyu Wang}, \bibinfo{person}{Li Xiong},
  \bibinfo{person}{Zheng Xu}, \bibinfo{person}{Qiang Yang},
  \bibinfo{person}{Felix~X. Yu}, \bibinfo{person}{Han Yu}, {and}
  \bibinfo{person}{Sen Zhao}.} \bibinfo{year}{2019}\natexlab{}.
\newblock \showarticletitle{{Advances and Open Problems in Federated
  Learning}}.
\newblock  (\bibinfo{year}{2019}).
\newblock
\showeprint[arxiv]{1912.04977}
\urldef\tempurl%
\url{http://arxiv.org/abs/1912.04977}
\showURL{%
\tempurl}


\bibitem[\protect\citeauthoryear{Kone{\v{c}}n{\'{y}}, McMahan, Yu,
  Richt{\'{a}}rik, Suresh, and Bacon}{Kone{\v{c}}n{\'{y}}
  et~al\mbox{.}}{2016}]%
        {Konecny2016}
\bibfield{author}{\bibinfo{person}{Jakub Kone{\v{c}}n{\'{y}}},
  \bibinfo{person}{H~Brendan McMahan}, \bibinfo{person}{Felix~X Yu},
  \bibinfo{person}{Peter Richt{\'{a}}rik}, \bibinfo{person}{Ananda~Theertha
  Suresh}, {and} \bibinfo{person}{Dave Bacon}.}
  \bibinfo{year}{2016}\natexlab{}.
\newblock \showarticletitle{{Federated Learning: Strategies for Improving
  Communication Efficiency}}.
\newblock \bibinfo{journal}{\emph{arXiv:1610.05492}} (\bibinfo{year}{2016}).
\newblock
\urldef\tempurl%
\url{http://arxiv.org/abs/1610.05492}
\showURL{%
\tempurl}


\bibitem[\protect\citeauthoryear{Krumm}{Krumm}{2007}]%
        {Krumm2007}
\bibfield{author}{\bibinfo{person}{John Krumm}.}
  \bibinfo{year}{2007}\natexlab{}.
\newblock \showarticletitle{Inference attacks on location tracks}. In
  \bibinfo{booktitle}{\emph{International Conference on Pervasive Computing}}.
  Springer, \bibinfo{pages}{127--143}.
\newblock


\bibitem[\protect\citeauthoryear{Kulkarni, Tagasovska, Vatter, and
  Garbinato}{Kulkarni et~al\mbox{.}}{2018}]%
        {Kulkarni2018}
\bibfield{author}{\bibinfo{person}{Vaibhav Kulkarni}, \bibinfo{person}{Natasa
  Tagasovska}, \bibinfo{person}{Thibault Vatter}, {and} \bibinfo{person}{Benoit
  Garbinato}.} \bibinfo{year}{2018}\natexlab{}.
\newblock \showarticletitle{{Generative Models for Simulating Mobility
  Trajectories}}.
\newblock \bibinfo{journal}{\emph{arXiv:1811.12801}} (\bibinfo{year}{2018}).
\newblock
\urldef\tempurl%
\url{http://arxiv.org/abs/1811.12801}
\showURL{%
\tempurl}


\bibitem[\protect\citeauthoryear{Kurakin, Goodfellow, and Bengio}{Kurakin
  et~al\mbox{.}}{2019}]%
        {Kurakin2019}
\bibfield{author}{\bibinfo{person}{Alexey Kurakin}, \bibinfo{person}{Ian~J.
  Goodfellow}, {and} \bibinfo{person}{Samy Bengio}.}
  \bibinfo{year}{2019}\natexlab{}.
\newblock \showarticletitle{{Adversarial machine learning at scale}}. In
  \bibinfo{booktitle}{\emph{Proceedings of the International Conference on
  Learning Representations (ICLR'19)}}.
\newblock


\bibitem[\protect\citeauthoryear{Lebanoff and Liu}{Lebanoff and Liu}{2018}]%
        {Lebanoff2018}
\bibfield{author}{\bibinfo{person}{Logan Lebanoff} {and} \bibinfo{person}{Fei
  Liu}.} \bibinfo{year}{2018}\natexlab{}.
\newblock \showarticletitle{{Automatic Detection of Vague Words and Sentences
  in Privacy Policies}}. In \bibinfo{booktitle}{\emph{Proceedings of the 2018
  Conference on Empirical Methods in Natural Language Processing}}.
  \bibinfo{publisher}{Association for Computational Linguistics},
  \bibinfo{address}{Stroudsburg, PA, USA}, \bibinfo{pages}{3508--3517}.
\newblock
\showISBNx{9781948087841}
\urldef\tempurl%
\url{https://doi.org/10.18653/v1/D18-1387}
\showDOI{\tempurl}


\bibitem[\protect\citeauthoryear{Lecuyer, Atlidakis, Geambasu, Hsu, and
  Jana}{Lecuyer et~al\mbox{.}}{2019}]%
        {lecuyer2019certified}
\bibfield{author}{\bibinfo{person}{Mathias Lecuyer}, \bibinfo{person}{Vaggelis
  Atlidakis}, \bibinfo{person}{Roxana Geambasu}, \bibinfo{person}{Daniel Hsu},
  {and} \bibinfo{person}{Suman Jana}.} \bibinfo{year}{2019}\natexlab{}.
\newblock \showarticletitle{Certified robustness to adversarial examples with
  differential privacy}. In \bibinfo{booktitle}{\emph{Proceedings of the IEEE
  Symposium on Security and Privacy (SP’19)}}. IEEE,
  \bibinfo{pages}{656--672}.
\newblock


\bibitem[\protect\citeauthoryear{Lee and Kobsa}{Lee and Kobsa}{2017}]%
        {Lee2017}
\bibfield{author}{\bibinfo{person}{Hosub Lee} {and} \bibinfo{person}{Alfred
  Kobsa}.} \bibinfo{year}{2017}\natexlab{}.
\newblock \showarticletitle{{Privacy preference modeling and prediction in a
  simulated campuswide IoT environment}}. In
  \bibinfo{booktitle}{\emph{Proceedings of the IEEE International Conference on
  Pervasive Computing and Communications (PerCom'17)}}.
  \bibinfo{pages}{276--285}.
\newblock


\bibitem[\protect\citeauthoryear{Li, Zhu, Du, Liang, and Shen}{Li
  et~al\mbox{.}}{2018}]%
        {Li2018}
\bibfield{author}{\bibinfo{person}{Huaxin Li}, \bibinfo{person}{Haojin Zhu},
  \bibinfo{person}{Suguo Du}, \bibinfo{person}{Xiaohui Liang}, {and}
  \bibinfo{person}{Xuemin~Sherman Shen}.} \bibinfo{year}{2018}\natexlab{}.
\newblock \showarticletitle{{Privacy leakage of location sharing in mobile
  social networks: Attacks and defense}}.
\newblock \bibinfo{journal}{\emph{IEEE Trans. Dependable Secure Comput.}}
  \bibinfo{volume}{15}, \bibinfo{number}{4} (\bibinfo{year}{2018}),
  \bibinfo{pages}{646--660}.
\newblock
\showISSN{19410018}
\urldef\tempurl%
\url{https://doi.org/10.1109/TDSC.2016.2604383}
\showDOI{\tempurl}


\bibitem[\protect\citeauthoryear{Li, Li, Huang, Li, Gao, Yiu, and Chen}{Li
  et~al\mbox{.}}{2017}]%
        {Li2017}
\bibfield{author}{\bibinfo{person}{Ping Li}, \bibinfo{person}{Jin Li},
  \bibinfo{person}{Zhengan Huang}, \bibinfo{person}{Tong Li},
  \bibinfo{person}{Chong~Zhi Gao}, \bibinfo{person}{Siu~Ming Yiu}, {and}
  \bibinfo{person}{Kai Chen}.} \bibinfo{year}{2017}\natexlab{}.
\newblock \showarticletitle{{Multi-key privacy-preserving deep learning in
  cloud computing}}.
\newblock \bibinfo{journal}{\emph{Future Generation Computer Systems}}
  \bibinfo{volume}{74} (\bibinfo{year}{2017}), \bibinfo{pages}{76--85}.
\newblock
\showISSN{0167739X}
\urldef\tempurl%
\url{https://doi.org/10.1016/j.future.2017.02.006}
\showDOI{\tempurl}


\bibitem[\protect\citeauthoryear{Li and Lin}{Li and Lin}{2019}]%
        {Li2019}
\bibfield{author}{\bibinfo{person}{Tao Li} {and} \bibinfo{person}{Lei Lin}.}
  \bibinfo{year}{2019}\natexlab{}.
\newblock \showarticletitle{{AnonymousNet: Natural face de-identification with
  measurable privacy}}.
\newblock \bibinfo{journal}{\emph{Proceedings of the IEEE Computer Society
  Conference on Computer Vision and Pattern Recognition Workshops (CVPRW'19))}}
   \bibinfo{volume}{2019-June} (\bibinfo{year}{2019}), \bibinfo{pages}{56--65}.
\newblock
\showISBNx{9781728125060}
\showISSN{21607516}
\urldef\tempurl%
\url{https://doi.org/10.1109/CVPRW.2019.00013}
\showDOI{\tempurl}


\bibitem[\protect\citeauthoryear{Li, Li, Wang, Zhang, and Gong}{Li
  et~al\mbox{.}}{2019}]%
        {li2019nattack}
\bibfield{author}{\bibinfo{person}{Yandong Li}, \bibinfo{person}{Lijun Li},
  \bibinfo{person}{Liqiang Wang}, \bibinfo{person}{Tong Zhang}, {and}
  \bibinfo{person}{Boqing Gong}.} \bibinfo{year}{2019}\natexlab{}.
\newblock \showarticletitle{NATTACK: Learning the Distributions of Adversarial
  Examples for an Improved Black-Box Attack on Deep Neural Networks}. In
  \bibinfo{booktitle}{\emph{Proceedings of the International Conference on
  Machine Learning (ICML'19)}}. \bibinfo{pages}{3866--3876}.
\newblock


\bibitem[\protect\citeauthoryear{Liu, Ding, Zhu, Xiang, and Zhou}{Liu
  et~al\mbox{.}}{2019}]%
        {Liu2019a}
\bibfield{author}{\bibinfo{person}{Bo Liu}, \bibinfo{person}{Ming Ding},
  \bibinfo{person}{Tianqing Zhu}, \bibinfo{person}{Yong Xiang}, {and}
  \bibinfo{person}{Wanlei Zhou}.} \bibinfo{year}{2019}\natexlab{}.
\newblock \showarticletitle{{Adversaries or allies? Privacy and deep learning
  in big data era}}. In \bibinfo{booktitle}{\emph{Concurrency Computation}},
  Vol.~\bibinfo{volume}{31}. \bibinfo{publisher}{Wiley Online Library},
  \bibinfo{pages}{e5102}.
\newblock
\showISSN{15320634}
\urldef\tempurl%
\url{https://doi.org/10.1002/cpe.5102}
\showDOI{\tempurl}


\bibitem[\protect\citeauthoryear{Liu, Zhou, Yu, Wang, Wang, Xiang, and Li}{Liu
  et~al\mbox{.}}{2017b}]%
        {Liu2017}
\bibfield{author}{\bibinfo{person}{Bo Liu}, \bibinfo{person}{Wanlei Zhou},
  \bibinfo{person}{Shui Yu}, \bibinfo{person}{Kun Wang}, \bibinfo{person}{Yu
  Wang}, \bibinfo{person}{Yong Xiang}, {and} \bibinfo{person}{Jin Li}.}
  \bibinfo{year}{2017}\natexlab{b}.
\newblock \showarticletitle{Home location protection in mobile social networks:
  a community based method (short paper)}. In
  \bibinfo{booktitle}{\emph{Proceedings of the International Conference on
  Information Security Practice and Experience (ISPEC'17)}}. Springer,
  \bibinfo{pages}{694--704}.
\newblock


\bibitem[\protect\citeauthoryear{Liu, Zhou, Zhu, Gao, Luan, and Zhou}{Liu
  et~al\mbox{.}}{2016}]%
        {Liu2016}
\bibfield{author}{\bibinfo{person}{Bo Liu}, \bibinfo{person}{Wanlei Zhou},
  \bibinfo{person}{Tianqing Zhu}, \bibinfo{person}{Longxiang Gao},
  \bibinfo{person}{Tom~H Luan}, {and} \bibinfo{person}{Haibo Zhou}.}
  \bibinfo{year}{2016}\natexlab{}.
\newblock \showarticletitle{{Silence is Golden: Enhancing Privacy of
  Location-Based Services by Content Broadcasting and Active Caching in
  Wireless Vehicular Networks}}.
\newblock \bibinfo{journal}{\emph{IEEE Trans. Veh. Technol.}}
  \bibinfo{volume}{65}, \bibinfo{number}{12} (\bibinfo{year}{2016}),
  \bibinfo{pages}{9942--9953}.
\newblock
\showISSN{00189545}
\urldef\tempurl%
\url{https://doi.org/10.1109/TVT.2016.2531185}
\showDOI{\tempurl}


\bibitem[\protect\citeauthoryear{Liu, Zhang, Zhou, Zhang, Fu, and
  Ramakrishnan}{Liu et~al\mbox{.}}{2014}]%
        {Liu2014}
\bibfield{author}{\bibinfo{person}{Hao Liu}, \bibinfo{person}{Yaoxue Zhang},
  \bibinfo{person}{Yuezhi Zhou}, \bibinfo{person}{Di Zhang},
  \bibinfo{person}{Xiaoming Fu}, {and} \bibinfo{person}{K~K Ramakrishnan}.}
  \bibinfo{year}{2014}\natexlab{}.
\newblock \showarticletitle{{Mining checkins from location-sharing services for
  client-independent IP geolocation}}. In \bibinfo{booktitle}{\emph{Proceedings
  of IEEE International Conference on Computer Communications (INFOCOM'14)}}.
  \bibinfo{pages}{619--627}.
\newblock
\showISBNx{9781479933600}
\showISSN{0743166X}
\urldef\tempurl%
\url{https://doi.org/10.1109/INFOCOM.2014.6847987}
\showDOI{\tempurl}


\bibitem[\protect\citeauthoryear{Liu, Li, and Gao}{Liu et~al\mbox{.}}{2018b}]%
        {Liu2018}
\bibfield{author}{\bibinfo{person}{Kin~Sum Liu}, \bibinfo{person}{Bo Li}, {and}
  \bibinfo{person}{Jie Gao}.} \bibinfo{year}{2018}\natexlab{b}.
\newblock \showarticletitle{{Generative model: Membership attack,
  generalization and diversity}}.
\newblock \bibinfo{journal}{\emph{CoRR, abs/1805.09898}}
  (\bibinfo{year}{2018}).
\newblock


\bibitem[\protect\citeauthoryear{Liu, Li, Zhao, Cai, Yu, and Leung}{Liu
  et~al\mbox{.}}{2018c}]%
        {Liu2018a}
\bibfield{author}{\bibinfo{person}{Qiang Liu}, \bibinfo{person}{Pan Li},
  \bibinfo{person}{Wentao Zhao}, \bibinfo{person}{Wei Cai},
  \bibinfo{person}{Shui Yu}, {and} \bibinfo{person}{Victor~C.M. Leung}.}
  \bibinfo{year}{2018}\natexlab{c}.
\newblock \showarticletitle{{A survey on security threats and defensive
  techniques of machine learning: A data driven view}}.
\newblock \bibinfo{journal}{\emph{IEEE Access}}  \bibinfo{volume}{6}
  (\bibinfo{year}{2018}), \bibinfo{pages}{12103--12117}.
\newblock
\showISSN{21693536}
\urldef\tempurl%
\url{https://doi.org/10.1109/ACCESS.2018.2805680}
\showDOI{\tempurl}


\bibitem[\protect\citeauthoryear{Liu, Chen, and Andris}{Liu
  et~al\mbox{.}}{2018a}]%
        {Liu2018b}
\bibfield{author}{\bibinfo{person}{Xi Liu}, \bibinfo{person}{Hanzhou Chen},
  {and} \bibinfo{person}{Clio Andris}.} \bibinfo{year}{2018}\natexlab{a}.
\newblock \showarticletitle{{trajGANs: Using generative adversarial networks
  for geo-privacy protection of trajectory data (Vision paper)}}. In
  \bibinfo{booktitle}{\emph{Location Privacy and Security Workshop}}.
  \bibinfo{pages}{1--7}.
\newblock


\bibitem[\protect\citeauthoryear{Liu, Ma, Aafer, Lee, Zhai, Wang, and
  Zhang}{Liu et~al\mbox{.}}{2018d}]%
        {Liu2018c}
\bibfield{author}{\bibinfo{person}{Yingqi Liu}, \bibinfo{person}{Shiqing Ma},
  \bibinfo{person}{Yousra Aafer}, \bibinfo{person}{Wen-Chuan Lee},
  \bibinfo{person}{Juan Zhai}, \bibinfo{person}{Weihang Wang}, {and}
  \bibinfo{person}{Xiangyu Zhang}.} \bibinfo{year}{2018}\natexlab{d}.
\newblock \showarticletitle{{Trojaning Attack on Neural Networks}}. In
  \bibinfo{booktitle}{\emph{Proceedings of Network and Distributed Systems
  Security Symposium (NDSS'18)}}.
\newblock
\showISBNx{1891562495}
\urldef\tempurl%
\url{https://doi.org/10.14722/ndss.2018.23291}
\showDOI{\tempurl}


\bibitem[\protect\citeauthoryear{Liu, Zhang, and Yu}{Liu
  et~al\mbox{.}}{2017a}]%
        {Liu2017a}
\bibfield{author}{\bibinfo{person}{Yujia Liu}, \bibinfo{person}{Weiming Zhang},
  {and} \bibinfo{person}{Nenghai Yu}.} \bibinfo{year}{2017}\natexlab{a}.
\newblock \showarticletitle{{Protecting Privacy in Shared Photos via
  Adversarial Examples Based Stealth}}.
\newblock \bibinfo{journal}{\emph{Security and Communication Networks}}
  \bibinfo{volume}{2017} (\bibinfo{year}{2017}).
\newblock
\showISSN{19390122}
\urldef\tempurl%
\url{https://doi.org/10.1155/2017/1897438}
\showDOI{\tempurl}


\bibitem[\protect\citeauthoryear{Long, Bindschaedler, Wang, Bu, Wang, Tang,
  Gunter, and Chen}{Long et~al\mbox{.}}{2018}]%
        {Long2018}
\bibfield{author}{\bibinfo{person}{Yunhui Long}, \bibinfo{person}{Vincent
  Bindschaedler}, \bibinfo{person}{Lei Wang}, \bibinfo{person}{Diyue Bu},
  \bibinfo{person}{Xiaofeng Wang}, \bibinfo{person}{Haixu Tang},
  \bibinfo{person}{Carl~A Gunter}, {and} \bibinfo{person}{Kai Chen}.}
  \bibinfo{year}{2018}\natexlab{}.
\newblock \showarticletitle{{Understanding Membership Inferences on
  Well-Generalized Learning Models}}.
\newblock \bibinfo{journal}{\emph{arXiv:1802.04889}} (\bibinfo{year}{2018}).
\newblock
\urldef\tempurl%
\url{http://arxiv.org/abs/1802.04889}
\showURL{%
\tempurl}


\bibitem[\protect\citeauthoryear{Lowd and Meek}{Lowd and Meek}{2005}]%
        {Lowd2005}
\bibfield{author}{\bibinfo{person}{Daniel Lowd} {and}
  \bibinfo{person}{Christopher Meek}.} \bibinfo{year}{2005}\natexlab{}.
\newblock \showarticletitle{{Adversarial learning}}. In
  \bibinfo{booktitle}{\emph{Proceedings of the ACM International Conference on
  Knowledge Discovery and Data Mining (KDD'05)}}. \bibinfo{pages}{641--647}.
\newblock
\urldef\tempurl%
\url{https://doi.org/10.1145/1081870.1081950}
\showDOI{\tempurl}


\bibitem[\protect\citeauthoryear{Mahmud, Nichols, and Drews}{Mahmud
  et~al\mbox{.}}{2014}]%
        {Mahmud2014}
\bibfield{author}{\bibinfo{person}{Jalal Mahmud}, \bibinfo{person}{Jeffrey
  Nichols}, {and} \bibinfo{person}{Clemens Drews}.}
  \bibinfo{year}{2014}\natexlab{}.
\newblock \showarticletitle{{Home location identification of twitter users}}.
\newblock \bibinfo{journal}{\emph{ACM Trans. Intell. Syst. Technol.}}
  \bibinfo{volume}{5}, \bibinfo{number}{3} (\bibinfo{year}{2014}),
  \bibinfo{pages}{47}.
\newblock
\showISSN{21576912}
\urldef\tempurl%
\url{https://doi.org/10.1145/2528548}
\showDOI{\tempurl}


\bibitem[\protect\citeauthoryear{Manek, Shenoy, Mohan, and Venugopal}{Manek
  et~al\mbox{.}}{2016}]%
        {Manek2016}
\bibfield{author}{\bibinfo{person}{Asha~S Manek}, \bibinfo{person}{P~Deepa
  Shenoy}, \bibinfo{person}{M~Chandra Mohan}, {and} \bibinfo{person}{K.R.
  Venugopal}.} \bibinfo{year}{2016}\natexlab{}.
\newblock \showarticletitle{{Detection of fraudulent and malicious websites by
  analysing user reviews for online shopping websites}}.
\newblock \bibinfo{journal}{\emph{International Journal of Knowledge and Web
  Intelligence}} \bibinfo{volume}{5}, \bibinfo{number}{3}
  (\bibinfo{year}{2016}), \bibinfo{pages}{171}.
\newblock
\showISSN{1755-8255}
\urldef\tempurl%
\url{https://doi.org/10.1504/ijkwi.2016.078712}
\showDOI{\tempurl}


\bibitem[\protect\citeauthoryear{McCrae, Costa, Terracciano, Parker, Mills, {De
  Fruyt}, and Mervielde}{McCrae et~al\mbox{.}}{2002}]%
        {McCrae2002}
\bibfield{author}{\bibinfo{person}{Robert~R McCrae}, \bibinfo{person}{Paul~T.
  Costa}, \bibinfo{person}{Antonio Terracciano}, \bibinfo{person}{Wayne~D
  Parker}, \bibinfo{person}{Carol~J Mills}, \bibinfo{person}{Filip {De Fruyt}},
  {and} \bibinfo{person}{Ivan Mervielde}.} \bibinfo{year}{2002}\natexlab{}.
\newblock \showarticletitle{{Personality trait development from age 12 to age
  18: Longitudinal, cross-sectional, and cross-cultural analyses}}.
\newblock \bibinfo{journal}{\emph{Journal of Personality and Social
  Psychology}} \bibinfo{volume}{83}, \bibinfo{number}{6}
  (\bibinfo{year}{2002}), \bibinfo{pages}{1456--1468}.
\newblock
\showISSN{00223514}
\urldef\tempurl%
\url{https://doi.org/10.1037/0022-3514.83.6.1456}
\showDOI{\tempurl}


\bibitem[\protect\citeauthoryear{McMahan, Moore, Ramage, Hampson, and
  Arcas}{McMahan et~al\mbox{.}}{2016}]%
        {McMahan2016}
\bibfield{author}{\bibinfo{person}{H.~Brendan McMahan}, \bibinfo{person}{Eider
  Moore}, \bibinfo{person}{Daniel Ramage}, \bibinfo{person}{Seth Hampson},
  {and} \bibinfo{person}{Blaise Ag{\"{u}}era~y Arcas}.}
  \bibinfo{year}{2016}\natexlab{}.
\newblock \showarticletitle{{Federated Learning of Deep Networks using Model
  Averaging}}.
\newblock \bibinfo{journal}{\emph{Arxiv}} \bibinfo{volume}{92},
  \bibinfo{number}{9} (\bibinfo{year}{2016}), \bibinfo{pages}{091118}.
\newblock
\showISBNx{0003-6951}
\showISSN{0003-6951}
\urldef\tempurl%
\url{https://doi.org/10.1063/1.2841713}
\showDOI{\tempurl}


\bibitem[\protect\citeauthoryear{McMahan, Ramage, Talwar, and Zhang}{McMahan
  et~al\mbox{.}}{2018}]%
        {McMahan2018}
\bibfield{author}{\bibinfo{person}{H~Brendan McMahan}, \bibinfo{person}{Daniel
  Ramage}, \bibinfo{person}{Kunal Talwar}, {and} \bibinfo{person}{Li Zhang}.}
  \bibinfo{year}{2018}\natexlab{}.
\newblock \showarticletitle{{Learning Differentially Private Recurrent Language
  Models Without Lossing Accuracy}}. In \bibinfo{booktitle}{\emph{Proceedings
  of the 6th International Conference on Learning Representations (ICLR'18)}},
  Vol.~\bibinfo{volume}{45}. \bibinfo{pages}{39--44}.
\newblock
\showISBNx{9781607419303}
\showISSN{00010782}
\urldef\tempurl%
\url{https://doi.org/10.1145/585597.585599}
\showDOI{\tempurl}


\bibitem[\protect\citeauthoryear{McPherson, Shokri, and Shmatikov}{McPherson
  et~al\mbox{.}}{2016}]%
        {McPherson2016}
\bibfield{author}{\bibinfo{person}{Richard McPherson}, \bibinfo{person}{Reza
  Shokri}, {and} \bibinfo{person}{Vitaly Shmatikov}.}
  \bibinfo{year}{2016}\natexlab{}.
\newblock \showarticletitle{{Defeating image obfuscation with deep learning}}.
\newblock \bibinfo{journal}{\emph{arXiv:1609.00408}} (\bibinfo{year}{2016}).
\newblock


\bibitem[\protect\citeauthoryear{Mehrpouyan, Azpiazu, and Pera}{Mehrpouyan
  et~al\mbox{.}}{2017}]%
        {Mehrpouyan2017}
\bibfield{author}{\bibinfo{person}{Hoda Mehrpouyan},
  \bibinfo{person}{Ion~Madrazo Azpiazu}, {and} \bibinfo{person}{Maria~Soledad
  Pera}.} \bibinfo{year}{2017}\natexlab{}.
\newblock \showarticletitle{{Measuring Personality for Automatic Elicitation of
  Privacy Preferences}}. In \bibinfo{booktitle}{\emph{Proceedings of the IEEE
  Symposium on Privacy-Aware Computing (PAC'17)}}. \bibinfo{pages}{84--95}.
\newblock


\bibitem[\protect\citeauthoryear{Melis, Song, De~Cristofaro, and
  Shmatikov}{Melis et~al\mbox{.}}{2019}]%
        {melis2019exploiting}
\bibfield{author}{\bibinfo{person}{Luca Melis}, \bibinfo{person}{Congzheng
  Song}, \bibinfo{person}{Emiliano De~Cristofaro}, {and}
  \bibinfo{person}{Vitaly Shmatikov}.} \bibinfo{year}{2019}\natexlab{}.
\newblock \showarticletitle{Exploiting unintended feature leakage in
  collaborative learning}. In \bibinfo{booktitle}{\emph{Proceedings of the IEEE
  Symposium on Security and Privacy (SP'19)}}. IEEE, \bibinfo{pages}{691--706}.
\newblock


\bibitem[\protect\citeauthoryear{Meng, Xing, Sheth, Weinsberg, and Lee}{Meng
  et~al\mbox{.}}{2014}]%
        {Meng2014}
\bibfield{author}{\bibinfo{person}{Wei Meng}, \bibinfo{person}{Xinyu Xing},
  \bibinfo{person}{Anmol Sheth}, \bibinfo{person}{Udi Weinsberg}, {and}
  \bibinfo{person}{Wenke Lee}.} \bibinfo{year}{2014}\natexlab{}.
\newblock \showarticletitle{{Your online interests: Pwned! a pollution attack
  against targeted advertising}}. In \bibinfo{booktitle}{\emph{Proceedings of
  the ACM Conference on Computer and Communications Security (CCS'14)}}.
  \bibinfo{pages}{129--140}.
\newblock


\bibitem[\protect\citeauthoryear{Mironov}{Mironov}{2017}]%
        {Mironov2017}
\bibfield{author}{\bibinfo{person}{Ilya Mironov}.}
  \bibinfo{year}{2017}\natexlab{}.
\newblock \showarticletitle{{R{\'{e}}nyi Differential Privacy}}. In
  \bibinfo{booktitle}{\emph{Proceedings of the IEEE Computer Security
  Foundations Symposium (CSF'17)}}. \bibinfo{publisher}{IEEE Computer Society},
  \bibinfo{pages}{263--275}.
\newblock
\showISBNx{9781538632161}
\showISSN{19401434}
\urldef\tempurl%
\url{https://doi.org/10.1109/CSF.2017.11}
\showDOI{\tempurl}


\bibitem[\protect\citeauthoryear{Mohassel and Zhang}{Mohassel and
  Zhang}{2017}]%
        {Mohassel2017}
\bibfield{author}{\bibinfo{person}{Payman Mohassel} {and}
  \bibinfo{person}{Yupeng Zhang}.} \bibinfo{year}{2017}\natexlab{}.
\newblock \showarticletitle{{SecureML: A system for scalable privacy-preserving
  machine learning}}. In \bibinfo{booktitle}{\emph{Proceedings of the IEEE
  Symposium on Security and Privacy (SP'17)}}. \bibinfo{pages}{19--38}.
\newblock


\bibitem[\protect\citeauthoryear{Moosavi-Dezfooli, Fawzi, Fawzi, and
  Frossard}{Moosavi-Dezfooli et~al\mbox{.}}{2017}]%
        {Moosavi-Dezfooli2017}
\bibfield{author}{\bibinfo{person}{Seyed~Mohsen Moosavi-Dezfooli},
  \bibinfo{person}{Alhussein Fawzi}, \bibinfo{person}{Omar Fawzi}, {and}
  \bibinfo{person}{Pascal Frossard}.} \bibinfo{year}{2017}\natexlab{}.
\newblock \showarticletitle{{Universal adversarial perturbations}}. In
  \bibinfo{booktitle}{\emph{Proceedings of the IEEE Conference on Computer
  Vision and Pattern Recognition (CVPR'17)}}. \bibinfo{pages}{86--94}.
\newblock
\showISBNx{9781538604571}
\urldef\tempurl%
\url{https://doi.org/10.1109/CVPR.2017.17}
\showDOI{\tempurl}


\bibitem[\protect\citeauthoryear{Moosavi-Dezfooli, Fawzi, and
  Frossard}{Moosavi-Dezfooli et~al\mbox{.}}{2016}]%
        {Moosavi-Dezfooli2016}
\bibfield{author}{\bibinfo{person}{Seyed-Mohsen Moosavi-Dezfooli},
  \bibinfo{person}{Alhussein Fawzi}, {and} \bibinfo{person}{Pascal Frossard}.}
  \bibinfo{year}{2016}\natexlab{}.
\newblock \showarticletitle{{Deepfool: a simple and accurate method to fool
  deep neural networks}}. In \bibinfo{booktitle}{\emph{Proceedings of the IEEE
  Conference on Computer Vision and Pattern Recognition (CVPR'16)}}.
  \bibinfo{pages}{2574--2582}.
\newblock


\bibitem[\protect\citeauthoryear{Narayanan and Shmatikov}{Narayanan and
  Shmatikov}{2006}]%
        {Narayanan2006}
\bibfield{author}{\bibinfo{person}{Arvind Narayanan} {and}
  \bibinfo{person}{Vitaly Shmatikov}.} \bibinfo{year}{2006}\natexlab{}.
\newblock \showarticletitle{{How To Break Anonymity of the Netflix Prize
  Dataset}}.
\newblock \bibinfo{journal}{\emph{cs/0610105}} (\bibinfo{year}{2006}).
\newblock
\urldef\tempurl%
\url{http://arxiv.org/abs/cs/0610105}
\showURL{%
\tempurl}


\bibitem[\protect\citeauthoryear{Nasr, Shokri, and Houmansadr}{Nasr
  et~al\mbox{.}}{2019}]%
        {Nasr2019}
\bibfield{author}{\bibinfo{person}{Milad Nasr}, \bibinfo{person}{Reza Shokri},
  {and} \bibinfo{person}{Amir Houmansadr}.} \bibinfo{year}{2019}\natexlab{}.
\newblock \showarticletitle{{Comprehensive privacy analysis of deep learning:
  Passive and active white-box inference attacks against centralized and
  federated learning}}. In \bibinfo{booktitle}{\emph{Proceedings of the IEEE
  Symposium on Security and Privacy (SP'19)}}. \bibinfo{pages}{739--753}.
\newblock
\showISBNx{9781538666609}
\showISSN{10816011}
\urldef\tempurl%
\url{https://doi.org/10.1109/SP.2019.00065}
\showDOI{\tempurl}


\bibitem[\protect\citeauthoryear{Neel, Roth, Vietri, and Wu}{Neel
  et~al\mbox{.}}{2019}]%
        {Neel2019}
\bibfield{author}{\bibinfo{person}{Seth Neel}, \bibinfo{person}{Aaron Roth},
  \bibinfo{person}{Giuseppe Vietri}, {and} \bibinfo{person}{Zhiwei~Steven Wu}.}
  \bibinfo{year}{2019}\natexlab{}.
\newblock \showarticletitle{{Oracle Efficient Private Non-Convex
  Optimization}}.
\newblock  (\bibinfo{year}{2019}).
\newblock
\showeprint[arxiv]{1909.01783}
\urldef\tempurl%
\url{http://arxiv.org/abs/1909.01783}
\showURL{%
\tempurl}


\bibitem[\protect\citeauthoryear{Nikolaenko, Weinsberg, Ioannidis, Joye, Boneh,
  and Taft}{Nikolaenko et~al\mbox{.}}{2013}]%
        {Nikolaenko2013}
\bibfield{author}{\bibinfo{person}{Valeria Nikolaenko}, \bibinfo{person}{Udi
  Weinsberg}, \bibinfo{person}{Stratis Ioannidis}, \bibinfo{person}{Marc Joye},
  \bibinfo{person}{Dan Boneh}, {and} \bibinfo{person}{Nina Taft}.}
  \bibinfo{year}{2013}\natexlab{}.
\newblock \showarticletitle{{Privacy-preserving ridge regression on hundreds of
  millions of records}}. In \bibinfo{booktitle}{\emph{Proceedings of the IEEE
  Symposium on Security and Privacy (SP'13)}}. \bibinfo{pages}{334--348}.
\newblock


\bibitem[\protect\citeauthoryear{Nugent}{Nugent}{2018}]%
        {Nugent2018}
\bibfield{author}{\bibinfo{person}{Ruair{\'\i} Nugent}.}
  \bibinfo{year}{2018}\natexlab{}.
\newblock \showarticletitle{{Assesing Completeness of Solvency and Financial
  Condition Reports through the use of Machine Learning and Text
  Classification}}.
\newblock  (\bibinfo{year}{2018}).
\newblock


\bibitem[\protect\citeauthoryear{Oh, Benenson, Fritz, and Schiele}{Oh
  et~al\mbox{.}}{2016}]%
        {Oh2016}
\bibfield{author}{\bibinfo{person}{Seong~Joon Oh}, \bibinfo{person}{Rodrigo
  Benenson}, \bibinfo{person}{Mario Fritz}, {and} \bibinfo{person}{Bernt
  Schiele}.} \bibinfo{year}{2016}\natexlab{}.
\newblock \showarticletitle{Faceless person recognition: Privacy implications
  in social media}. In \bibinfo{booktitle}{\emph{Proceedings of European
  Conference on Computer Vision (ECCV'16)}}. Springer, \bibinfo{pages}{19--35}.
\newblock


\bibitem[\protect\citeauthoryear{Oh, Fritz, and Schiele}{Oh
  et~al\mbox{.}}{2017}]%
        {oh2017adversarial}
\bibfield{author}{\bibinfo{person}{Seong~Joon Oh}, \bibinfo{person}{Mario
  Fritz}, {and} \bibinfo{person}{Bernt Schiele}.}
  \bibinfo{year}{2017}\natexlab{}.
\newblock \showarticletitle{Adversarial image perturbation for privacy
  protection a game theory perspective}. In
  \bibinfo{booktitle}{\emph{Proceedings of IEEE International Conference on
  Computer Vision (ICCV'17)}}. IEEE, \bibinfo{pages}{1491--1500}.
\newblock


\bibitem[\protect\citeauthoryear{Oh, Schiele, and Fritz}{Oh
  et~al\mbox{.}}{2019}]%
        {Oh2019}
\bibfield{author}{\bibinfo{person}{Seong~Joon Oh}, \bibinfo{person}{Bernt
  Schiele}, {and} \bibinfo{person}{Mario Fritz}.}
  \bibinfo{year}{2019}\natexlab{}.
\newblock \showarticletitle{{Towards reverse-engineering black-box neural
  networks}}.
\newblock In \bibinfo{booktitle}{\emph{Explainable AI: Interpreting, Explaining
  and Visualizing Deep Learning}}. \bibinfo{publisher}{Springer},
  \bibinfo{pages}{121--144}.
\newblock


\bibitem[\protect\citeauthoryear{Olejnik, Dacosta, Machado, Huguenin, Khan, and
  Hubaux}{Olejnik et~al\mbox{.}}{2017}]%
        {Olejnik2017}
\bibfield{author}{\bibinfo{person}{Katarzyna Olejnik}, \bibinfo{person}{Italo
  Dacosta}, \bibinfo{person}{Joana~Soares Machado}, \bibinfo{person}{Kevin
  Huguenin}, \bibinfo{person}{Mohammad~Emtiyaz Khan}, {and}
  \bibinfo{person}{Jean~Pierre Hubaux}.} \bibinfo{year}{2017}\natexlab{}.
\newblock \showarticletitle{{SmarPer: Context-Aware and Automatic
  Runtime-Permissions for Mobile Devices}}. In
  \bibinfo{booktitle}{\emph{Proceedings of the IEEE Symposium on Security and
  Privacy (SP'17)}}. \bibinfo{pages}{1058--1076}.
\newblock
\showISBNx{9781509055326}
\showISSN{10816011}
\urldef\tempurl%
\url{https://doi.org/10.1109/SP.2017.25}
\showDOI{\tempurl}


\bibitem[\protect\citeauthoryear{Orekondy, Fritz, and Schiele}{Orekondy
  et~al\mbox{.}}{2018}]%
        {Orekondy2018}
\bibfield{author}{\bibinfo{person}{Tribhuvanesh Orekondy},
  \bibinfo{person}{Mario Fritz}, {and} \bibinfo{person}{Bernt Schiele}.}
  \bibinfo{year}{2018}\natexlab{}.
\newblock \showarticletitle{{Connecting Pixels to Privacy and Utility:
  Automatic Redaction of Private Information in Images}}. In
  \bibinfo{booktitle}{\emph{Proceedings of the IEEE Computer Society Conference
  on Computer Vision and Pattern Recognition (CVPR'18)}}.
  \bibinfo{pages}{8466--8475}.
\newblock
\showISBNx{9781538664209}
\showISSN{10636919}
\urldef\tempurl%
\url{https://doi.org/10.1109/CVPR.2018.00883}
\showDOI{\tempurl}


\bibitem[\protect\citeauthoryear{Orekondy, Schiele, and Fritz}{Orekondy
  et~al\mbox{.}}{2017}]%
        {Orekondy2017}
\bibfield{author}{\bibinfo{person}{Tribhuvanesh Orekondy},
  \bibinfo{person}{Bernt Schiele}, {and} \bibinfo{person}{Mario Fritz}.}
  \bibinfo{year}{2017}\natexlab{}.
\newblock \showarticletitle{{Towards a Visual Privacy Advisor: Understanding
  and Predicting Privacy Risks in Images}}. In
  \bibinfo{booktitle}{\emph{Proceedings of the IEEE International Conference on
  Computer Vision (ICCV'17)}}. \bibinfo{pages}{3706--3715}.
\newblock
\showISBNx{9781538610329}
\showISSN{15505499}
\urldef\tempurl%
\url{https://doi.org/10.1109/ICCV.2017.398}
\showDOI{\tempurl}


\bibitem[\protect\citeauthoryear{Ouyang, Shokri, Rosenblum, and Yang}{Ouyang
  et~al\mbox{.}}{2018}]%
        {Ouyang2018}
\bibfield{author}{\bibinfo{person}{Kun Ouyang}, \bibinfo{person}{Reza Shokri},
  \bibinfo{person}{David~S Rosenblum}, {and} \bibinfo{person}{Wenzhuo Yang}.}
  \bibinfo{year}{2018}\natexlab{}.
\newblock \showarticletitle{{A non-parametric generative model for human
  trajectories}}. In \bibinfo{booktitle}{\emph{Proceedings of the International
  Joint Conference on Artificial Intelligence (IJCAI'18)}}.
  \bibinfo{pages}{3812--3817}.
\newblock
\showISBNx{9780999241127}
\showISSN{10450823}


\bibitem[\protect\citeauthoryear{Papernot, Goodfellow, Abadi, Talwar, and
  Erlingsson}{Papernot et~al\mbox{.}}{2019}]%
        {Papernot2019}
\bibfield{author}{\bibinfo{person}{Nicolas Papernot}, \bibinfo{person}{Ian
  Goodfellow}, \bibinfo{person}{Mart{\'{i}}n Abadi}, \bibinfo{person}{Kunal
  Talwar}, {and} \bibinfo{person}{{\'{U}}lfar Erlingsson}.}
  \bibinfo{year}{2019}\natexlab{}.
\newblock \showarticletitle{{Semi-supervised knowledge transfer for deep
  learning from private training data}}. In
  \bibinfo{booktitle}{\emph{Proceedings of the 5th International Conference on
  Learning Representations (ICLR'19)}}.
\newblock


\bibitem[\protect\citeauthoryear{Papernot, McDaniel, and Goodfellow}{Papernot
  et~al\mbox{.}}{2016a}]%
        {Papernot2016}
\bibfield{author}{\bibinfo{person}{Nicolas Papernot}, \bibinfo{person}{Patrick
  McDaniel}, {and} \bibinfo{person}{Ian Goodfellow}.}
  \bibinfo{year}{2016}\natexlab{a}.
\newblock \showarticletitle{{Transferability in Machine Learning: from
  Phenomena to Black-Box Attacks using Adversarial Samples}}.
\newblock \bibinfo{journal}{\emph{arXiv:1605.07277}} (\bibinfo{year}{2016}).
\newblock
\urldef\tempurl%
\url{http://arxiv.org/abs/1605.07277}
\showURL{%
\tempurl}


\bibitem[\protect\citeauthoryear{Papernot, McDaniel, Goodfellow, Jha, Celik,
  and Swami}{Papernot et~al\mbox{.}}{2017}]%
        {Papernot2017}
\bibfield{author}{\bibinfo{person}{Nicolas Papernot}, \bibinfo{person}{Patrick
  McDaniel}, \bibinfo{person}{Ian Goodfellow}, \bibinfo{person}{Somesh Jha},
  \bibinfo{person}{Z~Berkay Celik}, {and} \bibinfo{person}{Ananthram Swami}.}
  \bibinfo{year}{2017}\natexlab{}.
\newblock \showarticletitle{{Practical black-box attacks against machine
  learning}}. In \bibinfo{booktitle}{\emph{Proceedings of the ACM Asia
  Conference on Computer and Communications Security (ASIACCS'17)}}.
  \bibinfo{publisher}{ACM}, \bibinfo{pages}{506--519}.
\newblock
\showISBNx{9781450349444}
\urldef\tempurl%
\url{https://doi.org/10.1145/3052973.3053009}
\showDOI{\tempurl}


\bibitem[\protect\citeauthoryear{Papernot, McDaniel, Jha, Fredrikson, Celik,
  and Swami}{Papernot et~al\mbox{.}}{2016b}]%
        {Papernot2016a}
\bibfield{author}{\bibinfo{person}{Nicolas Papernot}, \bibinfo{person}{Patrick
  McDaniel}, \bibinfo{person}{Somesh Jha}, \bibinfo{person}{Matt Fredrikson},
  \bibinfo{person}{Z~Berkay Celik}, {and} \bibinfo{person}{Ananthram Swami}.}
  \bibinfo{year}{2016}\natexlab{b}.
\newblock \showarticletitle{{The limitations of deep learning in adversarial
  settings}}. In \bibinfo{booktitle}{\emph{Proceedings of the IEEE European
  Symposium on Security and Privacy (EuroS{\&}P'16)}}.
  \bibinfo{publisher}{IEEE}, \bibinfo{pages}{372--387}.
\newblock
\showISBNx{1509017526}


\bibitem[\protect\citeauthoryear{Park, Mohammadi, Gorde, Jajodia, Park, and
  Kim}{Park et~al\mbox{.}}{2018}]%
        {Park2018}
\bibfield{author}{\bibinfo{person}{Noseong Park}, \bibinfo{person}{Mahmoud
  Mohammadi}, \bibinfo{person}{Kshitij Gorde}, \bibinfo{person}{Sushil
  Jajodia}, \bibinfo{person}{Hongkyu Park}, {and} \bibinfo{person}{Youngmin
  Kim}.} \bibinfo{year}{2018}\natexlab{}.
\newblock \showarticletitle{{Data synthesis based on generative adversarial
  networks}}. In \bibinfo{booktitle}{\emph{Proceedings of the VLDB Endowment}},
  Vol.~\bibinfo{volume}{11}. \bibinfo{publisher}{VLDB Endowment},
  \bibinfo{pages}{1071--1083}.
\newblock
\showISSN{21508097}
\urldef\tempurl%
\url{https://doi.org/10.14778/3231751.3231757}
\showDOI{\tempurl}


\bibitem[\protect\citeauthoryear{Pathak, Rane, and Raj}{Pathak
  et~al\mbox{.}}{2010}]%
        {Pathak2010}
\bibfield{author}{\bibinfo{person}{Manas~A. Pathak}, \bibinfo{person}{Shantanu
  Rane}, {and} \bibinfo{person}{Bhiksha Raj}.} \bibinfo{year}{2010}\natexlab{}.
\newblock \showarticletitle{{Multiparty differential privacy via aggregation of
  locally trained classifiers}}. In \bibinfo{booktitle}{\emph{Advances in
  Neural Information Processing Systems (NIPS'10)}}.
  \bibinfo{pages}{1876--1884}.
\newblock
\showISBNx{9781617823800}


\bibitem[\protect\citeauthoryear{Pearson and Benameur}{Pearson and
  Benameur}{2010}]%
        {Pearson2010}
\bibfield{author}{\bibinfo{person}{Siani Pearson} {and}
  \bibinfo{person}{Azzedine Benameur}.} \bibinfo{year}{2010}\natexlab{}.
\newblock \showarticletitle{{Privacy, security and trust issues arising from
  cloud computing}}. In \bibinfo{booktitle}{\emph{Proceedings of the 2nd IEEE
  International Conference on Cloud Computing Technology and Science
  (CloudCom'10)}}. \bibinfo{pages}{693--702}.
\newblock
\showISBNx{9780769543024}
\urldef\tempurl%
\url{https://doi.org/10.1109/CloudCom.2010.66}
\showDOI{\tempurl}


\bibitem[\protect\citeauthoryear{Phan, Thai, Hu, Jin, Sun, and Dou}{Phan
  et~al\mbox{.}}{2020}]%
        {Phan2020}
\bibfield{author}{\bibinfo{person}{NhatHai Phan}, \bibinfo{person}{My~T. Thai},
  \bibinfo{person}{Han Hu}, \bibinfo{person}{Ruoming Jin},
  \bibinfo{person}{Tong Sun}, {and} \bibinfo{person}{Dejing Dou}.}
  \bibinfo{year}{2020}\natexlab{}.
\newblock \showarticletitle{{Scalable Differential Privacy with Certified
  Robustness in Adversarial Learning}}. In
  \bibinfo{booktitle}{\emph{Proceedings of the 37th International Conference on
  Machine Learning (PMLR'20)}}, Vol.~\bibinfo{volume}{6}.
\newblock
\showeprint[arxiv]{1903.09822}
\urldef\tempurl%
\url{http://arxiv.org/abs/1903.09822}
\showURL{%
\tempurl}


\bibitem[\protect\citeauthoryear{Phan, Wu, Hu, and Dou}{Phan
  et~al\mbox{.}}{2017}]%
        {Phan2017}
\bibfield{author}{\bibinfo{person}{Nhathai Phan}, \bibinfo{person}{Xintao Wu},
  \bibinfo{person}{Han Hu}, {and} \bibinfo{person}{Dejing Dou}.}
  \bibinfo{year}{2017}\natexlab{}.
\newblock \showarticletitle{{Adaptive laplace mechanism: Differential privacy
  preservation in deep learning}}. In \bibinfo{booktitle}{\emph{Proceedings of
  IEEE International Conference on Data Mining (ICDM'17)}},
  Vol.~\bibinfo{volume}{2017-November}. \bibinfo{publisher}{IEEE},
  \bibinfo{pages}{385--394}.
\newblock
\showISBNx{9781538638347}
\showISSN{15504786}
\urldef\tempurl%
\url{https://doi.org/10.1109/ICDM.2017.48}
\showDOI{\tempurl}


\bibitem[\protect\citeauthoryear{Phan, Vu, Liu, Jin, Dou, Wu, and Thai}{Phan
  et~al\mbox{.}}{2019}]%
        {Phan2019}
\bibfield{author}{\bibinfo{person}{Nhat~Hai Phan}, \bibinfo{person}{Minh~N.
  Vu}, \bibinfo{person}{Yang Liu}, \bibinfo{person}{Ruoming Jin},
  \bibinfo{person}{Dejing Dou}, \bibinfo{person}{Xintao Wu}, {and}
  \bibinfo{person}{My~T. Thai}.} \bibinfo{year}{2019}\natexlab{}.
\newblock \showarticletitle{{Heterogeneous Gaussian mechanism: Preserving
  differential privacy in deep learning with provable robustness}}. In
  \bibinfo{booktitle}{\emph{Proceedings of International Joint Conference on
  Artificial Intelligence (IJCAI'19)}}, Vol.~\bibinfo{volume}{2019-Augus}.
  \bibinfo{pages}{4753--4759}.
\newblock
\showISBNx{9780999241141}
\showISSN{10450823}
\urldef\tempurl%
\url{https://doi.org/10.24963/ijcai.2019/660}
\showDOI{\tempurl}
\showeprint[arxiv]{1906.01444}


\bibitem[\protect\citeauthoryear{Phong, Aono, Hayashi, Wang, and Moriai}{Phong
  et~al\mbox{.}}{2018}]%
        {Phong2018}
\bibfield{author}{\bibinfo{person}{Le~Trieu Phong}, \bibinfo{person}{Yoshinori
  Aono}, \bibinfo{person}{Takuya Hayashi}, \bibinfo{person}{Lihua Wang}, {and}
  \bibinfo{person}{Shiho Moriai}.} \bibinfo{year}{2018}\natexlab{}.
\newblock \showarticletitle{{Privacy-Preserving Deep Learning via Additively
  Homomorphic Encryption}}.
\newblock \bibinfo{journal}{\emph{IEEE Trans. Inf. Forensics Security}}
  \bibinfo{volume}{13}, \bibinfo{number}{5} (\bibinfo{year}{2018}),
  \bibinfo{pages}{1333--1345}.
\newblock
\showISSN{15566013}
\urldef\tempurl%
\url{https://doi.org/10.1109/TIFS.2017.2787987}
\showDOI{\tempurl}


\bibitem[\protect\citeauthoryear{Pinkas}{Pinkas}{2002}]%
        {Pinkas2002}
\bibfield{author}{\bibinfo{person}{Benny Pinkas}.}
  \bibinfo{year}{2002}\natexlab{}.
\newblock \showarticletitle{{Cryptographic techniques for privacy-preserving
  data mining}}.
\newblock \bibinfo{journal}{\emph{ACM SIGKDD Explorations Newsletter}}
  \bibinfo{volume}{4}, \bibinfo{number}{2} (\bibinfo{year}{2002}),
  \bibinfo{pages}{12--19}.
\newblock
\showISSN{19310145}
\urldef\tempurl%
\url{https://doi.org/10.1145/772862.772865}
\showDOI{\tempurl}


\bibitem[\protect\citeauthoryear{Poursaeed, Katsman, Gao, and
  Belongie}{Poursaeed et~al\mbox{.}}{2018}]%
        {Poursaeed2018}
\bibfield{author}{\bibinfo{person}{Omid Poursaeed}, \bibinfo{person}{Isay
  Katsman}, \bibinfo{person}{Bicheng Gao}, {and} \bibinfo{person}{Serge
  Belongie}.} \bibinfo{year}{2018}\natexlab{}.
\newblock \showarticletitle{{Generative Adversarial Perturbations}}. In
  \bibinfo{booktitle}{\emph{Proceedings of the IEEE Computer Society Conference
  on Computer Vision and Pattern Recognition (CVPR'18)}}.
  \bibinfo{pages}{4422--4431}.
\newblock
\showISBNx{9781538664209}
\showISSN{10636919}
\urldef\tempurl%
\url{https://doi.org/10.1109/CVPR.2018.00465}
\showDOI{\tempurl}


\bibitem[\protect\citeauthoryear{Rahimian, Orekondy, and Fritz}{Rahimian
  et~al\mbox{.}}{2020}]%
        {Rahimian2020}
\bibfield{author}{\bibinfo{person}{Shadi Rahimian},
  \bibinfo{person}{Tribhuvanesh Orekondy}, {and} \bibinfo{person}{Mario
  Fritz}.} \bibinfo{year}{2020}\natexlab{}.
\newblock \showarticletitle{{Sampling Attacks: Amplification of Membership
  Inference Attacks by Repeated Queries}}.
\newblock  (\bibinfo{year}{2020}).
\newblock
\showeprint[arxiv]{2009.00395}
\urldef\tempurl%
\url{http://arxiv.org/abs/2009.00395}
\showURL{%
\tempurl}


\bibitem[\protect\citeauthoryear{Reznichenko and Francis}{Reznichenko and
  Francis}{2014}]%
        {Reznichenko2014}
\bibfield{author}{\bibinfo{person}{Alexey Reznichenko} {and}
  \bibinfo{person}{Paul Francis}.} \bibinfo{year}{2014}\natexlab{}.
\newblock \showarticletitle{{Private-by-design advertising meets the real
  world}}. In \bibinfo{booktitle}{\emph{Proceedings of the ACM Conference on
  Computer and Communications Security (CCS'14)}}. \bibinfo{pages}{116--128}.
\newblock
\showISBNx{9781450329576}
\showISSN{15437221}
\urldef\tempurl%
\url{https://doi.org/10.1145/2660267.2660305}
\showDOI{\tempurl}


\bibitem[\protect\citeauthoryear{Rozsa, Rudd, and Boult}{Rozsa
  et~al\mbox{.}}{2016}]%
        {Rozsa2016}
\bibfield{author}{\bibinfo{person}{Andras Rozsa}, \bibinfo{person}{Ethan~M
  Rudd}, {and} \bibinfo{person}{Terrance~E Boult}.}
  \bibinfo{year}{2016}\natexlab{}.
\newblock \showarticletitle{{Adversarial Diversity and Hard Positive
  Generation}}. In \bibinfo{booktitle}{\emph{Proceedings of the IEEE Computer
  Society Conference on Computer Vision and Pattern Recognition Workshops
  (CVPW'16)}}. \bibinfo{pages}{410--417}.
\newblock
\showISBNx{9781467388504}
\showISSN{21607516}
\urldef\tempurl%
\url{https://doi.org/10.1109/CVPRW.2016.58}
\showDOI{\tempurl}


\bibitem[\protect\citeauthoryear{Rubinstein, Bartlett, Huang, and
  Taft}{Rubinstein et~al\mbox{.}}{2012}]%
        {Rubinstein2012}
\bibfield{author}{\bibinfo{person}{Benjamin I~P Rubinstein},
  \bibinfo{person}{Peter~L Bartlett}, \bibinfo{person}{Ling Huang}, {and}
  \bibinfo{person}{Nina Taft}.} \bibinfo{year}{2012}\natexlab{}.
\newblock \bibinfo{booktitle}{\emph{{Learning in a Large Function Space:
  Privacy-Preserving Mechanisms for SVM Learning}}}.
\newblock \bibinfo{type}{{T}echnical {R}eport}~1. \bibinfo{pages}{65--100}
  pages.
\newblock
\urldef\tempurl%
\url{http://repository.cmu.edu/jpc}
\showURL{%
\tempurl}


\bibitem[\protect\citeauthoryear{Salem, Wen, Backes, Ma, and Zhang}{Salem
  et~al\mbox{.}}{2020}]%
        {Salem2020}
\bibfield{author}{\bibinfo{person}{Ahmed Salem}, \bibinfo{person}{Rui Wen},
  \bibinfo{person}{Michael Backes}, \bibinfo{person}{Shiqing Ma}, {and}
  \bibinfo{person}{Yang Zhang}.} \bibinfo{year}{2020}\natexlab{}.
\newblock \showarticletitle{{Dynamic Backdoor Attacks Against Machine Learning
  Models}}.
\newblock  (\bibinfo{year}{2020}).
\newblock
\showeprint[arxiv]{2003.03675}
\urldef\tempurl%
\url{http://arxiv.org/abs/2003.03675}
\showURL{%
\tempurl}


\bibitem[\protect\citeauthoryear{Salem, Zhang, Humbert, Berrang, Fritz, and
  Backes}{Salem et~al\mbox{.}}{2019}]%
        {Salem2019}
\bibfield{author}{\bibinfo{person}{Ahmed Salem}, \bibinfo{person}{Yang Zhang},
  \bibinfo{person}{Mathias Humbert}, \bibinfo{person}{Pascal Berrang},
  \bibinfo{person}{Mario Fritz}, {and} \bibinfo{person}{Michael Backes}.}
  \bibinfo{year}{2019}\natexlab{}.
\newblock \showarticletitle{{ML-Leaks: Model and Data Independent Membership
  Inference Attacks and Defenses on Machine Learning Models}}. In
  \bibinfo{booktitle}{\emph{Proceedings of Network and Distributed Systems
  Security Symposium (NDSS'19)}}.
\newblock
\urldef\tempurl%
\url{https://doi.org/10.14722/ndss.2019.23119}
\showDOI{\tempurl}


\bibitem[\protect\citeauthoryear{Sanil, Karr, Lin, and Reiter}{Sanil
  et~al\mbox{.}}{2004}]%
        {Sanil2004}
\bibfield{author}{\bibinfo{person}{Ashish~P Sanil}, \bibinfo{person}{Alan~F
  Karr}, \bibinfo{person}{Xiaodong Lin}, {and} \bibinfo{person}{Jerome~P
  Reiter}.} \bibinfo{year}{2004}\natexlab{}.
\newblock \showarticletitle{{Privacy preserving regression modelling via
  distributed computation}}. In \bibinfo{booktitle}{\emph{Proceedings of the
  ACM SIGKDD International Conference on Knowledge Discovery and Data Mining
  (KDD'04)}}. \bibinfo{pages}{677--682}.
\newblock
\showISBNx{1581138881}
\urldef\tempurl%
\url{https://doi.org/10.1145/1014052.1014139}
\showDOI{\tempurl}


\bibitem[\protect\citeauthoryear{Schoppmann, Balle, Doerner, Zahur, and
  Evans}{Schoppmann et~al\mbox{.}}{2016}]%
        {Schoppmann2016}
\bibfield{author}{\bibinfo{person}{Phillipp Schoppmann}, \bibinfo{person}{Borja
  Balle}, \bibinfo{person}{Jack Doerner}, \bibinfo{person}{Samee Zahur}, {and}
  \bibinfo{person}{David Evans}.} \bibinfo{year}{2016}\natexlab{}.
\newblock \showarticletitle{{Secure Linear Regression on Vertically Partitioned
  Datasets}}.
\newblock \bibinfo{journal}{\emph{IACR Cryptology Eprint Archive}}
  \bibinfo{volume}{2016} (\bibinfo{year}{2016}), \bibinfo{pages}{1--27}.
\newblock


\bibitem[\protect\citeauthoryear{Sebastiani}{Sebastiani}{2002}]%
        {Sebastiani2002}
\bibfield{author}{\bibinfo{person}{Fabrizio Sebastiani}.}
  \bibinfo{year}{2002}\natexlab{}.
\newblock \showarticletitle{{Machine Learning in Automated Text
  Categorization}}.
\newblock \bibinfo{journal}{\emph{Comput. Surveys}} \bibinfo{volume}{34},
  \bibinfo{number}{1} (\bibinfo{year}{2002}), \bibinfo{pages}{1--47}.
\newblock
\showISSN{03600300}
\urldef\tempurl%
\url{https://doi.org/10.1145/505282.505283}
\showDOI{\tempurl}


\bibitem[\protect\citeauthoryear{Sharif, Bhagavatula, Bauer, and Reiter}{Sharif
  et~al\mbox{.}}{2016}]%
        {Sharif2016}
\bibfield{author}{\bibinfo{person}{Mahmood Sharif}, \bibinfo{person}{Sruti
  Bhagavatula}, \bibinfo{person}{Lujo Bauer}, {and} \bibinfo{person}{Michael~K
  Reiter}.} \bibinfo{year}{2016}\natexlab{}.
\newblock \showarticletitle{{Accessorize to a crime: Real and stealthy attacks
  on state-of-the-art face recognition}}. In
  \bibinfo{booktitle}{\emph{Proceedings of the ACM Conference on Computer and
  Communications Security (CCS'16)}}. \bibinfo{pages}{1528--1540}.
\newblock
\showISBNx{9781450341394}
\showISSN{15437221}
\urldef\tempurl%
\url{https://doi.org/10.1145/2976749.2978392}
\showDOI{\tempurl}


\bibitem[\protect\citeauthoryear{Shayegh and Ghanavati}{Shayegh and
  Ghanavati}{2017}]%
        {Shayegh2017}
\bibfield{author}{\bibinfo{person}{Parvaneh Shayegh} {and}
  \bibinfo{person}{Sepideh Ghanavati}.} \bibinfo{year}{2017}\natexlab{}.
\newblock \showarticletitle{{Toward an approach to privacy notices in IoT}}. In
  \bibinfo{booktitle}{\emph{Proceedings of the IEEE 25th International
  Requirements Engineering Conference Workshops (REW'17)}}.
  \bibinfo{pages}{104--110}.
\newblock
\showISBNx{9781538634882}
\urldef\tempurl%
\url{https://doi.org/10.1109/REW.2017.77}
\showDOI{\tempurl}


\bibitem[\protect\citeauthoryear{Shokri and Shmatikov}{Shokri and
  Shmatikov}{2015}]%
        {Shokri2015}
\bibfield{author}{\bibinfo{person}{Reza Shokri} {and} \bibinfo{person}{Vitaly
  Shmatikov}.} \bibinfo{year}{2015}\natexlab{}.
\newblock \showarticletitle{{Privacy-preserving deep learning}}. In
  \bibinfo{booktitle}{\emph{Proceedings of the ACM Conference on Computer and
  Communications Security (CCS'15)}}. \bibinfo{pages}{1310--1321}.
\newblock


\bibitem[\protect\citeauthoryear{Shokri, Stronati, Song, and Shmatikov}{Shokri
  et~al\mbox{.}}{2017}]%
        {Shokri2017}
\bibfield{author}{\bibinfo{person}{Reza Shokri}, \bibinfo{person}{Marco
  Stronati}, \bibinfo{person}{Congzheng Song}, {and} \bibinfo{person}{Vitaly
  Shmatikov}.} \bibinfo{year}{2017}\natexlab{}.
\newblock \showarticletitle{{Membership Inference Attacks Against Machine
  Learning Models}}. In \bibinfo{booktitle}{\emph{Proceedings of the IEEE
  Symposium on Security and Privacy (SP'17)}}. \bibinfo{publisher}{IEEE},
  \bibinfo{pages}{3--18}.
\newblock
\showISBNx{9781509055326}
\showISSN{10816011}
\urldef\tempurl%
\url{https://doi.org/10.1109/SP.2017.41}
\showDOI{\tempurl}


\bibitem[\protect\citeauthoryear{Shokri, Theodorakopoulos, Papadimitratos,
  Kazemi, and Hubaux}{Shokri et~al\mbox{.}}{2014}]%
        {Shokri2014}
\bibfield{author}{\bibinfo{person}{Reza Shokri}, \bibinfo{person}{George
  Theodorakopoulos}, \bibinfo{person}{Panos Papadimitratos},
  \bibinfo{person}{Ehsan Kazemi}, {and} \bibinfo{person}{Jean~Pierre Hubaux}.}
  \bibinfo{year}{2014}\natexlab{}.
\newblock \showarticletitle{{Hiding in the mobile crowd: Location privacy
  through collaboration}}.
\newblock \bibinfo{journal}{\emph{IEEE Trans. Dependable Secure Comput.}}
  \bibinfo{number}{3} (\bibinfo{year}{2014}), \bibinfo{pages}{266--279}.
\newblock
\showISSN{15455971}
\urldef\tempurl%
\url{https://doi.org/10.1109/TDSC.2013.57}
\showDOI{\tempurl}


\bibitem[\protect\citeauthoryear{Slavkovic, Nardi, and Tibbits}{Slavkovic
  et~al\mbox{.}}{2007}]%
        {Slavkovic2007}
\bibfield{author}{\bibinfo{person}{Aleksandra~B Slavkovic},
  \bibinfo{person}{Yuval Nardi}, {and} \bibinfo{person}{Matthew~M Tibbits}.}
  \bibinfo{year}{2007}\natexlab{}.
\newblock \showarticletitle{{Secure logistic regression of horizontally and
  vertically partitioned distributed databases}}. In
  \bibinfo{booktitle}{\emph{Proceedings of the IEEE International Conference on
  Data Mining (ICDM'07)}}. \bibinfo{pages}{723--728}.
\newblock
\showISBNx{0769530192}
\showISSN{15504786}
\urldef\tempurl%
\url{https://doi.org/10.1109/ICDMW.2007.114}
\showDOI{\tempurl}


\bibitem[\protect\citeauthoryear{Song, Ristenpart, and Shmatikov}{Song
  et~al\mbox{.}}{2017}]%
        {Song2017}
\bibfield{author}{\bibinfo{person}{Congzheng Song}, \bibinfo{person}{Thomas
  Ristenpart}, {and} \bibinfo{person}{Vitaly Shmatikov}.}
  \bibinfo{year}{2017}\natexlab{}.
\newblock \showarticletitle{{Machine learning models that remember too much}}.
  In \bibinfo{booktitle}{\emph{Proceedings of the ACM Conference on Computer
  and Communications Security (CCS'17)}}. \bibinfo{publisher}{ACM},
  \bibinfo{pages}{587--601}.
\newblock
\showISBNx{9781450349468}
\showISSN{15437221}
\urldef\tempurl%
\url{https://doi.org/10.1145/3133956.3134077}
\showDOI{\tempurl}


\bibitem[\protect\citeauthoryear{Song and Chai}{Song and Chai}{2018}]%
        {Song2018}
\bibfield{author}{\bibinfo{person}{Guocong Song} {and} \bibinfo{person}{Wei
  Chai}.} \bibinfo{year}{2018}\natexlab{}.
\newblock \showarticletitle{{Collaborative learning for deep neural networks}}.
  In \bibinfo{booktitle}{\emph{Advances in Neural Information Processing
  Systems (NIPS'18)}}, Vol.~\bibinfo{volume}{2018-Decem}.
  \bibinfo{pages}{1832--1841}.
\newblock
\showISSN{10495258}


\bibitem[\protect\citeauthoryear{Song, Chaudhuri, and Sarwate}{Song
  et~al\mbox{.}}{2013}]%
        {Song2013}
\bibfield{author}{\bibinfo{person}{Shuang Song}, \bibinfo{person}{Kamalika
  Chaudhuri}, {and} \bibinfo{person}{Anand~D Sarwate}.}
  \bibinfo{year}{2013}\natexlab{}.
\newblock \showarticletitle{{Stochastic gradient descent with differentially
  private updates}}. In \bibinfo{booktitle}{\emph{Proceedings of the IEEE
  Global Conference on Signal and Information Processing (GlobalSIP'13)}}.
  \bibinfo{pages}{245--248}.
\newblock
\showISBNx{9781479902484}
\urldef\tempurl%
\url{https://doi.org/10.1109/GlobalSIP.2013.6736861}
\showDOI{\tempurl}


\bibitem[\protect\citeauthoryear{Squicciarini, Caragea, and
  Balakavi}{Squicciarini et~al\mbox{.}}{2017}]%
        {Squicciarini2017}
\bibfield{author}{\bibinfo{person}{Anna Squicciarini},
  \bibinfo{person}{Cornelia Caragea}, {and} \bibinfo{person}{Rahul Balakavi}.}
  \bibinfo{year}{2017}\natexlab{}.
\newblock \showarticletitle{{Toward automated online photo privacy}}.
\newblock \bibinfo{journal}{\emph{ACM Trans. Web}} \bibinfo{volume}{11},
  \bibinfo{number}{1} (\bibinfo{year}{2017}), \bibinfo{pages}{2}.
\newblock
\showISSN{1559114X}
\urldef\tempurl%
\url{https://doi.org/10.1145/2983644}
\showDOI{\tempurl}


\bibitem[\protect\citeauthoryear{Squicciarini, Caragea, and
  Balakavi}{Squicciarini et~al\mbox{.}}{2014}]%
        {Squicciarini2014}
\bibfield{author}{\bibinfo{person}{Anna~C Squicciarini},
  \bibinfo{person}{Cornelia Caragea}, {and} \bibinfo{person}{Rahul Balakavi}.}
  \bibinfo{year}{2014}\natexlab{}.
\newblock \showarticletitle{{Analyzing images' privacy for the modern web}}. In
  \bibinfo{booktitle}{\emph{Proceedings of the 25th ACM Conference on Hypertext
  and Social Media (HT'14)}}. \bibinfo{pages}{136--147}.
\newblock
\showISBNx{9781450329545}
\urldef\tempurl%
\url{https://doi.org/10.1145/2631775.2631803}
\showDOI{\tempurl}


\bibitem[\protect\citeauthoryear{Sun, Ma, {Joon Oh}, Gool, Schiele, and
  Fritz}{Sun et~al\mbox{.}}{2018a}]%
        {sun2018natural}
\bibfield{author}{\bibinfo{person}{Qianru Sun}, \bibinfo{person}{Liqian Ma},
  \bibinfo{person}{Seong {Joon Oh}}, \bibinfo{person}{Luc~Van Gool},
  \bibinfo{person}{Bernt Schiele}, {and} \bibinfo{person}{Mario Fritz}.}
  \bibinfo{year}{2018}\natexlab{a}.
\newblock \showarticletitle{{Natural and Effective Obfuscation by Head
  Inpainting}}.
\newblock \bibinfo{journal}{\emph{Proceedings of the IEEE Computer Society
  Conference on Computer Vision and Pattern Recognition (CVPR'18)}}
  (\bibinfo{year}{2018}), \bibinfo{pages}{5050--5059}.
\newblock
\showISBNx{9781538664209}
\showISSN{10636919}
\urldef\tempurl%
\url{https://doi.org/10.1109/CVPR.2018.00530}
\showDOI{\tempurl}


\bibitem[\protect\citeauthoryear{Sun, Schiele, and Fritz}{Sun
  et~al\mbox{.}}{2017}]%
        {Sun2017}
\bibfield{author}{\bibinfo{person}{Qianru Sun}, \bibinfo{person}{Bernt
  Schiele}, {and} \bibinfo{person}{Mario Fritz}.}
  \bibinfo{year}{2017}\natexlab{}.
\newblock \showarticletitle{{A domain based approach to social relation
  recognition}}. In \bibinfo{booktitle}{\emph{Proceedings of the 30th IEEE
  Conference on Computer Vision and Pattern Recognition (CVPR'17)}}.
  \bibinfo{pages}{435--444}.
\newblock
\showISBNx{9781538604571}
\urldef\tempurl%
\url{https://doi.org/10.1109/CVPR.2017.54}
\showDOI{\tempurl}


\bibitem[\protect\citeauthoryear{Sun, Wu, and Hoi}{Sun et~al\mbox{.}}{2018b}]%
        {Sun2018}
\bibfield{author}{\bibinfo{person}{Xudong Sun}, \bibinfo{person}{Pengcheng Wu},
  {and} \bibinfo{person}{Steven~C.H. Hoi}.} \bibinfo{year}{2018}\natexlab{b}.
\newblock \showarticletitle{{Face detection using deep learning: An improved
  faster RCNN approach}}.
\newblock \bibinfo{journal}{\emph{Neurocomputing}}  \bibinfo{volume}{299}
  (\bibinfo{year}{2018}), \bibinfo{pages}{42--50}.
\newblock
\showISSN{18728286}
\urldef\tempurl%
\url{https://doi.org/10.1016/j.neucom.2018.03.030}
\showDOI{\tempurl}


\bibitem[\protect\citeauthoryear{Szegedy, Zaremba, Sutskever, Bruna, Erhan,
  Goodfellow, and Fergus}{Szegedy et~al\mbox{.}}{2014}]%
        {Szegedy2014}
\bibfield{author}{\bibinfo{person}{Christian Szegedy},
  \bibinfo{person}{Wojciech Zaremba}, \bibinfo{person}{Ilya Sutskever},
  \bibinfo{person}{Joan Bruna}, \bibinfo{person}{Dumitru Erhan},
  \bibinfo{person}{Ian Goodfellow}, {and} \bibinfo{person}{Rob Fergus}.}
  \bibinfo{year}{2014}\natexlab{}.
\newblock \showarticletitle{{Intriguing properties of neural networks}}. In
  \bibinfo{booktitle}{\emph{Proceedings of the International Conference on
  Learning Representations (ICLR'14)}}.
\newblock


\bibitem[\protect\citeauthoryear{Tesfay, Hofmann, Nakamura, Kiyomoto, and
  Serna}{Tesfay et~al\mbox{.}}{2018}]%
        {Tesfay2018}
\bibfield{author}{\bibinfo{person}{Welderufael~B Tesfay},
  \bibinfo{person}{Peter Hofmann}, \bibinfo{person}{Toru Nakamura},
  \bibinfo{person}{Shinsaku Kiyomoto}, {and} \bibinfo{person}{Jetzabel Serna}.}
  \bibinfo{year}{2018}\natexlab{}.
\newblock \showarticletitle{{I Read but Don't Agree: Privacy Policy
  Benchmarking using Machine Learning and the EU GDPR}}.
\newblock \bibinfo{journal}{\emph{Companion Proceedings of the The Web
  Conference 2018}}  \bibinfo{volume}{2} (\bibinfo{year}{2018}),
  \bibinfo{pages}{163--166}.
\newblock
\showISBNx{9781450356404}


\bibitem[\protect\citeauthoryear{Times}{Times}{2020}]%
        {Times2020}
\bibfield{author}{\bibinfo{person}{Financial Times}.}
  \bibinfo{year}{2020}\natexlab{}.
\newblock \showarticletitle{{Facebook privacy breach}}.
\newblock \bibinfo{journal}{\emph{Financial Times}} (\bibinfo{year}{2020}),
  \bibinfo{pages}{11--12}.
\newblock
\urldef\tempurl%
\url{https://www.ft.com/content/87184c40-2cfe-11e8-9b4b-bc4b9f08f381}
\showURL{%
\tempurl}


\bibitem[\protect\citeauthoryear{Tram{\`{e}}r, Zhang, Juels, Reiter, and
  Ristenpart}{Tram{\`{e}}r et~al\mbox{.}}{2016}]%
        {Tramer2016}
\bibfield{author}{\bibinfo{person}{Florian Tram{\`{e}}r}, \bibinfo{person}{Fan
  Zhang}, \bibinfo{person}{Ari Juels}, \bibinfo{person}{Michael~K Reiter},
  {and} \bibinfo{person}{Thomas Ristenpart}.} \bibinfo{year}{2016}\natexlab{}.
\newblock \showarticletitle{{Stealing machine learning models via prediction
  APIs}}. In \bibinfo{booktitle}{\emph{Proceedings of the 25th USENIX Security
  Symposium (USENIX'16)}}. \bibinfo{pages}{601--618}.
\newblock
\showISBNx{9781931971324}


\bibitem[\protect\citeauthoryear{Triastcyn and Faltings}{Triastcyn and
  Faltings}{2019}]%
        {Triastcyn2019}
\bibfield{author}{\bibinfo{person}{Aleksei Triastcyn} {and}
  \bibinfo{person}{Boi Faltings}.} \bibinfo{year}{2019}\natexlab{}.
\newblock \showarticletitle{{Generating artificial data for private deep
  learning}}. In \bibinfo{booktitle}{\emph{Proceedings of the 2019 CEUR
  Workshop}}, Vol.~\bibinfo{volume}{2335}. \bibinfo{pages}{33--40}.
\newblock
\showISSN{16130073}


\bibitem[\protect\citeauthoryear{Vaidya, Kantarcoglu, and Clifton}{Vaidya
  et~al\mbox{.}}{2008}]%
        {Vaidya2008}
\bibfield{author}{\bibinfo{person}{Jaideep Vaidya}, \bibinfo{person}{Murat
  Kantarcoglu}, {and} \bibinfo{person}{Chris Clifton}.}
  \bibinfo{year}{2008}\natexlab{}.
\newblock \showarticletitle{{Privacy-preserving naive bayes classification}}.
\newblock \bibinfo{journal}{\emph{The VLDB Journal}} \bibinfo{volume}{17},
  \bibinfo{number}{4} (\bibinfo{year}{2008}), \bibinfo{pages}{879--898}.
\newblock


\bibitem[\protect\citeauthoryear{van~den Oord, Dieleman, Zen, Simonyan,
  Vinyals, Graves, Kalchbrenner, Senior, and Kavukcuoglu}{van~den Oord
  et~al\mbox{.}}{2016}]%
        {Oord2016}
\bibfield{author}{\bibinfo{person}{Aaron van~den Oord}, \bibinfo{person}{Sander
  Dieleman}, \bibinfo{person}{Heiga Zen}, \bibinfo{person}{Karen Simonyan},
  \bibinfo{person}{Oriol Vinyals}, \bibinfo{person}{Alex Graves},
  \bibinfo{person}{Nal Kalchbrenner}, \bibinfo{person}{Andrew Senior}, {and}
  \bibinfo{person}{Koray Kavukcuoglu}.} \bibinfo{year}{2016}\natexlab{}.
\newblock \showarticletitle{{WaveNet: A Generative Model for Raw Audio}}.
\newblock \bibinfo{journal}{\emph{CoRR abs/1609.03499}} (\bibinfo{year}{2016}).
\newblock
\urldef\tempurl%
\url{http://arxiv.org/abs/1609.03499}
\showURL{%
\tempurl}


\bibitem[\protect\citeauthoryear{Vepakomma, Gupta, Swedish, and
  Raskar}{Vepakomma et~al\mbox{.}}{2018}]%
        {Vepakomma2018}
\bibfield{author}{\bibinfo{person}{Praneeth Vepakomma},
  \bibinfo{person}{Otkrist Gupta}, \bibinfo{person}{Tristan Swedish}, {and}
  \bibinfo{person}{Ramesh Raskar}.} \bibinfo{year}{2018}\natexlab{}.
\newblock \showarticletitle{{Split learning for health: Distributed deep
  learning without sharing raw patient data}}.
\newblock \bibinfo{journal}{\emph{arXiv:1812.00564}} (\bibinfo{year}{2018}).
\newblock
\urldef\tempurl%
\url{http://arxiv.org/abs/1812.00564}
\showURL{%
\tempurl}


\bibitem[\protect\citeauthoryear{Vinyals, Toshev, Bengio, and Erhan}{Vinyals
  et~al\mbox{.}}{2015}]%
        {Vinyals2015}
\bibfield{author}{\bibinfo{person}{Oriol Vinyals}, \bibinfo{person}{Alexander
  Toshev}, \bibinfo{person}{Samy Bengio}, {and} \bibinfo{person}{Dumitru
  Erhan}.} \bibinfo{year}{2015}\natexlab{}.
\newblock \showarticletitle{{Show and tell: A neural image caption generator}}.
  In \bibinfo{booktitle}{\emph{Proceedings of the IEEE Computer Society
  Conference on Computer Vision and Pattern Recognition (CVPR'15)}},
  Vol.~\bibinfo{volume}{07-12-June}. \bibinfo{pages}{3156--3164}.
\newblock
\showISBNx{9781467369640}
\showISSN{10636919}
\urldef\tempurl%
\url{https://doi.org/10.1109/CVPR.2015.7298935}
\showDOI{\tempurl}


\bibitem[\protect\citeauthoryear{Wang and Gong}{Wang and Gong}{2018}]%
        {Wang2018}
\bibfield{author}{\bibinfo{person}{Binghui Wang} {and}
  \bibinfo{person}{Neil~Zhenqiang Gong}.} \bibinfo{year}{2018}\natexlab{}.
\newblock \showarticletitle{{Stealing Hyperparameters in Machine Learning}}. In
  \bibinfo{booktitle}{\emph{Proceedings of the IEEE Symposium on Security and
  Privacy (SP'18)}}. \bibinfo{pages}{36--52}.
\newblock
\showISBNx{9781538643525}
\showISSN{10816011}
\urldef\tempurl%
\url{https://doi.org/10.1109/SP.2018.00038}
\showDOI{\tempurl}


\bibitem[\protect\citeauthoryear{Wang, Chen, Fung, and Yu}{Wang
  et~al\mbox{.}}{2010}]%
        {Wang2010}
\bibfield{author}{\bibinfo{person}{K Wang}, \bibinfo{person}{R Chen},
  \bibinfo{person}{B~C Fung}, {and} \bibinfo{person}{P~S Yu}.}
  \bibinfo{year}{2010}\natexlab{}.
\newblock \showarticletitle{{Privacy-preserving data publishing: A survey on
  recent developments}}.
\newblock \bibinfo{journal}{\emph{Comput. Surveys}} (\bibinfo{year}{2010}).
\newblock


\bibitem[\protect\citeauthoryear{Wang, Song, Zhang, Song, Wang, and Qi}{Wang
  et~al\mbox{.}}{2019}]%
        {Wang2019}
\bibfield{author}{\bibinfo{person}{Zhibo Wang}, \bibinfo{person}{Mengkai Song},
  \bibinfo{person}{Zhifei Zhang}, \bibinfo{person}{Yang Song},
  \bibinfo{person}{Qian Wang}, {and} \bibinfo{person}{Hairong Qi}.}
  \bibinfo{year}{2019}\natexlab{}.
\newblock \showarticletitle{{Beyond Inferring Class Representatives: User-Level
  Privacy Leakage from Federated Learning}}.
\newblock \bibinfo{journal}{\emph{Proceedings of IEEE International Conference
  on Computer Communications (INFOCOM'19)}}  \bibinfo{volume}{2019-April}
  (\bibinfo{year}{2019}), \bibinfo{pages}{2512--2520}.
\newblock
\showISBNx{9781728105154}
\showISSN{0743166X}
\urldef\tempurl%
\url{https://doi.org/10.1109/INFOCOM.2019.8737416}
\showDOI{\tempurl}


\bibitem[\protect\citeauthoryear{Wei, Luo, Li, Liu, and Xu}{Wei
  et~al\mbox{.}}{2018}]%
        {Wei2018}
\bibfield{author}{\bibinfo{person}{Lingxiao Wei}, \bibinfo{person}{Bo Luo},
  \bibinfo{person}{Yu Li}, \bibinfo{person}{Yannan Liu}, {and}
  \bibinfo{person}{Qiang Xu}.} \bibinfo{year}{2018}\natexlab{}.
\newblock \showarticletitle{{I know what you see: Power side-channel attack on
  convolutional neural network accelerators}}. In \bibinfo{booktitle}{\emph{ACM
  International Conference Proceeding Series}}. \bibinfo{pages}{393--406}.
\newblock
\showISBNx{9781450365697}
\urldef\tempurl%
\url{https://doi.org/10.1145/3274694.3274696}
\showDOI{\tempurl}


\bibitem[\protect\citeauthoryear{Wijesekera, Baokar, Hosseini, Egelman, Wagner,
  and Beznosov}{Wijesekera et~al\mbox{.}}{2015}]%
        {Wijesekera2015}
\bibfield{author}{\bibinfo{person}{Primal Wijesekera}, \bibinfo{person}{Arjun
  Baokar}, \bibinfo{person}{Ashkan Hosseini}, \bibinfo{person}{Serge Egelman},
  \bibinfo{person}{David Wagner}, {and} \bibinfo{person}{Konstantin Beznosov}.}
  \bibinfo{year}{2015}\natexlab{}.
\newblock \showarticletitle{{Android permissions remystified: A field study on
  contextual integrity}}. In \bibinfo{booktitle}{\emph{Proceedings of the 24th
  USENIX Security Symposium (USENIX'15)}}. \bibinfo{pages}{499--514}.
\newblock
\showISBNx{9781931971232}


\bibitem[\protect\citeauthoryear{Wijesekera, Baokar, Tsai, Reardon, Egelman,
  Wagner, and Beznosov}{Wijesekera et~al\mbox{.}}{2017}]%
        {Wijesekera2017}
\bibfield{author}{\bibinfo{person}{Primal Wijesekera}, \bibinfo{person}{Arjun
  Baokar}, \bibinfo{person}{Lynn Tsai}, \bibinfo{person}{Joel Reardon},
  \bibinfo{person}{Serge Egelman}, \bibinfo{person}{David Wagner}, {and}
  \bibinfo{person}{Konstantin Beznosov}.} \bibinfo{year}{2017}\natexlab{}.
\newblock \showarticletitle{{The Feasibility of Dynamically Granted
  Permissions: Aligning Mobile Privacy with User Preferences}}. In
  \bibinfo{booktitle}{\emph{Proceedings of the IEEE Symposium on Security and
  Privacy (SP'17)}}. \bibinfo{pages}{1077--1093}.
\newblock
\showISBNx{9781509055326}
\showISSN{10816011}
\urldef\tempurl%
\url{https://doi.org/10.1109/SP.2017.51}
\showDOI{\tempurl}


\bibitem[\protect\citeauthoryear{Wijesekera, Reardon, Reyes, Tsai, Chen, Good,
  Wagner, Beznosov, and Egelman}{Wijesekera et~al\mbox{.}}{2018}]%
        {Wijesekera2018}
\bibfield{author}{\bibinfo{person}{Primal Wijesekera}, \bibinfo{person}{Joel
  Reardon}, \bibinfo{person}{Irwin Reyes}, \bibinfo{person}{Lynn Tsai},
  \bibinfo{person}{Jung~Wei Chen}, \bibinfo{person}{Nathan Good},
  \bibinfo{person}{David Wagner}, \bibinfo{person}{Konstantin Beznosov}, {and}
  \bibinfo{person}{Serge Egelman}.} \bibinfo{year}{2018}\natexlab{}.
\newblock \showarticletitle{{Contextualizing privacy decisions for better
  prediction (and protection)}}. In \bibinfo{booktitle}{\emph{Proceedings of
  the Conference on Human Factors in Computing Systems (CHI'18)}},
  Vol.~\bibinfo{volume}{2018-April}. \bibinfo{pages}{268}.
\newblock
\showISBNx{9781450356206}
\urldef\tempurl%
\url{https://doi.org/10.1145/3173574.3173842}
\showDOI{\tempurl}


\bibitem[\protect\citeauthoryear{Wilber, Shmatikov, and Belongie}{Wilber
  et~al\mbox{.}}{2016}]%
        {Wilber2016}
\bibfield{author}{\bibinfo{person}{Michael~J Wilber}, \bibinfo{person}{Vitaly
  Shmatikov}, {and} \bibinfo{person}{Serge Belongie}.}
  \bibinfo{year}{2016}\natexlab{}.
\newblock \showarticletitle{{Can we still avoid automatic face detection?}}. In
  \bibinfo{booktitle}{\emph{Proceedings of the IEEE Winter Conference on
  Applications of Computer Vision (WACV'16)}}. \bibinfo{pages}{1--9}.
\newblock
\showISBNx{9781509006410}
\urldef\tempurl%
\url{https://doi.org/10.1109/WACV.2016.7477452}
\showDOI{\tempurl}


\bibitem[\protect\citeauthoryear{Wu, Farokhi, Smith, and Kaafar}{Wu
  et~al\mbox{.}}{2020}]%
        {wu2019value}
\bibfield{author}{\bibinfo{person}{Nan Wu}, \bibinfo{person}{Farhad Farokhi},
  \bibinfo{person}{David Smith}, {and} \bibinfo{person}{Mohamed~Ali Kaafar}.}
  \bibinfo{year}{2020}\natexlab{}.
\newblock \showarticletitle{The value of collaboration in convex machine
  learning with differential privacy}. In \bibinfo{booktitle}{\emph{2020 IEEE
  Symposium on Security and Privacy (SP)}}. \bibinfo{pages}{304--317}.
\newblock


\bibitem[\protect\citeauthoryear{Wu, Teruya, Kawamoto, Sakuma, and Kikuchi}{Wu
  et~al\mbox{.}}{2013}]%
        {Wu2013}
\bibfield{author}{\bibinfo{person}{Shuang Wu}, \bibinfo{person}{Tadanori
  Teruya}, \bibinfo{person}{Junpei Kawamoto}, \bibinfo{person}{Jun Sakuma},
  {and} \bibinfo{person}{Hiroaki Kikuchi}.} \bibinfo{year}{2013}\natexlab{}.
\newblock \showarticletitle{{Privacy-preservation for Stochastic Gradient
  Descent Application to Secure Logistic Regression}}. In
  \bibinfo{booktitle}{\emph{Proceedings of the 27th Annual Conference of the
  Japanese Society for Artificial Intelligence (JSAI'13)}}.
  \bibinfo{pages}{6--9}.
\newblock


\bibitem[\protect\citeauthoryear{Xu and Veeramachaneni}{Xu and
  Veeramachaneni}{2018}]%
        {Xu2018}
\bibfield{author}{\bibinfo{person}{Lei Xu} {and} \bibinfo{person}{Kalyan
  Veeramachaneni}.} \bibinfo{year}{2018}\natexlab{}.
\newblock \showarticletitle{{Synthesizing Tabular Data using Generative
  Adversarial Networks}}.
\newblock \bibinfo{journal}{\emph{arXiv:1811.11264}} (\bibinfo{year}{2018}).
\newblock
\urldef\tempurl%
\url{http://arxiv.org/abs/1811.11264}
\showURL{%
\tempurl}


\bibitem[\protect\citeauthoryear{Yao, Zheng, Li, and Zhao}{Yao
  et~al\mbox{.}}{2019}]%
        {Yao2019}
\bibfield{author}{\bibinfo{person}{Yuanshun Yao}, \bibinfo{person}{Haitao
  Zheng}, \bibinfo{person}{Huiying Li}, {and} \bibinfo{person}{Ben~Y. Zhao}.}
  \bibinfo{year}{2019}\natexlab{}.
\newblock \showarticletitle{{Latent backdoor attacks on deep neural networks}}.
  In \bibinfo{booktitle}{\emph{Proceedings of the ACM Conference on Computer
  and Communications Security (CCS'19)}}. \bibinfo{pages}{2041--2055}.
\newblock
\showISBNx{9781450367479}
\showISSN{15437221}
\urldef\tempurl%
\url{https://doi.org/10.1145/3319535.3354209}
\showDOI{\tempurl}


\bibitem[\protect\citeauthoryear{Yeom, Giacomelli, Fredrikson, and Jha}{Yeom
  et~al\mbox{.}}{2018}]%
        {Yeom2018}
\bibfield{author}{\bibinfo{person}{Samuel Yeom}, \bibinfo{person}{Irene
  Giacomelli}, \bibinfo{person}{Matt Fredrikson}, {and} \bibinfo{person}{Somesh
  Jha}.} \bibinfo{year}{2018}\natexlab{}.
\newblock \showarticletitle{{Privacy risk in machine learning: Analyzing the
  connection to overfitting}}. In \bibinfo{booktitle}{\emph{Proceedings of the
  IEEE Computer Security Foundations Symposium (CSF'18)}},
  Vol.~\bibinfo{volume}{2018-July}. \bibinfo{pages}{268--282}.
\newblock
\showISBNx{9781538666807}
\showISSN{19401434}
\urldef\tempurl%
\url{https://doi.org/10.1109/CSF.2018.00027}
\showDOI{\tempurl}


\bibitem[\protect\citeauthoryear{Yu, Zhang, Kuang, Lin, and Fan}{Yu
  et~al\mbox{.}}{2017}]%
        {Yu2017}
\bibfield{author}{\bibinfo{person}{Jun Yu}, \bibinfo{person}{Baopeng Zhang},
  \bibinfo{person}{Zhengzhong Kuang}, \bibinfo{person}{Dan Lin}, {and}
  \bibinfo{person}{Jianping Fan}.} \bibinfo{year}{2017}\natexlab{}.
\newblock \showarticletitle{{IPrivacy: Image Privacy Protection by Identifying
  Sensitive Objects via Deep Multi-Task Learning}}.
\newblock \bibinfo{journal}{\emph{IEEE Trans. Inf. Forensics Security}}
  \bibinfo{volume}{12}, \bibinfo{number}{5} (\bibinfo{year}{2017}),
  \bibinfo{pages}{1005--1016}.
\newblock
\showISSN{15566013}
\urldef\tempurl%
\url{https://doi.org/10.1109/TIFS.2016.2636090}
\showDOI{\tempurl}


\bibitem[\protect\citeauthoryear{Yuan, Theytaz, and Ebrahimi}{Yuan
  et~al\mbox{.}}{2017}]%
        {Yuan2017}
\bibfield{author}{\bibinfo{person}{Lin Yuan}, \bibinfo{person}{Jo{\"{e}}l
  Theytaz}, {and} \bibinfo{person}{Touradj Ebrahimi}.}
  \bibinfo{year}{2017}\natexlab{}.
\newblock \showarticletitle{{Context-dependent privacy-aware photo sharing
  based on machine learning}}. In \bibinfo{booktitle}{\emph{IFIP Advances in
  Information and Communication Technology}}, Vol.~\bibinfo{volume}{502}.
  \bibinfo{pages}{93--107}.
\newblock
\showISBNx{9783319584683}
\showISSN{18684238}
\urldef\tempurl%
\url{https://doi.org/10.1007/978-3-319-58469-0_7}
\showDOI{\tempurl}


\bibitem[\protect\citeauthoryear{Zerr, Siersdorfer, Hare, and Demidova}{Zerr
  et~al\mbox{.}}{2012}]%
        {Zerr2012}
\bibfield{author}{\bibinfo{person}{Sergej Zerr}, \bibinfo{person}{Stefan
  Siersdorfer}, \bibinfo{person}{Jonathon Hare}, {and} \bibinfo{person}{Elena
  Demidova}.} \bibinfo{year}{2012}\natexlab{}.
\newblock \showarticletitle{{Privacy-aware image classification and search}}.
  In \bibinfo{booktitle}{\emph{Proceedings of the International ACM SIGIR
  Conference on Research and Development in Information Retrieval (SIGIR'12)}}.
  \bibinfo{pages}{35--44}.
\newblock
\showISBNx{9781450316583}
\urldef\tempurl%
\url{https://doi.org/10.1145/2348283.2348292}
\showDOI{\tempurl}


\bibitem[\protect\citeauthoryear{Zhang, Chen, Wang, and Shi}{Zhang
  et~al\mbox{.}}{2018a}]%
        {Zhang2018}
\bibfield{author}{\bibinfo{person}{Dayin Zhang}, \bibinfo{person}{Xiaojun
  Chen}, \bibinfo{person}{Dakui Wang}, {and} \bibinfo{person}{Jinqiao Shi}.}
  \bibinfo{year}{2018}\natexlab{a}.
\newblock \showarticletitle{{A survey on collaborative deep learning and
  privacy-preserving}}. In \bibinfo{booktitle}{\emph{Proceedings of the IEEE
  3rd International Conference on Data Science in Cyberspace (DSC'18)}}.
  \bibinfo{publisher}{IEEE}, \bibinfo{pages}{652--658}.
\newblock
\showISBNx{9781538642108}
\urldef\tempurl%
\url{https://doi.org/10.1109/DSC.2018.00104}
\showDOI{\tempurl}


\bibitem[\protect\citeauthoryear{Zhang, Zhang, Xiao, Yang, and Winslett}{Zhang
  et~al\mbox{.}}{2012}]%
        {Zhang2012}
\bibfield{author}{\bibinfo{person}{Jun Zhang}, \bibinfo{person}{Zhenjie Zhang},
  \bibinfo{person}{Xiaokui Xiao}, \bibinfo{person}{Yin Yang}, {and}
  \bibinfo{person}{Marianne Winslett}.} \bibinfo{year}{2012}\natexlab{}.
\newblock \showarticletitle{{Functional mechanism: Regression analysis under
  differential privacy}}.
\newblock \bibinfo{journal}{\emph{Proceedings of the International Conference
  on Very Large Data Bases (VLDB'12)}} \bibinfo{volume}{5},
  \bibinfo{number}{11} (\bibinfo{year}{2012}), \bibinfo{pages}{1364--1375}.
\newblock
\showISSN{21508097}
\urldef\tempurl%
\url{https://doi.org/10.14778/2350229.2350253}
\showDOI{\tempurl}


\bibitem[\protect\citeauthoryear{Zhang, He, and Lee}{Zhang
  et~al\mbox{.}}{2018b}]%
        {Zhang2018a}
\bibfield{author}{\bibinfo{person}{Tianwei Zhang}, \bibinfo{person}{Zecheng
  He}, {and} \bibinfo{person}{Ruby~B. Lee}.} \bibinfo{year}{2018}\natexlab{b}.
\newblock \showarticletitle{{Privacy-preserving Machine Learning through Data
  Obfuscation}}.
\newblock \bibinfo{journal}{\emph{arXiv:1807.01860}} (\bibinfo{year}{2018}).
\newblock
\urldef\tempurl%
\url{http://arxiv.org/abs/1807.01860}
\showURL{%
\tempurl}


\bibitem[\protect\citeauthoryear{Zhang, Ji, and Wang}{Zhang
  et~al\mbox{.}}{2018c}]%
        {Zhang2018b}
\bibfield{author}{\bibinfo{person}{Xinyang Zhang}, \bibinfo{person}{Shouling
  Ji}, {and} \bibinfo{person}{Ting Wang}.} \bibinfo{year}{2018}\natexlab{c}.
\newblock \showarticletitle{{Differentially Private Releasing via Deep
  Generative Model (Technical Report)}}.
\newblock \bibinfo{journal}{\emph{arXiv:1801.01594}} (\bibinfo{year}{2018}).
\newblock
\urldef\tempurl%
\url{http://arxiv.org/abs/1801.01594}
\showURL{%
\tempurl}


\bibitem[\protect\citeauthoryear{Zhou, Li, Lu, and Tian}{Zhou
  et~al\mbox{.}}{2012}]%
        {Zhou2012}
\bibfield{author}{\bibinfo{person}{Wengang Zhou}, \bibinfo{person}{Houqiang
  Li}, \bibinfo{person}{Yijuan Lu}, {and} \bibinfo{person}{Qi Tian}.}
  \bibinfo{year}{2012}\natexlab{}.
\newblock \showarticletitle{{Principal visual word discovery for automatic
  license plate detection}}.
\newblock \bibinfo{journal}{\emph{IEEE Trans. Image Process.}}
  \bibinfo{volume}{21}, \bibinfo{number}{9} (\bibinfo{year}{2012}),
  \bibinfo{pages}{4269--4279}.
\newblock
\showISSN{10577149}
\urldef\tempurl%
\url{https://doi.org/10.1109/TIP.2012.2199506}
\showDOI{\tempurl}


\end{thebibliography}

%

\end{document}